\documentclass{article}

\usepackage{microtype}
\usepackage{graphicx}
\usepackage{subfigure}
\usepackage{booktabs} % for professional tables
\usepackage{caption}

\usepackage{hyperref}

\usepackage[accepted]{icml2025}

\usepackage{amsmath}
\usepackage{amssymb}
\usepackage{mathtools}
\usepackage{amsthm}

\usepackage[capitalize,noabbrev]{cleveref}

\theoremstyle{plain}

\theoremstyle{definition}

\theoremstyle{remark}

\usepackage[textsize=tiny]{todonotes}

\def\softmax{\boldsymbol{\sigma}_{\text{SM}}}

\icmltitlerunning{SketchDNN: Joint Continuous-Discrete Diffusion for CAD Sketch Generation}

\begin{document}

\twocolumn[
\icmltitle{SketchDNN: Joint Continuous-Discrete Diffusion for CAD Sketch Generation}

% It is OKAY to include author information, even for blind
% submissions: the style file will automatically remove it for you
% unless you've provided the [accepted] option to the icml2025
% package.

% List of affiliations: The first argument should be a (short)
% identifier you will use later to specify author affiliations
% Academic affiliations should list Department, University, City, Region, Country
% Industry affiliations should list Company, City, Region, Country

% You can specify symbols, otherwise they are numbered in order.
% Ideally, you should not use this facility. Affiliations will be numbered
% in order of appearance and this is the preferred way.
\icmlsetsymbol{equal}{*}

\begin{icmlauthorlist}
\icmlauthor{Sathvik Chereddy}{yyy}
\icmlauthor{John Femiani}{yyy}
%\icmlauthor{}{sch}
%\icmlauthor{}{sch}
\end{icmlauthorlist}

\icmlaffiliation{yyy}{Department of Computer Science, Miami-Oxford University, Oxford OH, USA}
% \icmlaffiliation{comp}{Company Name, Location, Country}
% \icmlaffiliation{sch}{School of ZZZ, Institute of WWW, Location, Country}

\icmlcorrespondingauthor{Sathvik Chereddy (M.S.)}{sathware@outlook.com}
\icmlcorrespondingauthor{John Femiani}{femianjc@miamioh.edu}

% You may provide any keywords that you
% find helpful for describing your paper; these are used to populate
% the "keywords" metadata in the PDF but will not be shown in the document
\icmlkeywords{Diffusion, Discrete Diffusion, CAD, CAD Sketch, Generative, AI, ML, Gaussian-Softmax}

% TEASER FIGURE ---------
{
    \renewcommand\tabcolsep{0pt}
    \centering
    \includegraphics[clip, trim=3cm 1.5cm 3cm 1.5cm, width=0.22\linewidth]{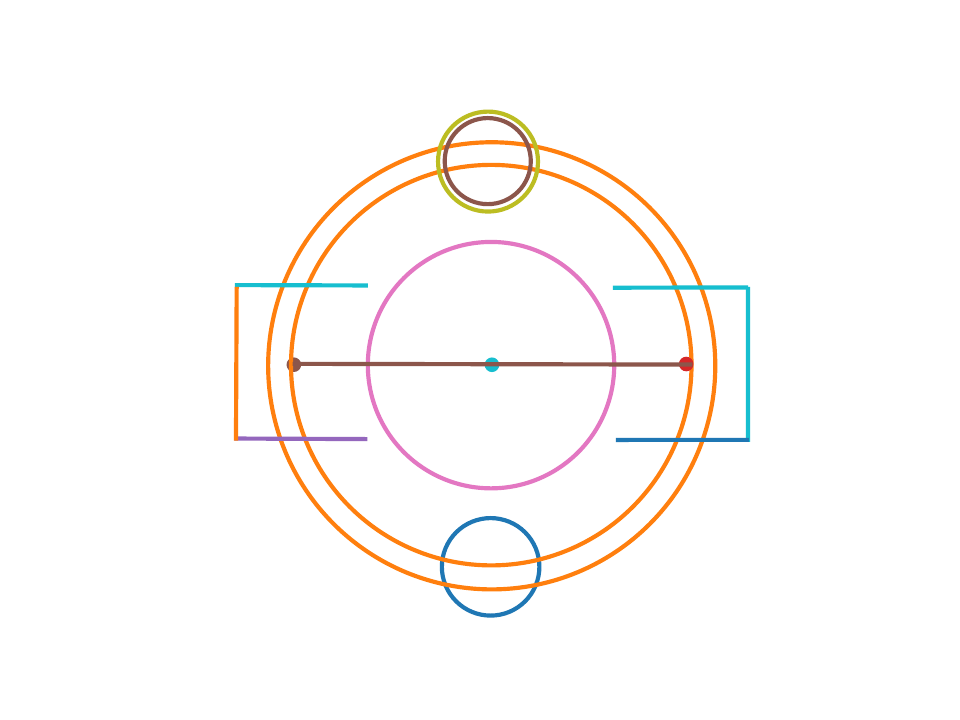}
    \includegraphics[clip, trim=3cm 1.5cm 3cm 1.5cm, width=0.22\linewidth]{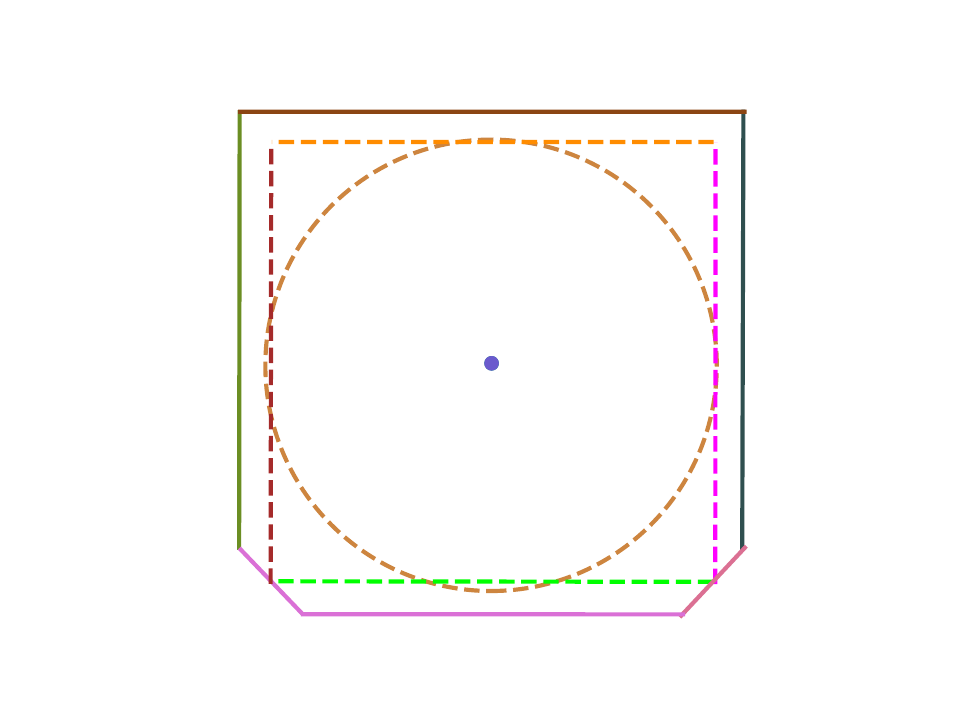}
    \includegraphics[clip, trim=3cm 1.5cm 3cm 1.5cm, width=0.22\linewidth]{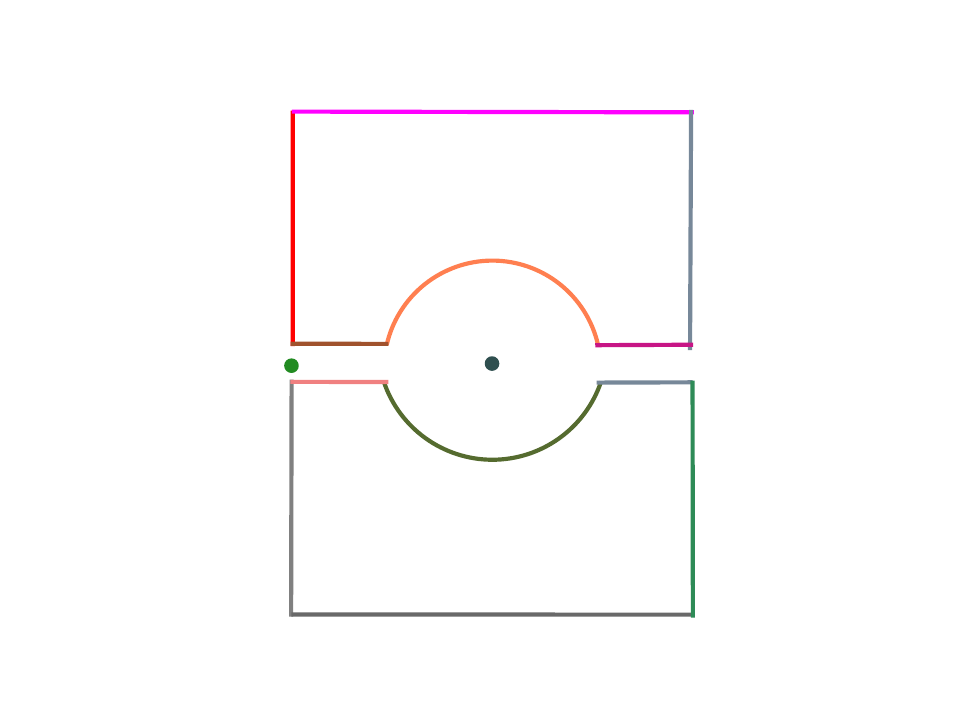}
    \includegraphics[clip, trim=3cm 1.5cm 3cm 1.5cm, width=0.22\linewidth]{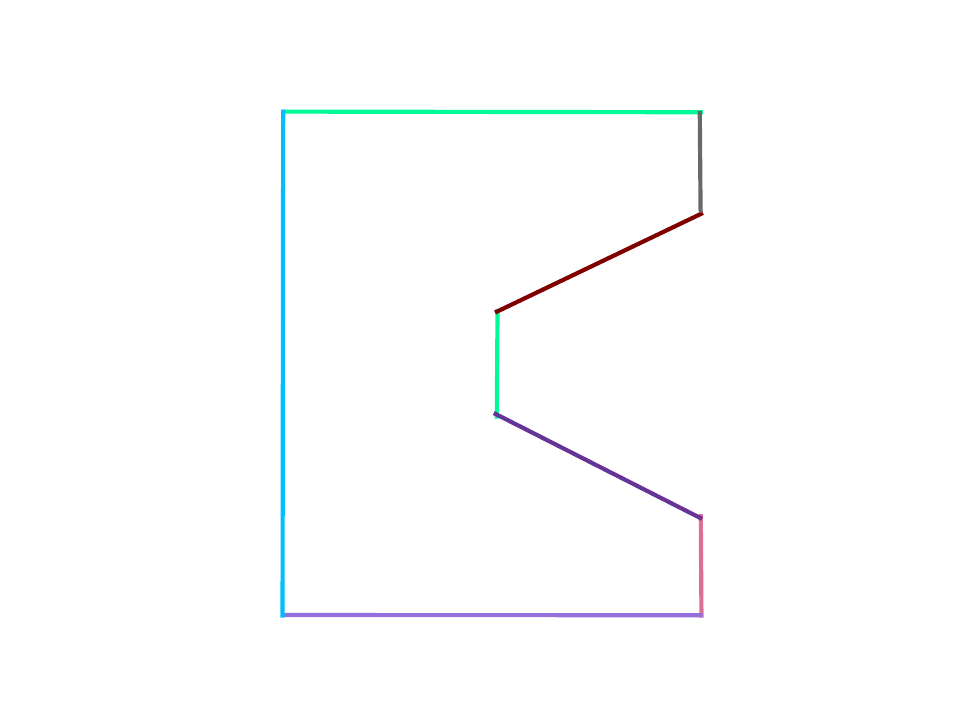}
    \includegraphics[clip, trim=3cm 1.5cm 3cm 1.5cm, width=0.22\linewidth]{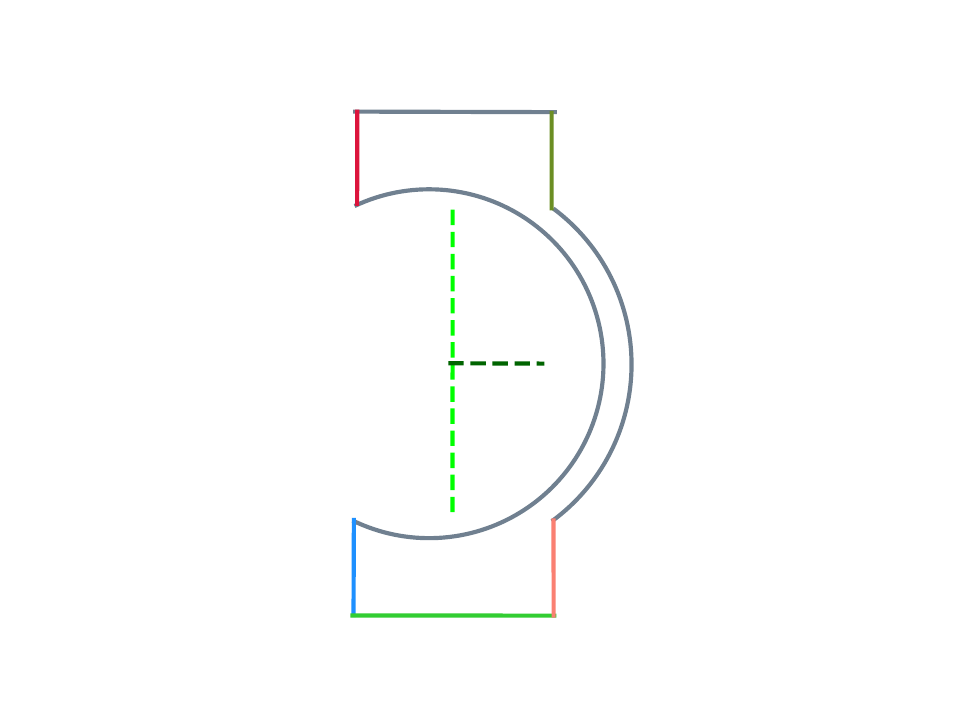}
    \includegraphics[clip, trim=2cm 1.5cm 2cm 1.5cm, width=0.22\linewidth]{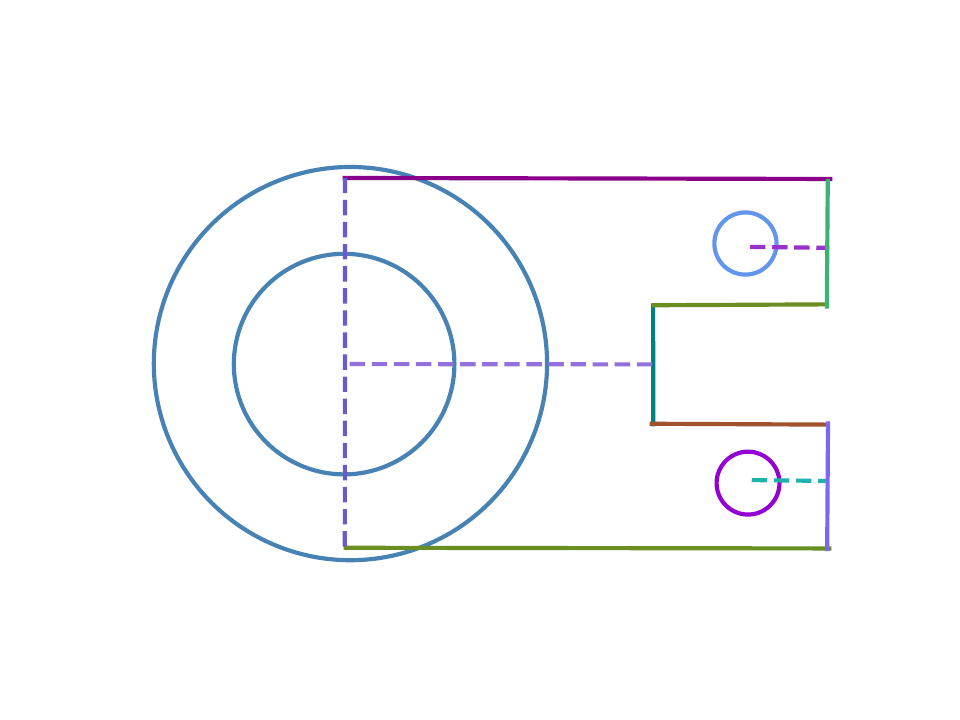}
    \includegraphics[clip, trim=3cm 1.5cm 3cm 1.5cm, width=0.22\linewidth]{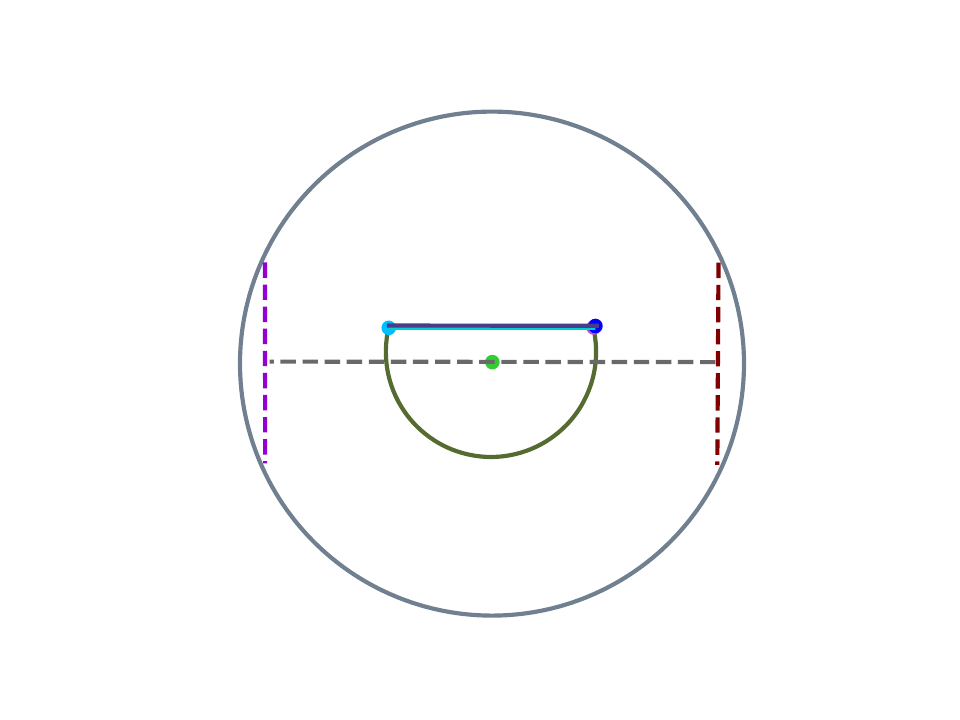}
    \includegraphics[clip, trim=3cm 1.5cm 3cm 1.5cm, width=0.22\linewidth]{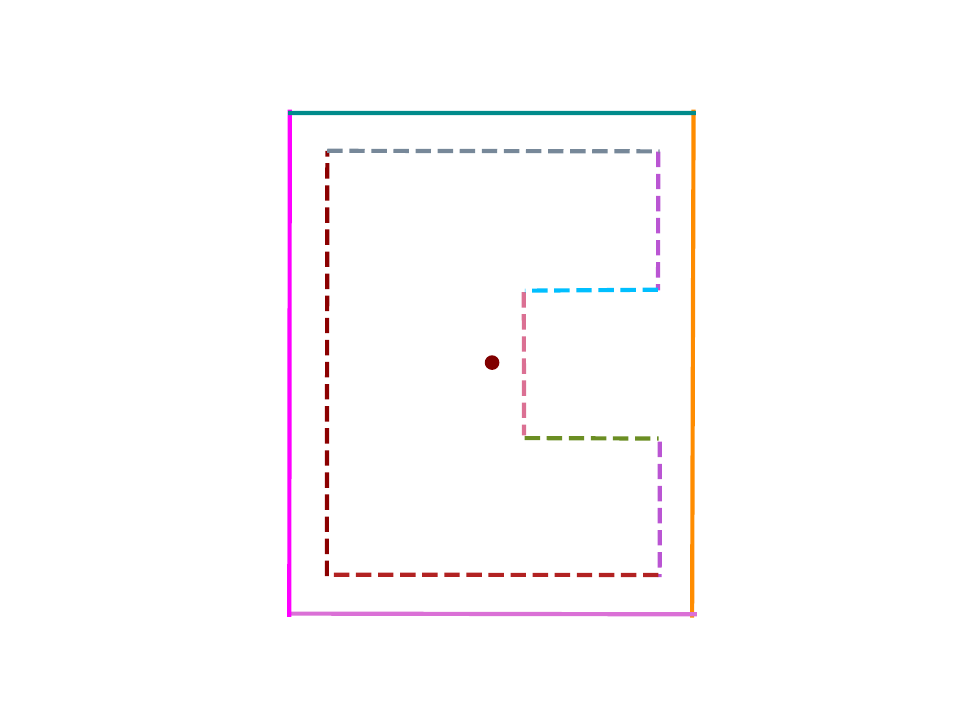}
    \includegraphics[clip, trim=3cm 1.5cm 3cm 1.5cm, width=0.22\linewidth]{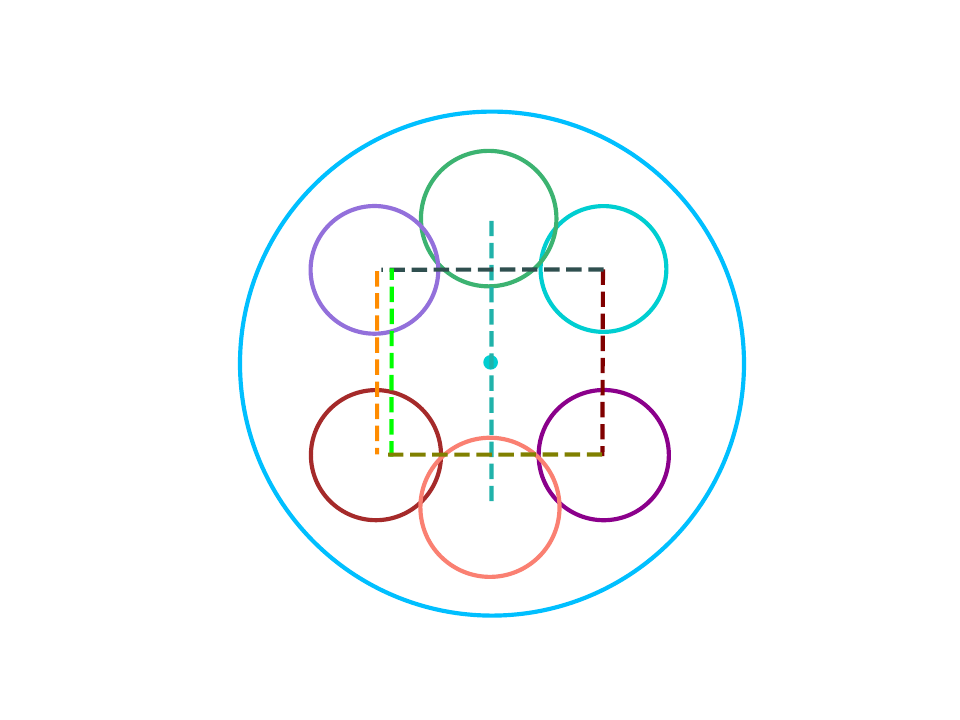}
    \includegraphics[clip, trim=3cm 1.5cm 3cm 1.5cm, width=0.22\linewidth]{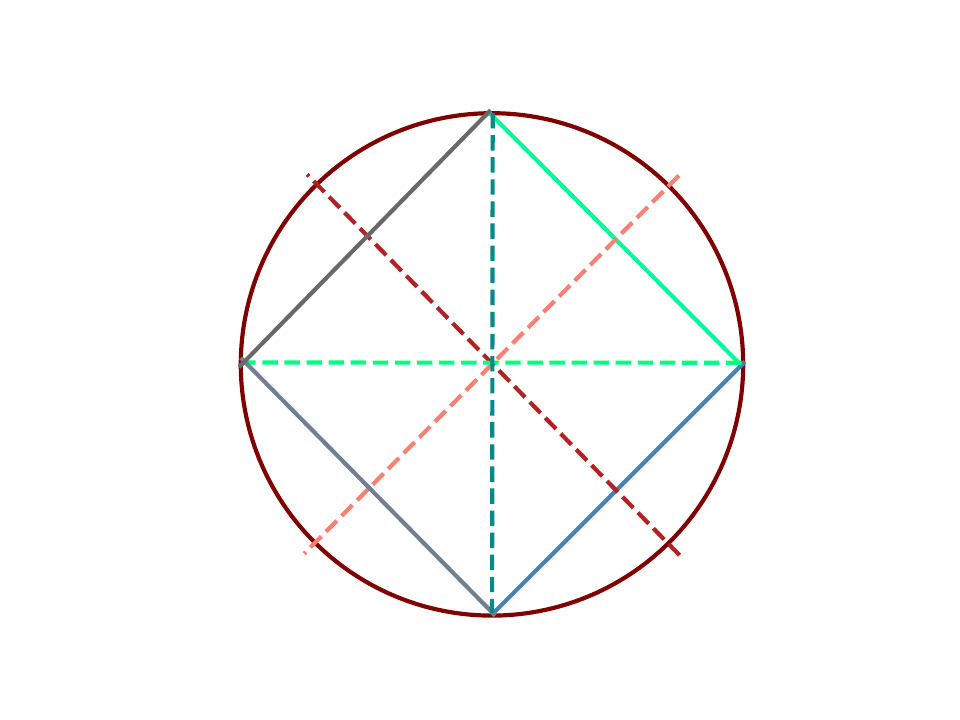}
    \includegraphics[clip, trim=3cm 1.5cm 3cm 1.5cm, width=0.22\linewidth]{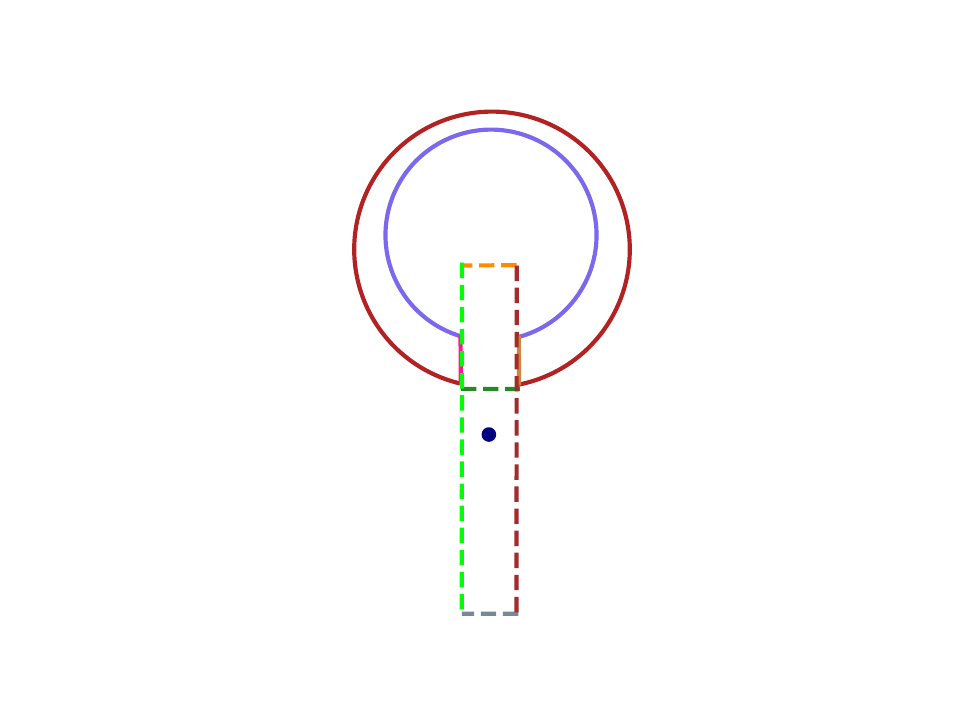}
    \includegraphics[clip, trim=3cm 1.5cm 3cm 1.5cm, width=0.22\linewidth]{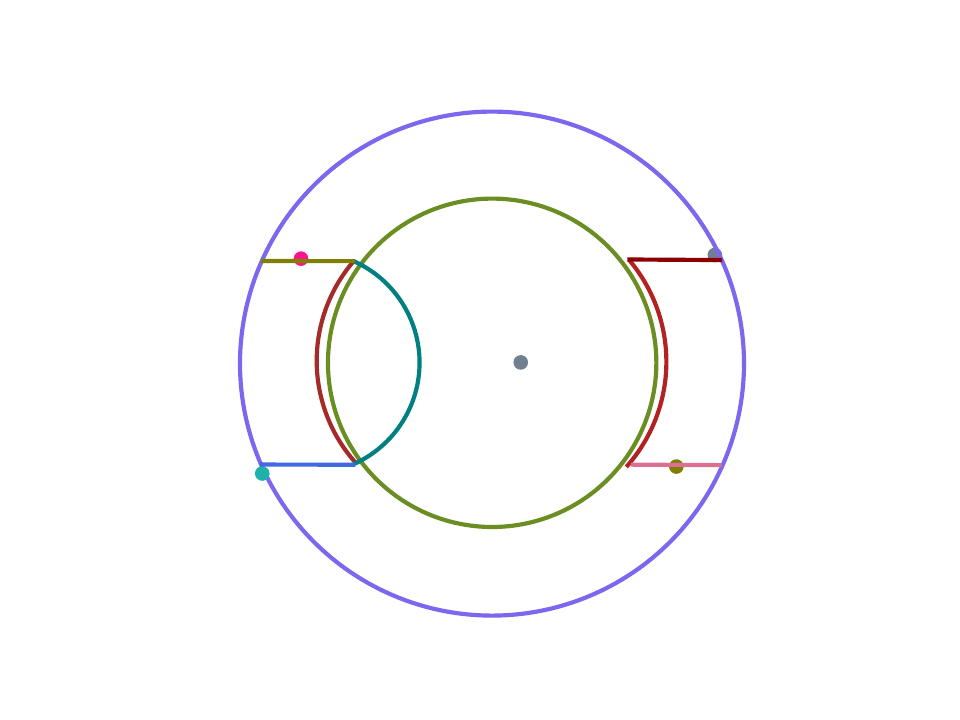}
    
    \captionof{figure}{CAD sketches generated by our diffusion model, showcasing its ability to produce diverse, high-fidelity designs. Geometric primitives—such as circles, arcs, lines, and points—are randomly colored to differentiate separate primitives, and primitives tagged as construction aids are represented with dashed lines.}
    \label{fig:teaser}
    \bigskip
}

\vskip 0.3in
]

% this must go after the closing bracket ] following \twocolumn[ ...

% This command actually creates the footnote in the first column
% listing the affiliations and the copyright notice.
% The command takes one argument, which is text to display at the start of the footnote.
% The \icmlEqualContribution command is standard text for equal contribution.
% Remove it (just {}) if you do not need this facility.

\printAffiliationsAndNotice{}  % leave blank if no need to mention equal contribution
% \printAffiliationsAndNotice{\icmlEqualContribution} % otherwise use the standard text.

\begin{abstract}
%Typical computer-aided design (CAD) workflows require manual construction of 2D planar sketches, which if automated can streamline CAD design, improve user productivity, and unlock new workflows. While existing methods have focused on autoregressive synthesis of sketches similar to language models, alternative generative paradigms remain relatively unexplored. To address this, 
% We present SketchDNN, a generative model that synthesizes CAD sketches by modeling both the continuous parameters and discrete class labels of primitives using a joint continuous-discrete diffusion process. A key component of SketchDNN is a novel Gaussian-Softmax diffusion strategy for discrete variables, which perturbs logits with Gaussian noise and projects them onto the probability simplex using a softmax transformation, enabling blended class labels in the diffusion process as opposed to conventional methods. Our model achieves state-of-the-art results, reducing the Fréchet Inception Distance (FID) from 16.04 to 7.80 and the negative log-likelihood (NLL) from 84.8 to 81.33.

We present SketchDNN, a generative model for synthesizing CAD sketches that jointly models both continuous parameters and discrete class labels through a unified continuous-discrete diffusion process. Our core innovation is Gaussian-Softmax diffusion, where logits perturbed with Gaussian noise are projected onto the probability simplex via a softmax transformation, facilitating blended class labels for discrete variables. This formulation addresses 2 key challenges, namely, the heterogeneity of primitive parameterizations and the permutation invariance of primitives in CAD sketches. Our approach significantly improves generation quality, reducing Fréchet Inception Distance (FID) from 16.04 to 7.80 and negative log-likelihood (NLL) from 84.8 to 81.33, establishing a new state-of-the-art in CAD sketch generation on the SketchGraphs dataset.
\end{abstract}

\section{Introduction}

Computer-aided design (CAD) modeling is a cornerstone of modern engineering that is heavily reliant on human ingenuity. The typical workflow in CAD modeling starts with designing 2D planar sketches comprised of primitives (e.g., lines, arcs) that define geometry and constraints (e.g., coincidence, orthogonality) that define relationships between primitives. Such diagrams are then expanded into 3D volumes using operations such as extrusion and revolution. Over the course of the CAD modeling process, these 3D volumes are iteratively aggregated to form the final 3D model. As a result, sketch construction takes a substantial portion of user's time and effort in CAD design.

A number of researchers have identified CAD diagrams as an ideal domain for generative models \cite{seff_sketchgraphs_2020, ganin_computer-aided_2021, para_sketchgen_2021, Willis2021EngineeringSG, Wu2021DeepCADAD, seff2021vitruvion}. Primitive generation is of particular interest, as constraints do not define geometry but instead serve as editing aids for users. Analogous to image generative models such as Stable Diffusion, CAD generative models can streamline design and improve user productivity \cite{seff_sketchgraphs_2020}. By automating low-level CAD drawing construction, these models could allow designers to focus more on higher-level design tasks \cite{ganin_computer-aided_2021}. Furthermore, they could facilitate novel workflows \cite{para_sketchgen_2021}, such as simultaneously exploring multiple sketch variations to meet design requirements.

However, generating CAD sketches is difficult due to two fundamental challenges. The first challenge arises from the heterogeneity of sketch primitives, as each primitive type is defined by its own distinct parameterization. The second challenge lies in the permutation invariance of primitives, where any ordering of primitives encodes the same geometry. This poses a particular challenge for autoregressive models, the dominant generation paradigm in this domain, as they generate primitives sequentially. While prior approaches have made significant progress, they do not adequately address these challenges, which we believe limits the diversity and quality of their generated sketches. Diffusion models, which have demonstrated remarkable success in image generation, are a promising avenue for addressing these challenges.

In this work, we introduce a novel generative diffusion model for unconditional primitive generation and propose a discrete diffusion strategy built on the Gaussian-Softmax distribution. Our method addresses both challenges in CAD sketch generation by employing superposition, wherein each primitive is represented as a probabilistic mixture of all primitive types, and permutation-invariant denoising, where the denoising of each primitive is independent of its ordering. Our approach improves the state-of-the-art (SOA) in terms of both generation quality and diversity.

Our key contributions are as follows:
\begin{enumerate}
\item We propose the first sketch-space/data-space generative diffusion model for parametric CAD sketches.
\item We introduce a Gaussian-Softmax based diffusion paradigm for modeling discrete variables.
\item We advance the state of the art in CAD sketch generation. Our model reduces the Fréchet Inception Distance (FID) from 16.04 to 7.80 and the Negative Log-Likelihood (NLL) from 84.8 to 81.33 on the SketchGraphs dataset, demonstrating significant improvements in both fidelity and diversity.
\end{enumerate}

\section{CAD Sketches}\label{sec:representation}

A CAD sketch is naturally represented as a graph \(\mathcal{G} = (V, E)\), where the set of nodes \(V = \{v_i\}\) corresponds to geometric primitives, and the set of edges \(E = \{e_{ij}\}\) encodes constraints between them. While constraints play a crucial role in defining design intent, they introduce additional dependencies that significantly complicate generative modeling. In this work, we focus on synthesizing the geometric primitives alone, deferring constraints to future work. Some primitives may be designated as construction aids; such primitives are not directly rendered in the final 3D CAD model but assist in specifying more complex design requirements (dashed lines in Fig.~\ref{fig:teaser}).

Each primitive \(v \in V\) is characterized by three attributes: (1) a boolean \(b\) indicating whether it is a construction aid, (2) a class label \(c \in \mathcal{C}\) specifying its type where \(\mathcal{C}=\{\text{\textsc{Line}, \textsc{Circle}, \textsc{Arc}, \textsc{Point}}\}\), and (3) a set of parameters \(p \in \mathbb{R}^{d_c}\), where \(d_c\) depends on the primitive class. The parameters used to define each primitive type is as follows:  
\begin{align*}\label{sec:cad-prim}  
\textsc{Circle}: &\quad (x, y, r) & (\text{center coords, radius})\\  
\textsc{Line}: &\quad (x_1, y_1, x_2, y_2) & (\text{start/end coords}) \\  
\textsc{Arc}: &\quad (x_1, y_1, x_2, y_2, \kappa) & (\text{start/end coords, radius}) \\  
\textsc{Point}: &\quad (x, y) & (\text{xy-coordinates})  
\end{align*}
We note that $\kappa$ is the curvature or signed radius, negating the curvature will reflect the \textsc{Arc} across the line defined by its starting and ending terminal points.

We resolve the heterogeneity of primitive parameterizations by modeling each primitive as a composite structure that encodes all possible primitive types. Specifically, we represent each primitive as:
\[
(\mathbf{b}, \mathbf{c}, \mathbf{p}_{\textsc{line}}, \mathbf{p}_{\textsc{circle}}, \mathbf{p}_{\textsc{arc}}, \mathbf{p}_{\textsc{point}})
\]  
where \(\mathbf{b}\) is a one-hot encoding of the construction aid flag, \(\textbf{c} \in \{0,1\}^{|\mathcal{C}|+1}\) is a one-hot encoding of the class label with an additional \textsc{None} type, and \(\mathbf{p}_{\textsc{line}}, \mathbf{p}_{\textsc{circle}}, \mathbf{p}_{\textsc{arc}}, \mathbf{p}_{\textsc{point}}\) are the parameters for each primitive type. This representation enables SketchDNN to view each primitive as a probabilistic mixture, or superposition, of all primitive types. Thus, the full sketch representation is given by a matrix \(X \in \mathbb{R}^{n \times d}\), where \(n\) is the maximum number of primitives, and \(d\) is the feature dimension. For this work \(n=16\) and \(d=20\). To recover the standard representation from our composite encoding, we extract the class label with the highest confidence (i.e., $c=\text{argmax}\left\{\mathbf{c}\right\}$) along with its corresponding parameters.

% \begin{figure}[h]
% \centering
% \includegraphics[width=\columnwidth]{Figures/CAD Graph Example.pdf}
% \caption{A rendering of a CAD sketch where every primitive is assigned a unique color and primitives tagged as construction aids are dotted on the left. The corresponding graph representation is provided on the right where cyan represents coincidence, red a midpoint constraint, and purple a parallel constraint. Construction aids are omitted from the graph for clarity and readability.}
% \label{fig:cad sketch}
% \end{figure}

\section{Methodology}\label{sec:diff}
In this work, we depart from the conventional diffusion parameterization of predicting noise, and instead parameterize our model to predict the ground truth datapoint directly. We observed in our preliminary experiments that this approach yielded higher fidelity samples for CAD diagram generation than predicting the noise added. This finding parallels the results of \cite{ho2020denoising}, who demonstrated that predicting noise led to superior sample quality in the context of image generation, and so it appears the converse is true for CAD sketch generation. For completeness, and because the foundational principles of diffusion models extend naturally to our proposed Gaussian-Softmax diffusion paradigm, we include a concise review of Gaussian diffusion formulated in terms of the ground truth parameterization. 

\subsection{Continuous Diffusion}
Diffusion models, first introduced by \cite{sohlthermo} and later popularized by \cite{ho2020denoising}, represent a class of generative models that learn to undo a gradual noising process. The forward noising process is a Markov chain that gradually adds Gaussian noise to a clean datapoint $x_0$ over a series of \( T \) timesteps, so that \( \mathbf{x}_T \) is indistinguishable from pure noise. The forward transition is defined as:  
\begin{equation}
 \mathbf{x}_t = \sqrt{\alpha_t} \mathbf{x}_{t-1} + \sqrt{\left(1 - \alpha_t\right)} \boldsymbol{\epsilon}
\end{equation}  
where \( \boldsymbol{\epsilon} \sim \mathcal{N}\left(0, \mathbf{I}\right) \) and \( \alpha_0, \alpha_1, \ldots, \alpha_T \) is a noise schedule that controls the amount of noise added at each timestep. A noise schedule is a sequence of monotonically decreasing scalar values satisfying \( \alpha_0 = 1 \) and \( \alpha_T = 0 \). We use the cosine variance schedule introduced by Nichol \& Dhariwal \cite{nichol2021} for this work.

Conveniently, iterative compositions of the forward transition can be expressed in closed form as the cumulative transition:  
\begin{equation}
\label{eqn:continous cumulative transition}  
\mathbf{x}_t = \sqrt{\overline{\alpha_t}} \mathbf{x}_{0} + \sqrt{\left(1 - \overline{\alpha_t}\right)} \boldsymbol{\epsilon}
\end{equation}  
where \( \overline{\alpha_t} = \prod_{i=1}^t \alpha_i \). This allows for efficient training, as we can in one shot sample \( \mathbf{x}_t \) from the original datapoint $\mathbf{x}_0$ for any arbitrary \( t \in \{0,\ldots,T\} \). The reverse transition is then given by:  
\begin{equation}
\label{eqn:continous reverse transition}
\mathbf{x}_{t-1} = \boldsymbol\mu_{t-1} + \sigma_{t-1} \boldsymbol{\epsilon}
\end{equation}  
where
$$
\boldsymbol\mu_{t-1}=\frac{\sqrt{\alpha_t}\left(1 - \overline{\alpha}_{t-1}\right) \mathbf{x}_t + \sqrt{\overline{\alpha}_{t-1}}\left(1 - \alpha_t\right) \mathbf{x}_0}{1 - \overline{\alpha}_t}
$$
and
$$\sigma_{t-1}=\sqrt{\frac{\left(1 - \alpha_t\right)\left(1 - \overline{\alpha}_{t-1}\right)}{1 - \overline{\alpha}_t}}$$

We note that $\boldsymbol\mu_{t-1}$ is simply a function of $\mathbf{x}_t$ and $\mathbf{x}_0$, so in more explicit terms $\boldsymbol\mu_{t-1}:=\boldsymbol\mu_{t-1}\left(\mathbf{x}_t,\mathbf{x}_0\right)$. The reverse process is learned by training a denoiser network to predict the clean data point given a noisy sample. The denoiser network can be expressed as \( \hat{\mathbf{x}}^\theta_0\left(\mathbf{x}_t, t\right) \), and the learned reverse transition is given by:
$$
\mathbf{x}_{t-1} = \boldsymbol\mu_{t-1}(\mathbf{x}_t, \hat{\mathbf{x}}^\theta_0\left(\mathbf{x}_t, t\right)) + \sigma_{t-1} \boldsymbol{\epsilon}
$$ 
where novel samples are generated from pure noise $\mathbf{x}_T$ by iteratively applying the learned reverse transition from $t=T$ to $t=0$.

\subsection{Discrete Diffusion}\label{sec:disc-diff}

Building on the success of continuous diffusion models in domains such as image generation, \citet{hoogeboom2021argmax} introduced a discrete diffusion framework utilizing the Categorical distribution, that was later expanded by \cite{austin2021structured}. In Multinomial diffusion, the forward process is a discrete Markov chain that jumbles the state of a one-hot vector $\mathbf{y}_0$ over $T$ steps. The forward transition is defined to be 
$$
\mathbf{y}_t\sim\text{Cat}(\mathbf{y}_{t-1}Q_t)
$$ 
where $Q_t$ is a probability matrix. In other words, the forward transition stochastically permutes/shuffles the index of the $1$ in a one-hot vector. As a result, the output of the forward transition is always a one-hot vector, and similarly the cumulative transition and reverse transition also output one-hot vectors. This naturally renders superposition impossible in conventional discrete diffusion, because there is no way to accommodate blended class labels to express uncertainty. This is detrimental because there is no gradual destruction of information, once a class label is shuffled all information is destroyed. Preliminary experiments with generating CAD sketches using Multinomial diffusion yielded poor results, we believe the root cause to be the inability to accommodate superposition. This is validated by our results in Section \ref{sec:results}.

\subsection{Gaussian-Softmax Diffusion}
To overcome the limitations of Multinomial diffusion, we introduce the Gaussian-Softmax ($\mathcal{GS}$) distribution as a continuous relaxation of the Categorical distribution. We propose a novel simplex-constrained discrete diffusion framework utilizing the Gaussian-Softmax distribution to enable superposition, allowing information to be gradually destroyed in the forward process and gradually recovered in the reverse process. Essentially, if $\mathbf{x} \sim \mathcal{N}(\boldsymbol{\mu}, \sigma^2 \mathbf{I})$, then $\softmax{}(\mathbf{x}) \sim \mathcal{GS}(\boldsymbol{\mu}, \sigma^2 \mathbf{I})$ where $\softmax$ is the softmax transformation. The softmax function transforms arbitrary vectors $\mathbf{x}$ into probability vectors $\mathbf{p}$, satisfying $\sum_i \mathbf{p}_i = 1$ and $\mathbf{p}_i \geq 0, \forall i$, which geometrically means that the softmax operation maps vectors in $\mathbb{R}^D$ onto the probability simplex $\Delta^{D-1}$. Thus the support of the Gaussian-Softmax distribution is the probability simplex, and the density (derived in Appendix \ref{gauss-soft-density}) is:
\begin{equation}
    p(\mathbf{y}|\boldsymbol{\mu},\sigma^2\mathbf{I}) = \frac{\prod^D_{i=1} \mathbf{y}_i^{-1}}{Z(\sigma)} 
    \exp\left(-\frac{1}{2\sigma^2} \|\mathbf{\tilde{y}} - \boldsymbol{\mu}'\|^2_{\perp} \right)
\end{equation}
where
$$
Z(\sigma) = \sqrt{D(2\pi\sigma^2)^{(D-1)}}
$$
$$
\|\mathbf{\tilde{y}} - \boldsymbol{\mu}'\|^2_{\perp} = \|\mathbf{\tilde{y}} - \boldsymbol{\mu}'\|^2 - \frac{1}{D} (\mathbf{1}^T (\mathbf{\tilde{y}} - \boldsymbol{\mu}'))^2
$$
Here, \(\boldsymbol{\mu}' = \boldsymbol{\mu} - (\boldsymbol{\mu}_D)\mathbf{1}\), where every element in $\boldsymbol{\mu}$ is shifted by its last element, and similarly for \(\tilde{\mathbf{y}} = \log \mathbf{y} - (\log \mathbf{y}_D) \mathbf{1}\), every element in $\log \mathbf{y}$ is shifted by its last element.

\begin{figure*}[t!]
\begin{center}
\includegraphics[width=0.49\linewidth]{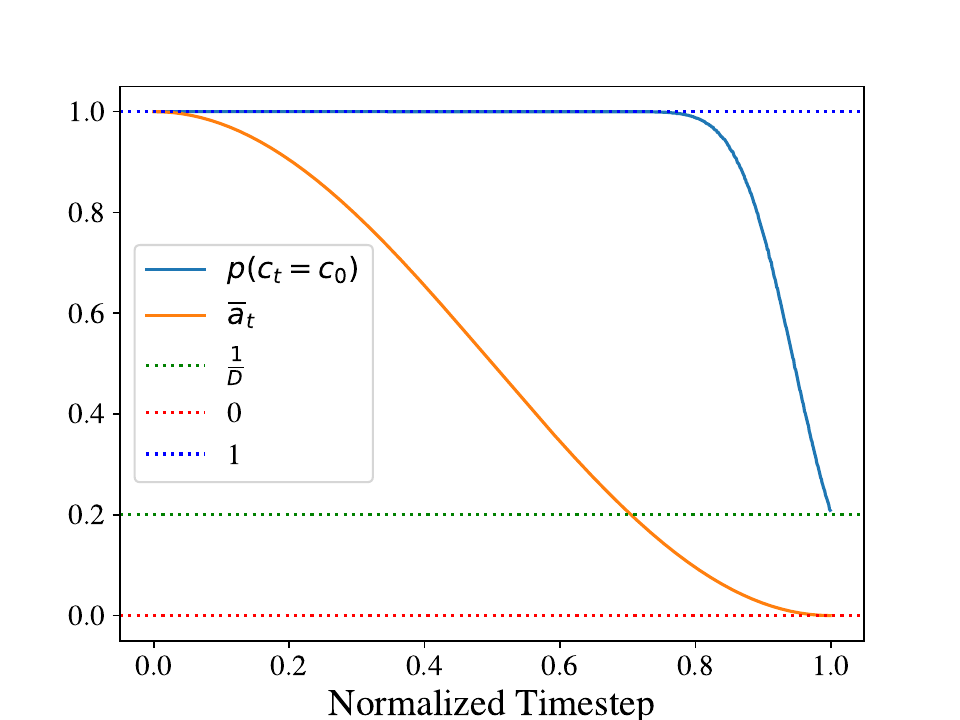}
\includegraphics[width=0.49\linewidth]{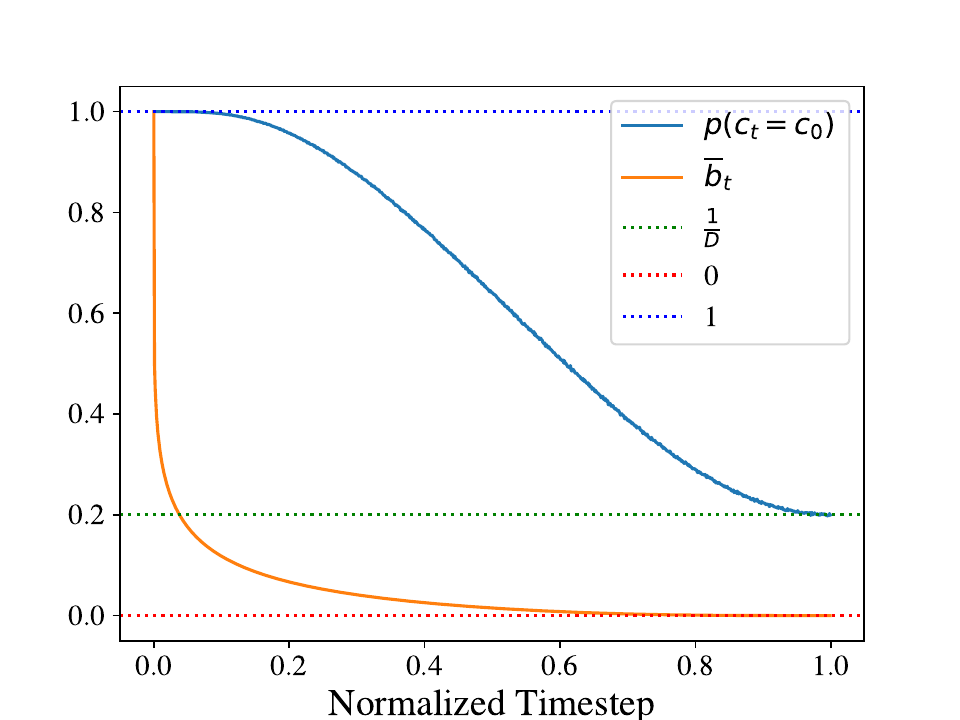}
\caption{\textbf{Left:} The orange curve represents the raw cosine variance schedule $\overline{a_t}$, while the blue curve depicts the probability that the class label remains unchanged. \textbf{Right:} The orange curve shows the augmented variance schedule $\overline{b_t}$, as defined in \cref{eqn:variance modification}, and the blue curve again represents the probability of the class label not switching. The probabilities were calculated using Monte Carlo estimation over 100,000 samples. The augmented variance schedule results in a more gradual discrete forward process compared to the raw variance schedule.}
\label{fig:variance-aug}
\end{center}
\end{figure*}

\subsubsection{Forward Process}

% Analogous to continuous diffusion, the forward noising process is a Markov chain that stochastically smooths the label of a one-hot vector \( \mathbf{y}_0 \) to produce a sequence of increasingly noisy probability vectors \( \mathbf{y}_1, \mathbf{y}_2, \ldots, \mathbf{y}_T \). The class label of a noisy vector $\mathbf{y}_t$ is just the class with the highest confidence, so $c_t=\text{argmax}\left\{ \mathbf{y}_t \right\}$. Thus, at $t=0$ the class label should have no noise meaning $c_0 \sim \mathcal{C}\left(\mathbf{y}_0\right)$ and at $t=T$ the class label should be pure noise meaning $c_T \sim \mathcal{C}\left(\frac{\mathbf{1}}{D}\right)$. Here $\mathcal{C}(\mathbf{p})$ is the Categorical distribution, so at $t=0$ is the noiseless class label and at $t=T$ the class label follows the Uniform distribution meaning all class labels are equally likely. 
Analogous to continuous diffusion, the forward process is a Markov chain that progressively noises a one-hot vector \( \mathbf{y}_0 \) via $\mathcal{GS}$ noise over $T$ steps, generating a sequence of increasingly noisy probability vectors \( \mathbf{y}_1, \mathbf{y}_2, \dots, \mathbf{y}_T \). As noise accumulates, information about the true label is destroyed, and by \( t = T \), the class label becomes fully randomized where $\mathbf{y}_T \sim \mathcal{GS}(0,\mathbf{I})$. To achieve this, we define the forward transition as  
\begin{equation}
\label{eqn:discrete_forward_transition}
\mathbf{y}_{t+1} = \softmax\left(\sqrt{\alpha_{t+1}} \log \mathbf{y}_t + \sqrt{1 - \alpha_{t+1}} \boldsymbol{\epsilon} \right)
\end{equation}
where $\boldsymbol{\epsilon} \sim \mathcal{N}(\mathbf{0}, \mathbf{I})$. Additionally, we can conveniently sample $\mathbf{y}_t$ directly from the one-hot vector $\mathbf{y}_0$ using the cumulative transition (derivation in Appendix \ref{cumulative_forward}):  
\begin{equation}
\label{eqn:discrete_cumulative_transition}
\mathbf{y}_t = \softmax\left(\sqrt{\overline{\alpha}_t} \log \mathbf{y}'_0 + \sqrt{1 - \overline{\alpha}_t} \boldsymbol{\epsilon} \right)
\end{equation}
here $\mathbf{y}'_0=k\mathbf{y}_0+\frac{1-k}{D}\mathbf{1}$ and $k$ is an user-defined constant close to 1, we set $k=0.99$. We slightly label smooth $\mathbf{x}_0$ to avoid singularities from computing the logarithm of 0. We remark that at $t=T$ the cumulative transition simply takes the softmax of an i.i.d standard Gaussian vector, which by symmetry makes the argmax or label $c_T$ follow a uniform distribution, satisfying $c_T \sim \mathcal{C}\left(\frac{\mathbf{1}}{D}\right)$.

% \subsection{Reverse Process}
% Notably, we find that, analogous to the continuous reverse transition, the discrete reverse transition is of the form:  
% \begin{equation}
% \label{eqn:discrete_reverse_transition}
% \mathbf{x}_{t-1} = S\left(\mu_{t-1}(\mathbf{y}_t, \mathbf{y}_0(\mathbf{y}_t, t)) + \sigma_{t-1}\boldsymbol{\epsilon} \right),
% \end{equation}
% Since the only unknown for the reverse transition is the noiseless one-hot vector $\mathbf{y}_0$, the reverse process can be learned by training a denoiser network to learn $\mathbf{y}_0\left(\mathbf{y}_t,t\right)$. The mean term is simply an interpolation between the clean one-hot vector and a noisy probability vector where:
% \begin{equation}
% \mu_{t-1}(\mathbf{y}_t, \mathbf{y}_0) = \frac{\sqrt{a_t} (1 - \overline{a}_{t-1}) \mathbf{y}'_t + \sqrt{\overline{a}_{t-1}} (1 - a_t) \mathbf{y}'_0}{1 - \overline{a}_t}.
% \end{equation}
% for which $\mathbf{y}'_t=\log\mathbf{y}_t$ and $\mathbf{y}'_0=\log\mathbf{y}_0$ are just log-space transformations. The standard deviation is similarly:
% \begin{equation}
% \sigma_{t-1}=\sqrt{\frac{(1 - a_t)(1 - \overline{a}_{t-1})}{1 - \overline{a}_t}}
% \end{equation}

% We provide a full derivation of the reverse transition in \ref{}. Intuitively, for both the forward and reverse process, continuous diffusion is performed in log-space before projecting the result onto the probability simplex using the softmax function. Similarly, novel samples can be generated by iteratively applying the reverse transition from $t=T$ to $t=0$.

\subsubsection{Reverse Process}  
Notably, we find that the discrete reverse transition is analogous to the continuous case and takes the form:  
\begin{equation}
\label{eqn:discrete_reverse_transition}
\mathbf{y}_{t-1} = \softmax\left(\boldsymbol\mu_{t-1}(\mathbf{y}_t, \mathbf{y}_0) + \sigma_{t-1}\boldsymbol{\epsilon} \right)
\end{equation}  
Here, the mean \(\boldsymbol\mu_{t-1}\) is simply an interpolation between the logits of \(\mathbf{y}_0\) and the logits of \(\mathbf{y}_t\):  
$$
\mu_{t-1}(\mathbf{y}_t, \mathbf{y}_0) = \frac{\sqrt{\alpha_t} (1 - \overline{\alpha}_{t-1}) \mathbf{y}'_t + \sqrt{\overline{\alpha}_{t-1}} (1 - \alpha_t) \mathbf{y}'_0}{1 - \overline{\alpha}_t}
$$
where \(\mathbf{y}'_t = \log\mathbf{y}_t\) and \(\mathbf{y}'_0 = \log\mathbf{y}_0\). Again, The standard deviation is similarly given as:  
$$
\sigma_{t-1} = \sqrt{\frac{(1 - \alpha_t)(1 - \overline{\alpha}_{t-1})}{1 - \overline{\alpha}_t}}
$$
A proof of the reverse process is provided in Appendix \ref{reverse-derivation}.

Intuitively, we perform continuous diffusion in log-space and project back onto the probability simplex using the softmax function. The reverse transition is learned by training a denoiser network to predict the ground-truth label from a noisy version, expressed as \(\hat{\mathbf{y}}^\theta_0\left(\mathbf{y}_t, t\right)\) where: 
$$
\mathbf{y}_{t-1} = \softmax\left(\mu_{t-1}\left(\mathbf{y}_t, \hat{\mathbf{y}}^\theta_0\left(\mathbf{y}_t, t\right)\right) + \sigma_{t-1}\boldsymbol{\epsilon} \right)
$$
Novel samples can be generated by iteratively applying the learned reverse transition from \(t = T\) to \(t = 0\).

% \subsection{Variance Schedule Augmentation}
% For Gaussian-Softmax diffusion, we found that variance schedules can not be used directly, since the softmax projection distorts the effect of the injected noise. To address this, we propose the following augmentation such that
% $
% \text{argmax}(\mathbf{y}_t)\sim\mathcal{C}(\overline{a_t} \mathbf{y}_0 + (1-\overline{a_t})/D)
% $
% achieved via the augmentation:
% \begin{equation}
%     \overline{b_t} = \frac{f(\overline{a_t})^2}{f(\overline{a_t})^2 + f(k)^2}, \hspace{0.5em} f(x) = \log \left( \frac{1-x}{(D-1)x+1} \right)
%     \label{eqn:variance modification}
% \end{equation}
%  Figure \ref{fig:variance-aug} highlights the importance of the proposed variance schedule augmentation, demonstrating its advantages over directly using the raw variance schedule. We provide a derivation in \ref{}.

\subsection{Variance Schedule Augmentation}  
In Gaussian-Softmax diffusion, we observed that variance schedules cannot be used directly as-is due to the distortion introduced by the softmax projection on the injected noise. To address this, we propose an augmentation to the variance schedule, ensuring that the distribution of $\text{argmax}(\mathbf{y}_t) \mathrel{\dot\sim} \mathcal{C}\left(\overline{\alpha_t} \mathbf{y}_0 + (1-\overline{\alpha_t})/D\right)$. This is achieved through the following augmentation:  
\begin{equation}
    \overline{b_t} = \frac{f(\overline{\alpha_t})^2}{f(\overline{\alpha_t})^2 + f(k)^2}, \quad f(x) = \log \left( \frac{1-x}{(D-1)x + 1} \right)
    \label{eqn:variance modification}
\end{equation}  
Figure \ref{fig:variance-aug} illustrates the significance of this augmentation, showcasing its necessity over directly using the raw variance schedule. We also provide the derivation in Appendix \ref{variance-augmentation}.

\section{Sketch Diffusion} \label{sec:sketch diffusion}
Since we focus solely on generating primitives, we represent a CAD sketch \(X\) as a set of primitives \(X = \{x_i\}\), where \(X \in \mathbb{R}^{n \times d}\), \(n\) denotes the maximum number of primitives, and \(d\) represents the dimensionality of primitive features. Importantly, all \(n!\) permutations of \(X\) are equivalent in terms of geometry, since permuting \(X\) doesn't change the actual geometry of each primitive. This invariance poses a challenge for generative modeling, as the ordering of primitives must not influence the learned generative process.

To address this, we adopt a permutation-equivariant diffusion methodology, ensuring that the denoising procedure remains consistent irrespective of primitive ordering. To apply noise to a sketch, we perturb each primitive independently, such that the forward transition distribution and cumulative transition distribution, respectively, factorize as:  
\begin{align}
    p(X_{t+1} \mid X_t) &= \prod_{i=1}^{n} p(x_{t+1,i} \mid x_{t,i}) \\
    p(X_{t} \mid X_0) &= \prod_{i=1}^{n} p(x_{t,i} \mid x_{0,i})
\end{align}

Since each primitive undergoes the same noising process independently of the others, permuting the order of primitives in \(X\), before the forward process, results in an equivalent permutation of the noised sketch \(X_t\). That is, if \(\pi\) is any permutation of \(\{1, \dots, n\}\), then  
\(
p(\pi(X_t) \mid \pi(X_0)) = p(X_t \mid X_0)
\).
Thus, the forward process itself is permutation-equivariant.

Applying Bayes’ rule, the posterior distribution can be expressed as:  
\begin{align}
\begin{split}
p(X_{t-1} \mid X_t, X_0) &= \frac{p(X_{t}|X_{t-1})p(X_{t-1}|X_0)}{p(X_t|X_0)} \\
&= \prod_{i=1}^{n}\frac{p(x_{t,i}|x_{t-1,i})p(x_{t-1,i}|x_{0,i})}{p(x_{t,i}|x_{0,i})} \\
&= \prod_{i=1}^{n} p(x_{t-1,i} \mid x_{t,i}, x_{0,i})
\end{split}
\end{align}
Thus, using the same line of reasoning as for the forward process, we can see that the reverse process is also permutation equivariant. To ensure the learned reverse process maintains permutation equivariance, the denoiser network must not introduce positional dependencies. This can be facilitated by using an architecture that is inherently permutation-equivariant. In our case, we adopt the diffusion transformer architecture by \cite{peebles2023scalablediffusionmodelstransformers} and simply omit positional encodings, since transformers have been shown to be intrinsically permutation-equivariant \cite{vaswani_attention_2017, Xu_2024_CVPR}. A similar architecture is used in Brepgen by \cite{brepgen} for permutation invariant, instead of equivariant, generation of B-rep splines.

% \subsection{Primitive Diffusion}
% Primitives consist of both discrete and continuous variables, necessitating joint continuous-discrete diffusion. Let each primitive be denoted as \(x = (\mathbf{b}, \mathbf{c}, \mathbf{p})\), where \(\mathbf{b}\) is the construction label, \(\mathbf{c}\) is the class label, and \(\mathbf{p}\) represents the continuous parameters. We independently apply noise and denoise each component, more specifically, for the discrete variables \(\mathbf{b}\) and \(\mathbf{c}\) we use Equations \ref{eqn:discrete_cumulative_transition} for the forward process and \ref{eqn:discrete_reverse_transition} for the reverse process. For the parameters \(\mathbf{p}\) we employ continuous diffusion where we use \ref{eqn:continous cumulative transition} for the forward process and \ref{eqn:continous reverse transition} for the reverse process. Thus we independently perform Gaussian-Softmax diffusion for discrete variables, and standard Gaussain diffusion for continuous variables.

\subsection{Primitive Diffusion}  
Each primitive is characterized by both discrete and continuous variables, necessitating a diffusion process that simultaneously operates in both domains. Formally, each primitive is represented as \(x = (\mathbf{b}, \mathbf{c}, \mathbf{p})\), where \(\mathbf{b}\) denotes the construction label, \(\mathbf{c}\) the class label, and \(\mathbf{p}\) the continuous parameters.  

To model the diffusion process, we treat each component independently. The discrete variables \(\mathbf{b}\) and \(\mathbf{c}\) undergo a forward process governed by Equation~\ref{eqn:discrete_cumulative_transition} and a reverse process dictated by Equation~\ref{eqn:discrete_reverse_transition}. The continuous parameters \(\mathbf{p}\), in contrast, follow a standard diffusion process, where the forward and reverse transitions are defined by Equations~\ref{eqn:continous cumulative transition} and~\ref{eqn:continous reverse transition}, respectively. Essentially, we perform joint discrete-continuous diffusion by independently performing Gaussian-Softmax diffusion for discrete variables and standard Gaussian diffusion for continuous parameters.
% A graphical illustration of the generation process is provided in Figure \ref{fig:diff-gen}.

% ----------- Diffusion Generation Figure ------------
\begin{figure*}[t!]
\begin{center}
\includegraphics[width=0.9\linewidth]{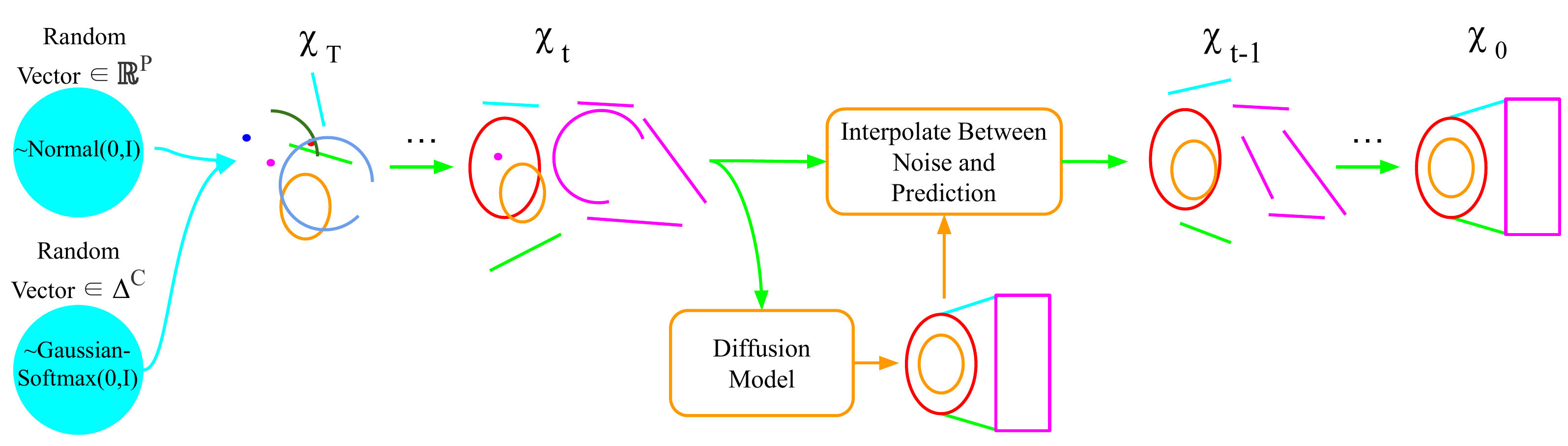}
\caption{The generation pipeline of SketchDNN. Starting from a pure noise seed \(\mathcal{X}_T\), the denoiser network iteratively refines the sample by interpolating between the noisy input and the model's prediction of the clean output. The final generated sketch, \(\mathcal{X}_0\), is obtained after $T$ successive denoising steps. The interpolation formulae are given by Equation~\ref{eqn:discrete_reverse_transition} for discrete variables and Equation~\ref{eqn:continous reverse transition} for continuous variables.}
\label{fig:diff-gen}
\end{center}
\end{figure*}

\section{Experiments}\label{sec:experiments}
\begin{figure*}[ht]
    \centering

    % Top: SketchGraphs Dataset
    \begin{tabular}{cccccccc}
        \includegraphics[clip, trim=1.5cm 1.5cm 1.5cm 1.5cm, width=0.13\textwidth, height=0.08\textheight]{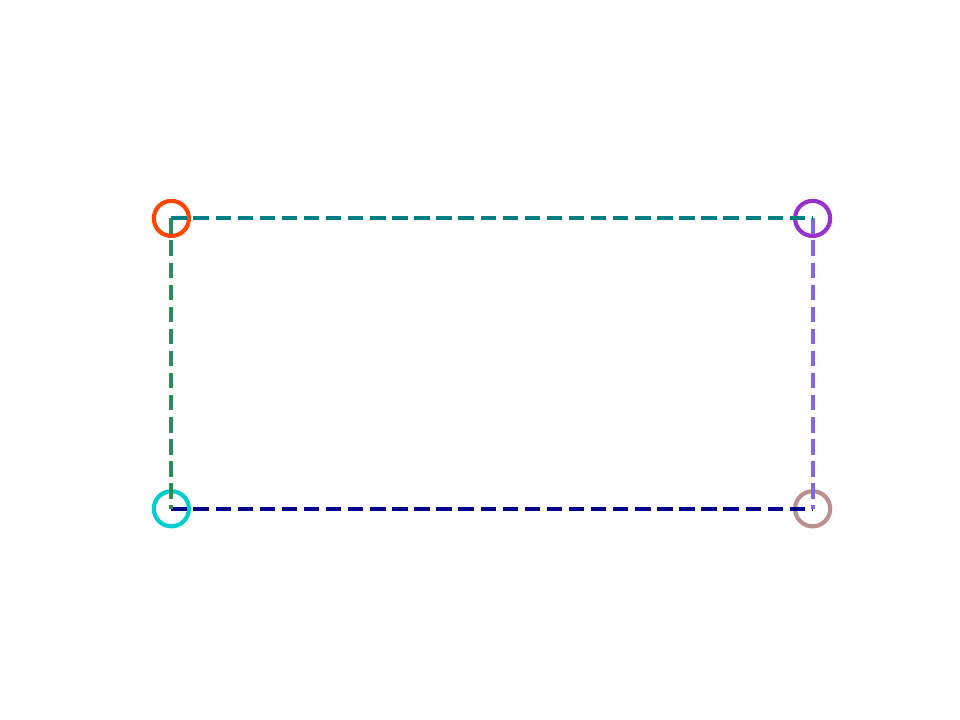} &
        \includegraphics[clip, trim=1.5cm 1.5cm 1.5cm 1.5cm, width=0.09\textwidth, height=0.05\textheight]{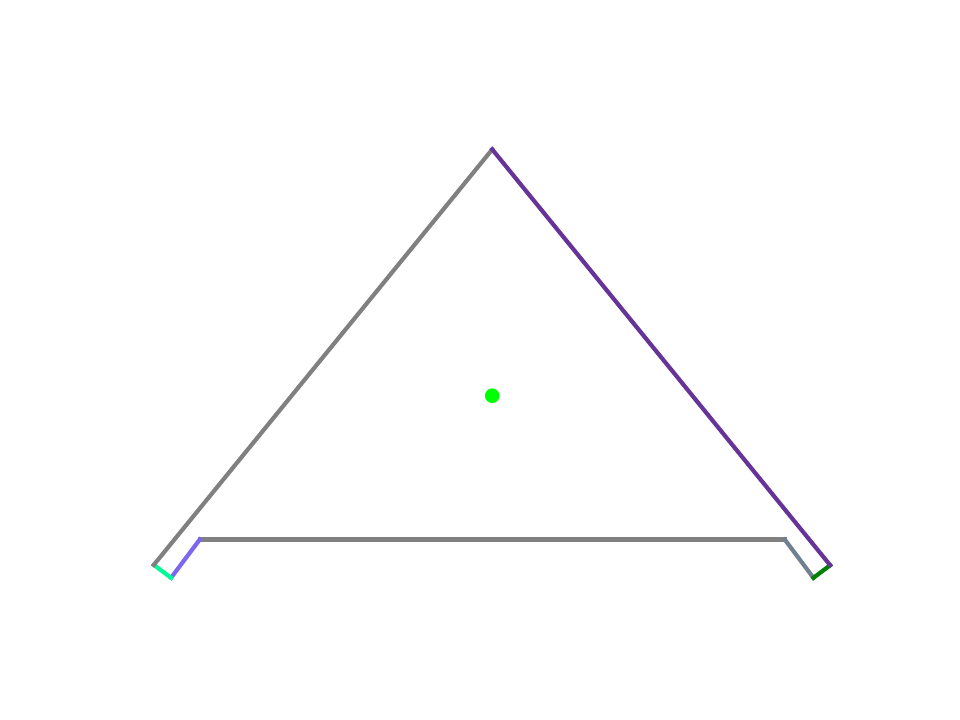} &
        \includegraphics[clip, trim=1.5cm 1.5cm 1.5cm 1.5cm, width=0.09\textwidth, height=0.05\textheight]{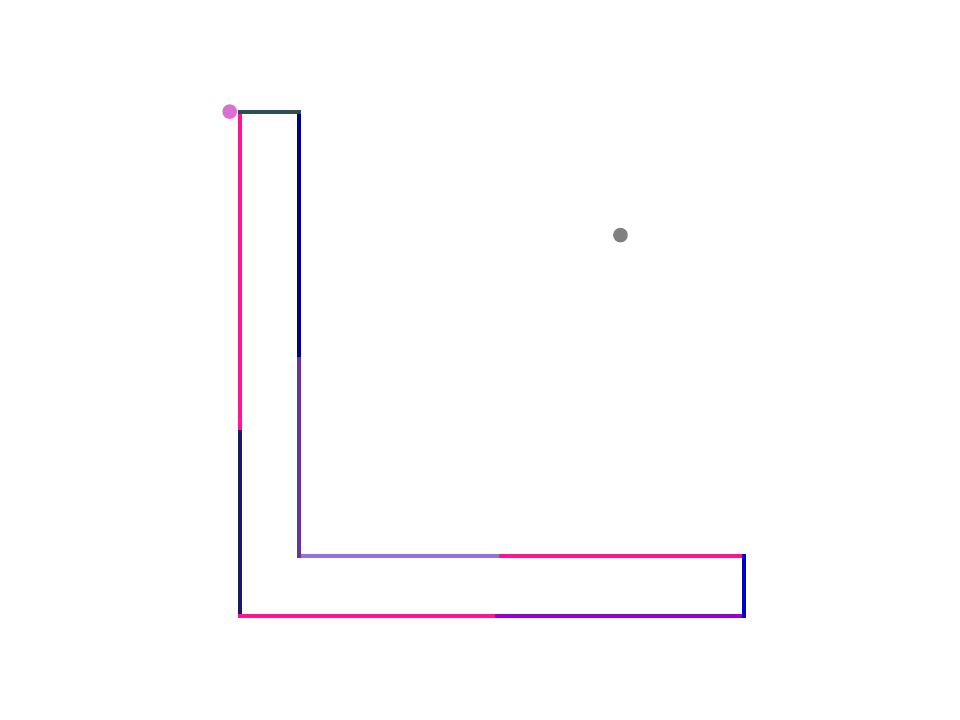} &
        \includegraphics[clip, trim=1.5cm 1.5cm 1.5cm 1.5cm, width=0.09\textwidth, height=0.05\textheight]{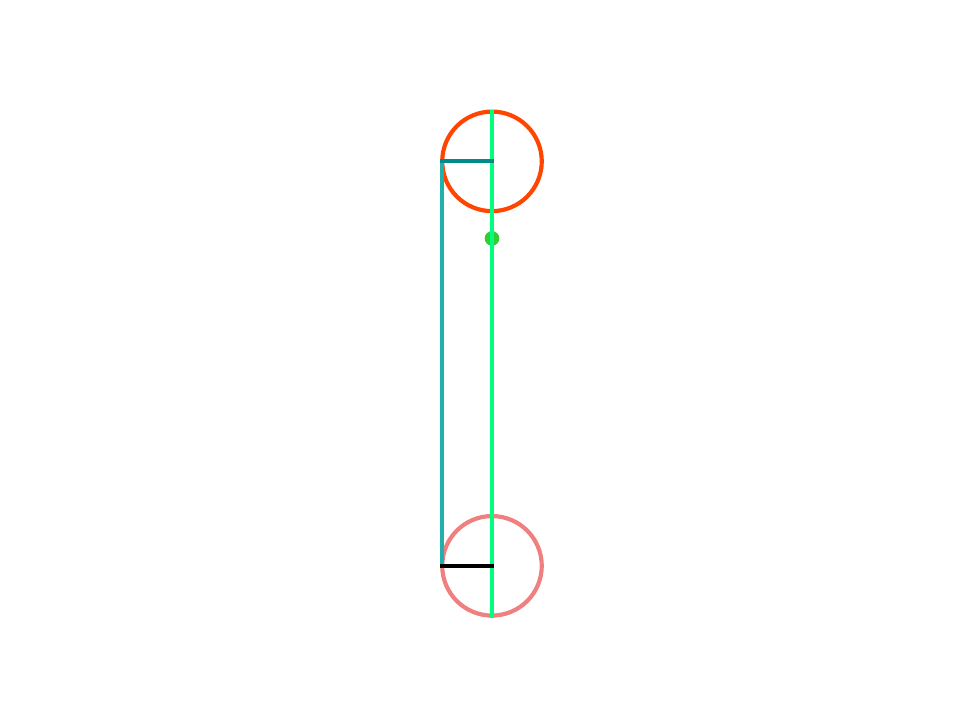} &
        \includegraphics[clip, trim=1.5cm 1.5cm 1.5cm 1.5cm, width=0.09\textwidth, height=0.05\textheight]{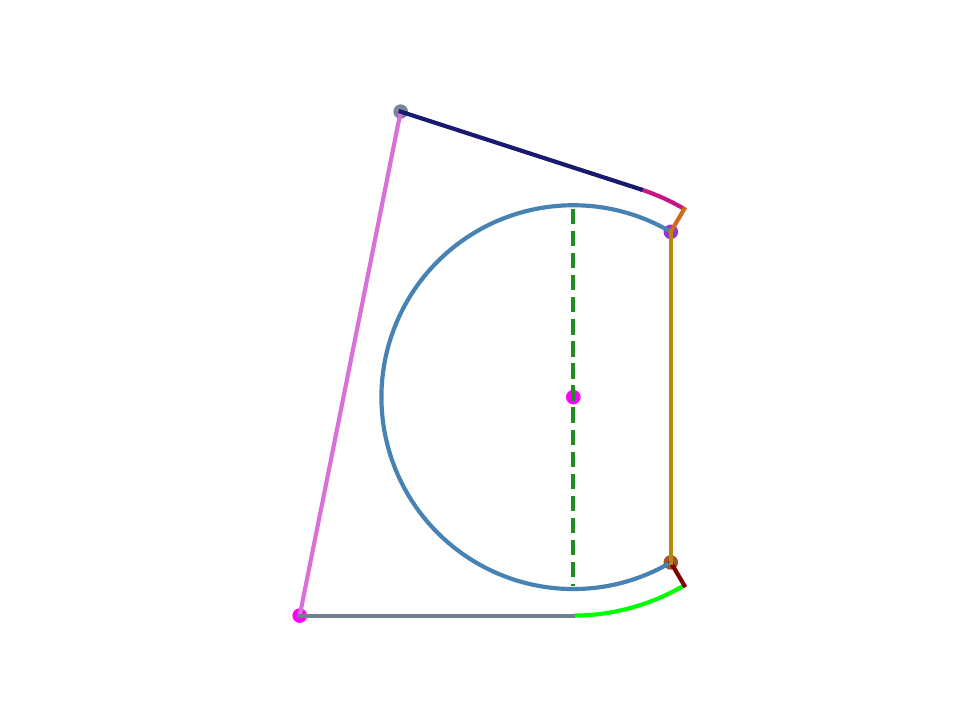} &
        \includegraphics[clip, trim=1.5cm 1.5cm 1.5cm 1.5cm, width=0.09\textwidth, height=0.05\textheight]{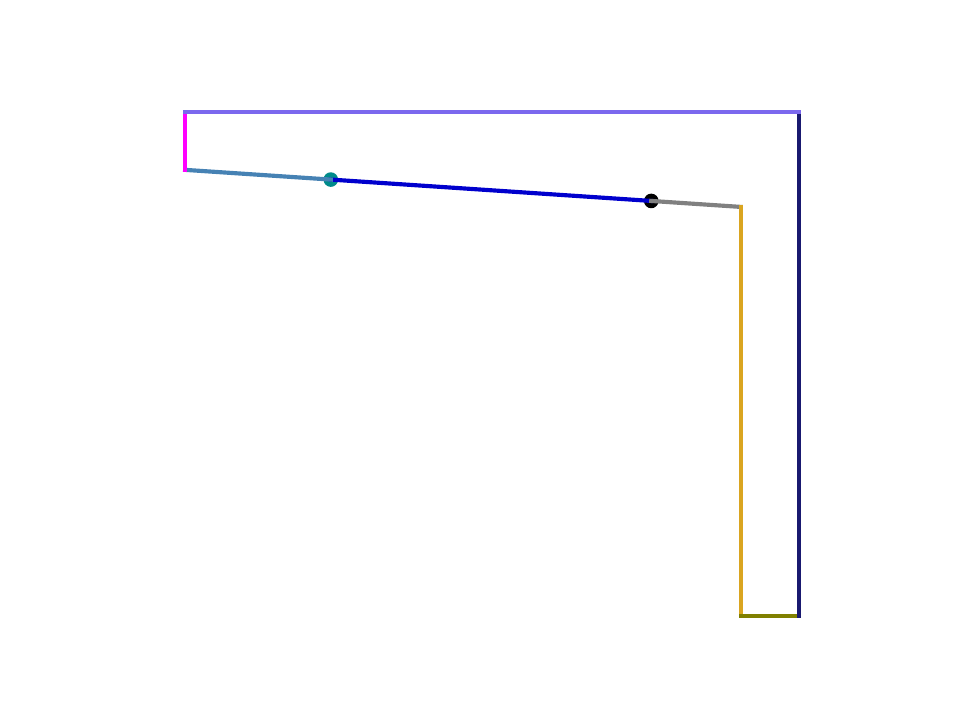} &
        \includegraphics[clip, trim=1.5cm 1.5cm 1.5cm 1.5cm, width=0.09\textwidth, height=0.05\textheight]{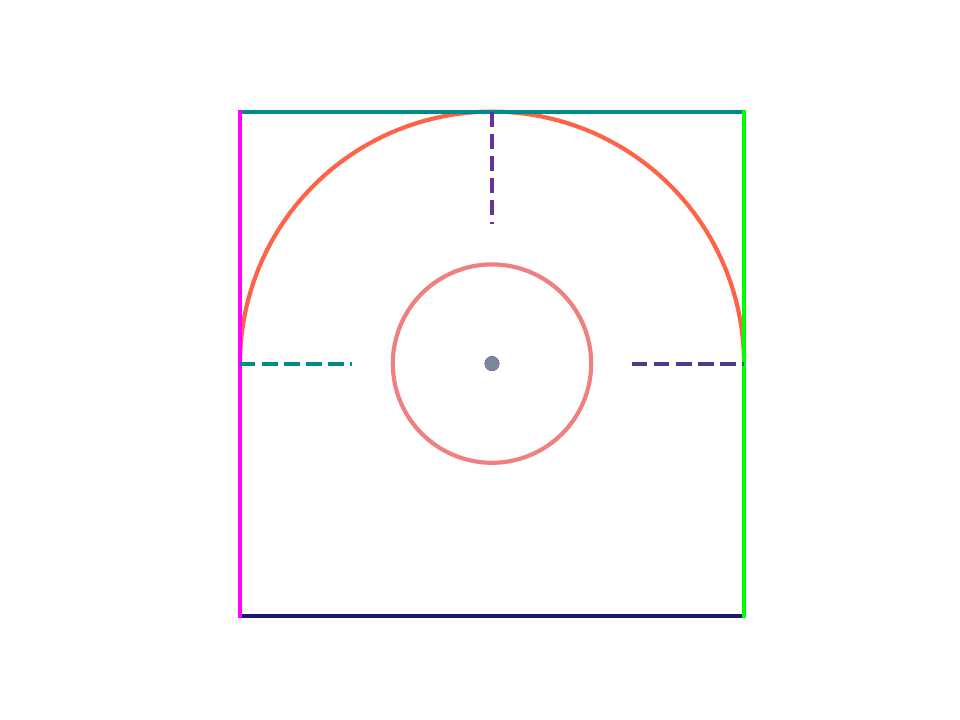} &
        \includegraphics[clip, trim=1.5cm 1.5cm 1.5cm 1.5cm, width=0.09\textwidth, height=0.05\textheight]{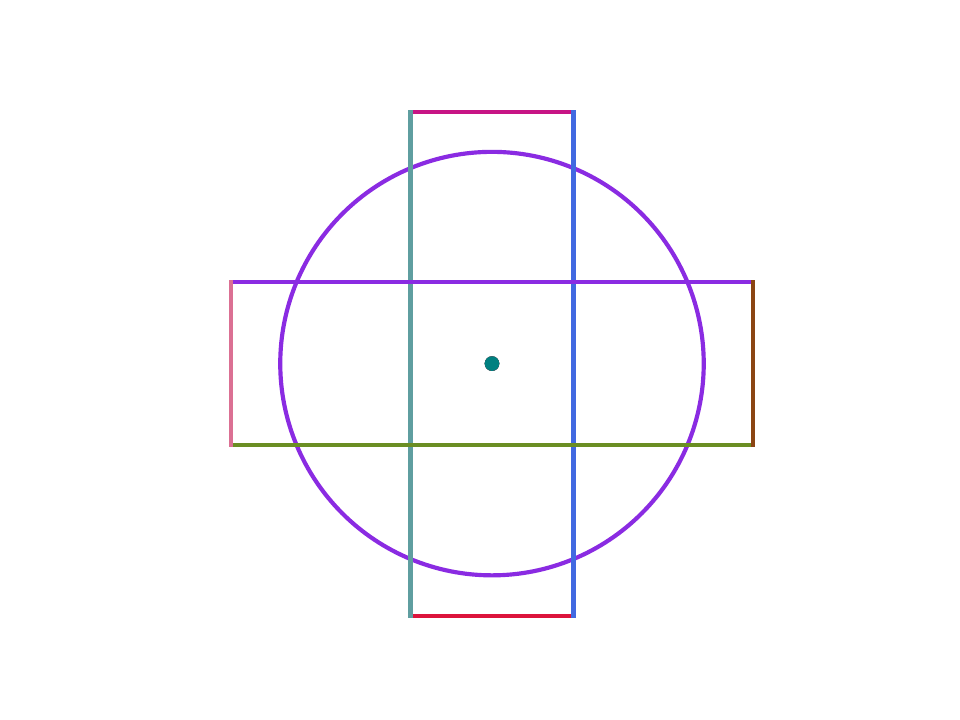} \\

        \includegraphics[clip, trim=1.5cm 1.5cm 1.5cm 1.5cm, width=0.09\textwidth, height=0.05\textheight]{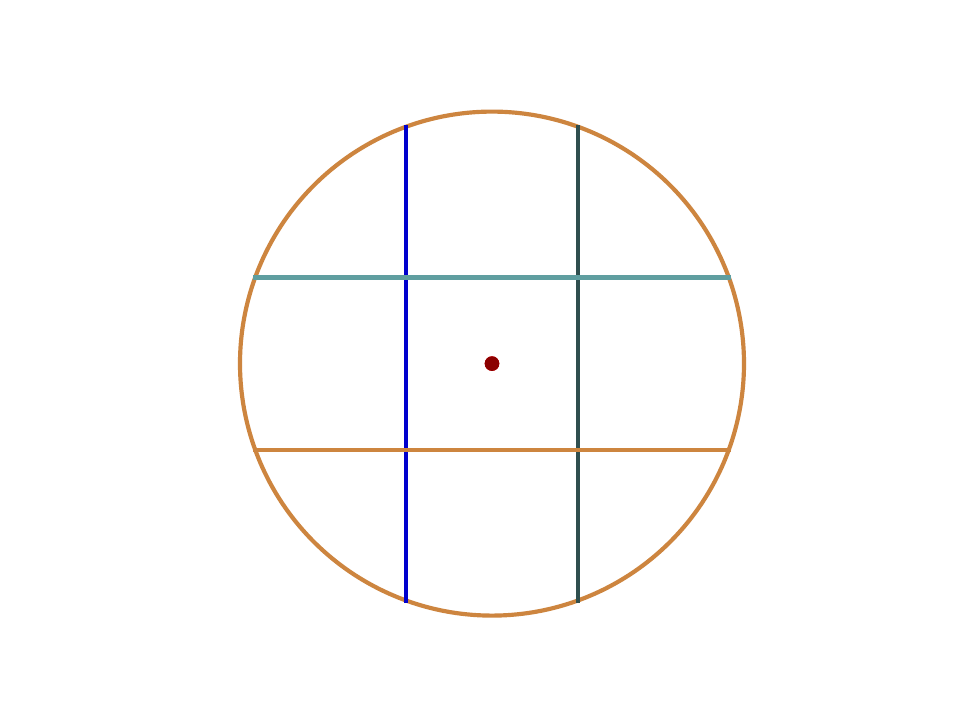} &
        \includegraphics[clip, trim=1.5cm 1.5cm 1.5cm 1.5cm, width=0.09\textwidth, height=0.05\textheight]{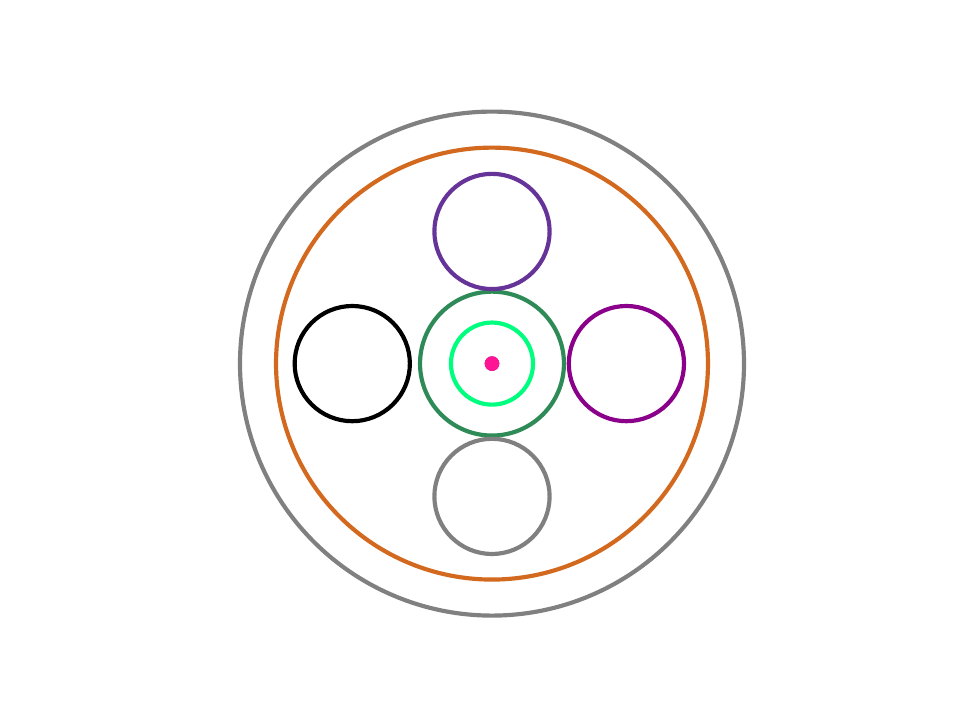} &
        \includegraphics[clip, trim=1.5cm 1.5cm 1.5cm 1.5cm, width=0.09\textwidth, height=0.05\textheight]{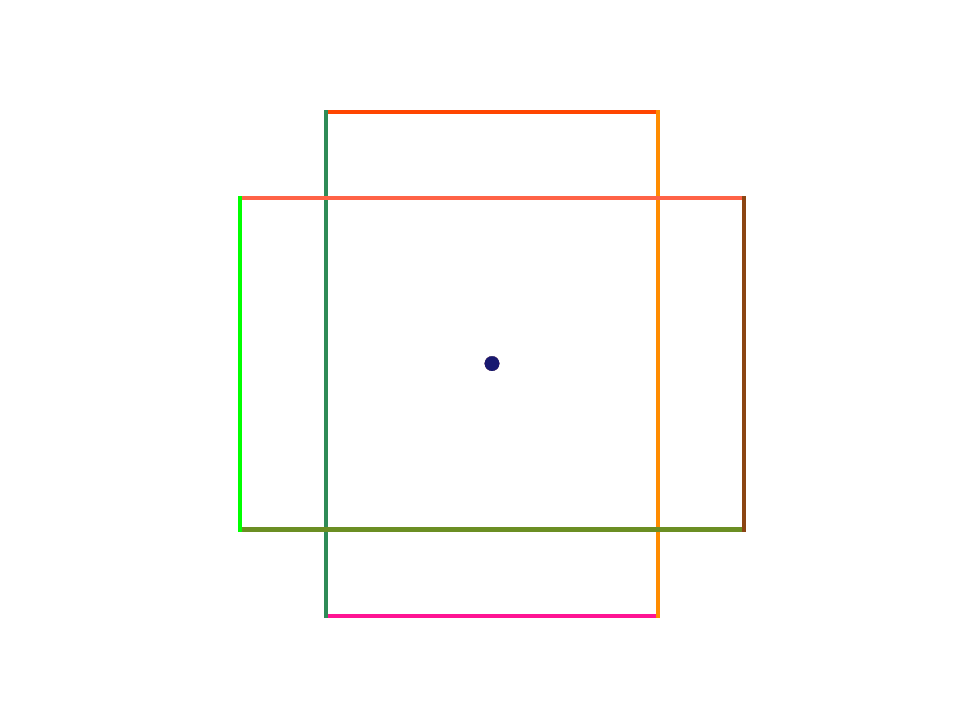} &
        \includegraphics[clip, trim=1.5cm 1.5cm 1.5cm 1.5cm, width=0.09\textwidth, height=0.05\textheight]{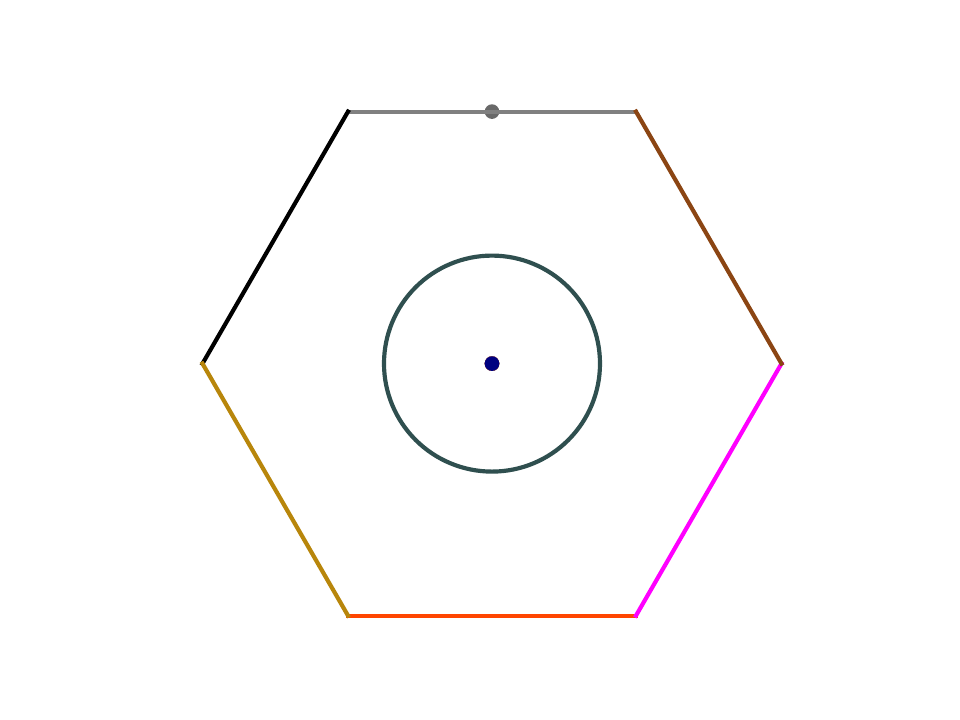} &
        \includegraphics[clip, trim=1.5cm 1.5cm 1.5cm 1.5cm, width=0.09\textwidth, height=0.05\textheight]{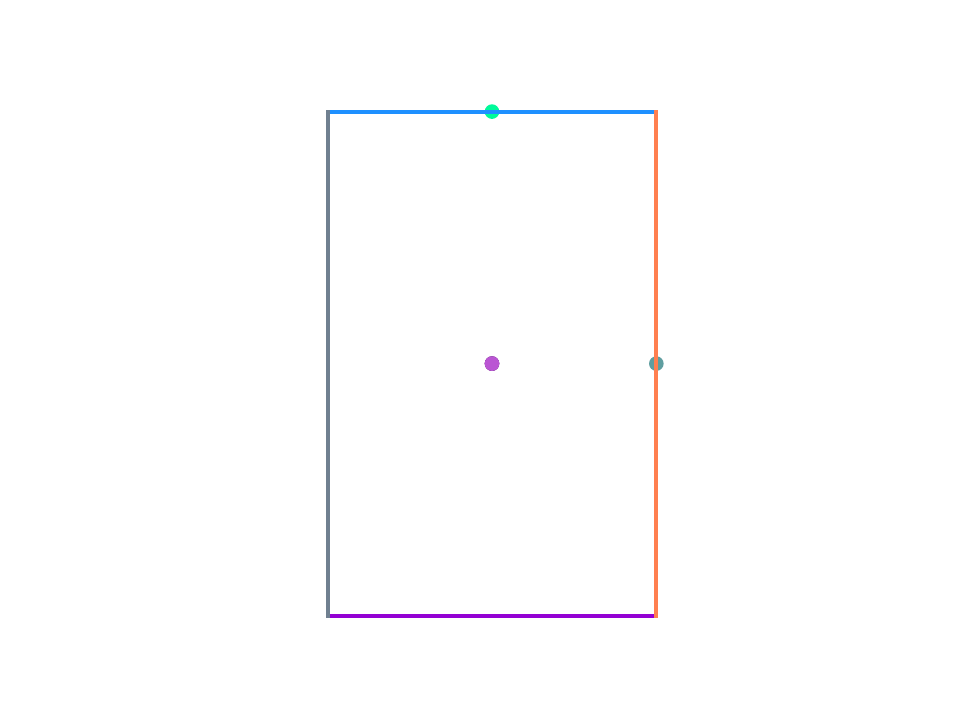} &
        \includegraphics[clip, trim=1.5cm 1.5cm 1.5cm 1.5cm, width=0.09\textwidth, height=0.05\textheight]{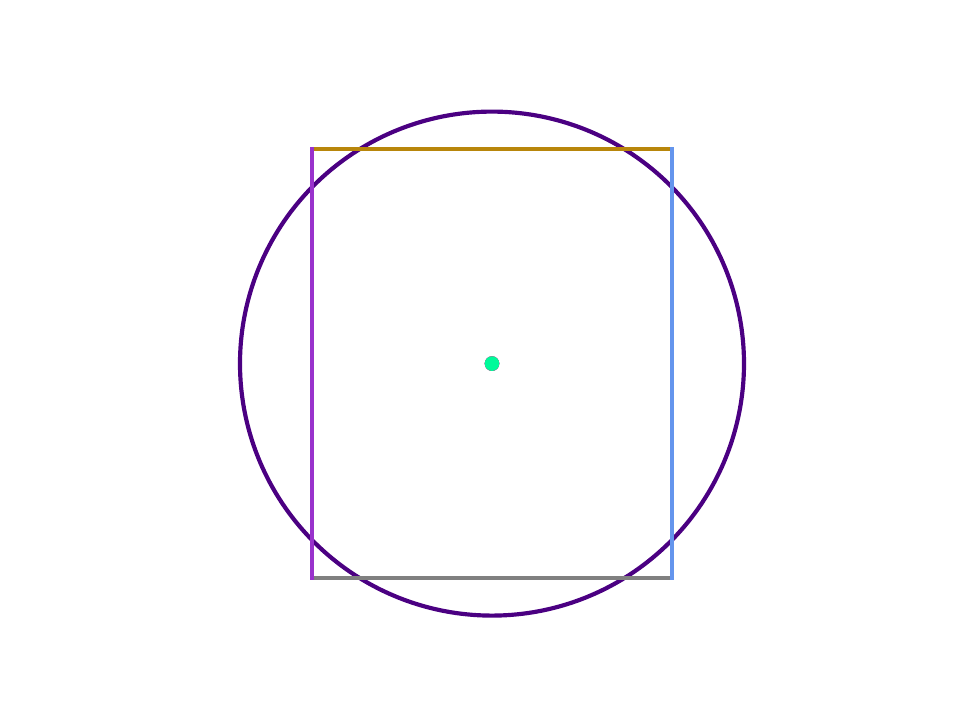} &
        \includegraphics[clip, trim=1.5cm 1.5cm 1.5cm 1.5cm, width=0.09\textwidth, height=0.05\textheight]{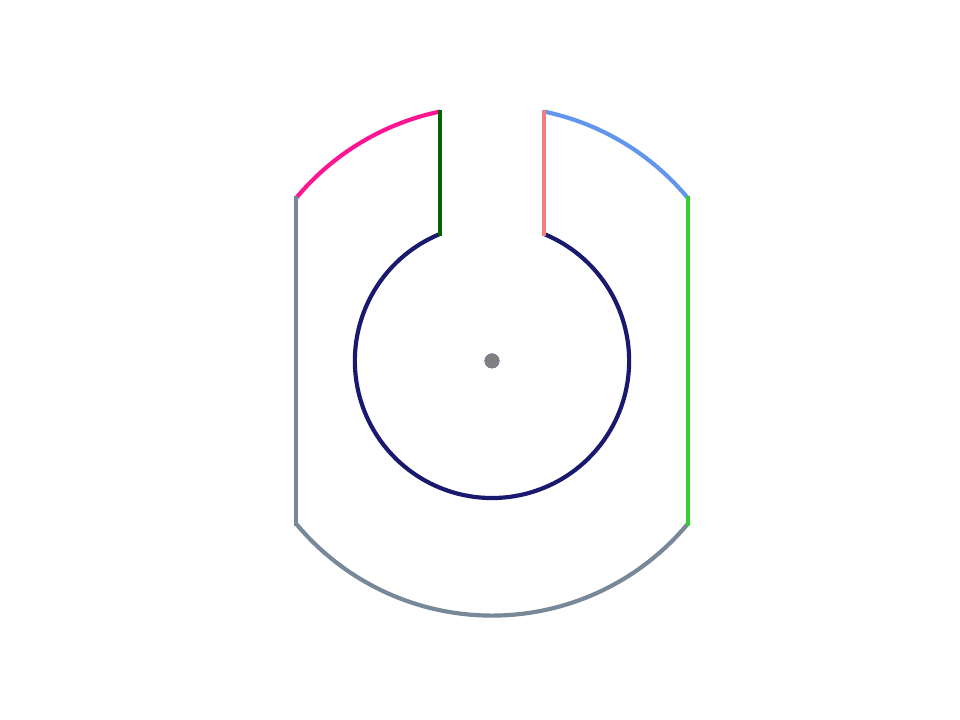} &
        \includegraphics[clip, trim=1.5cm 1.5cm 1.5cm 1.5cm, width=0.09\textwidth, height=0.05\textheight]{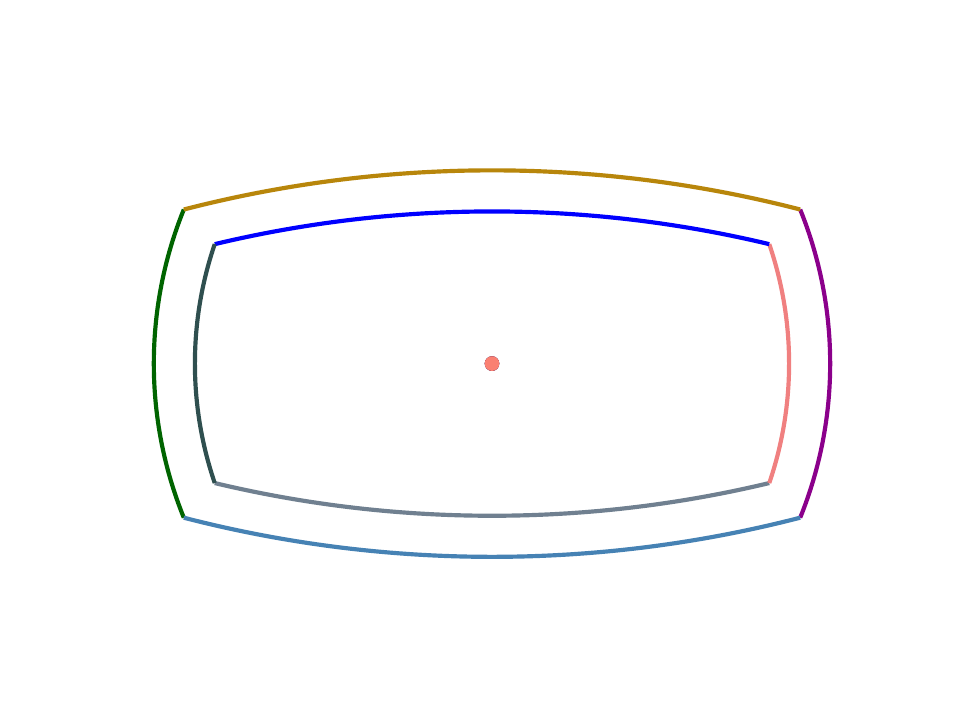} \\
    \end{tabular}

    \vspace{0.5em}
    \rule{0.9\textwidth}{0.4pt}
    \vspace{0.5em}

    % Middle: Diffusion Samples
    \begin{tabular}{cccccccc}
        \includegraphics[clip, trim=1.5cm 1.5cm 1.5cm 1.5cm, width=0.09\textwidth, height=0.05\textheight]{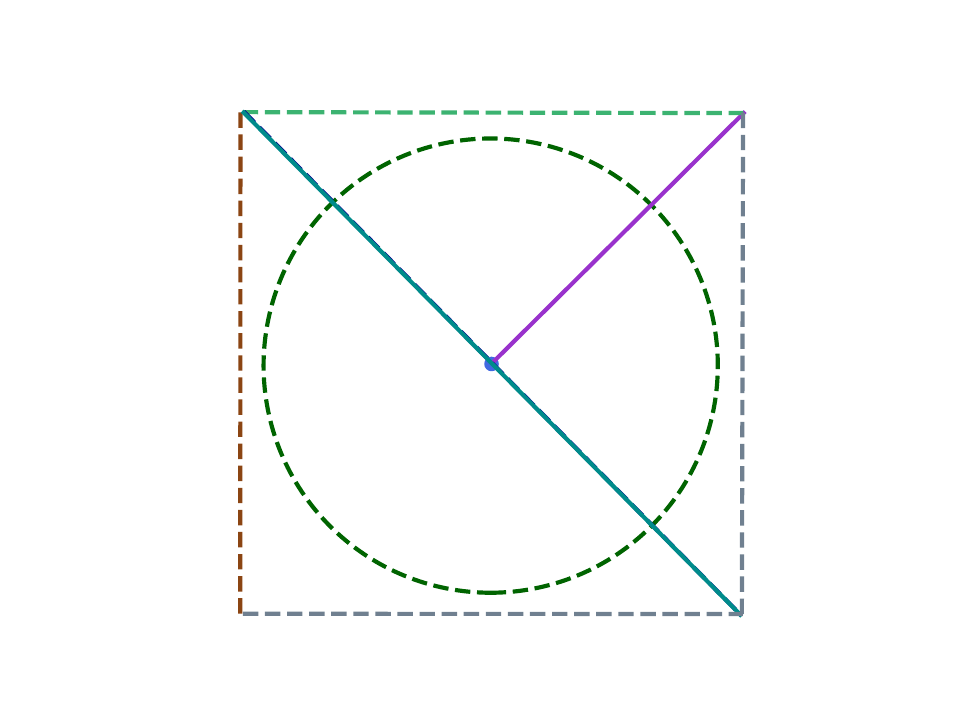} &
        \includegraphics[clip, trim=1.5cm 1.5cm 1.5cm 1.5cm, width=0.09\textwidth, height=0.05\textheight]{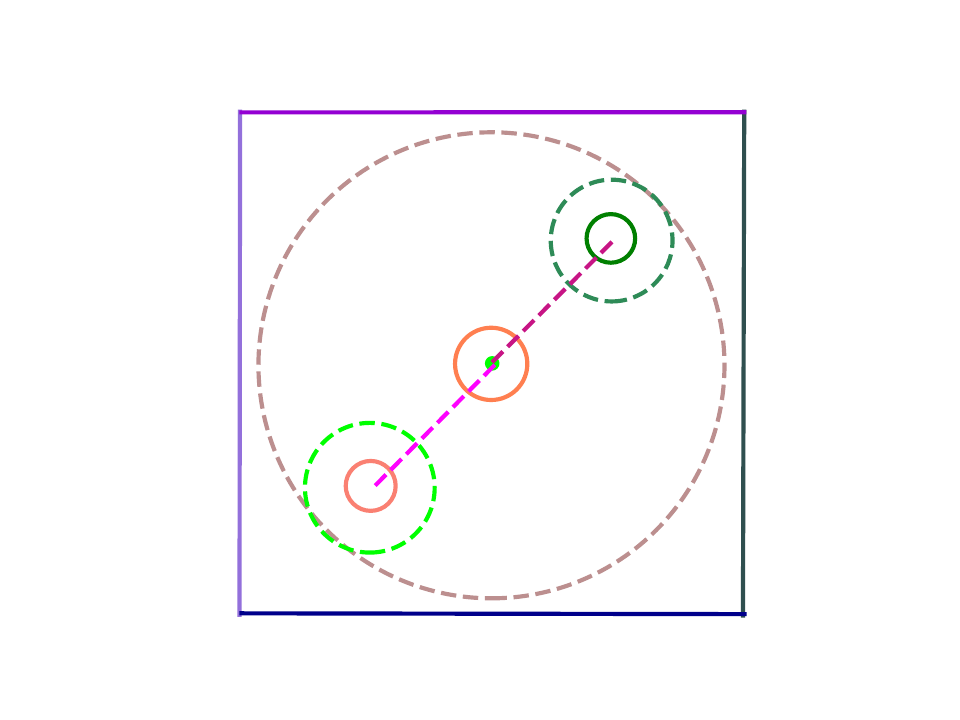} &
        \includegraphics[clip, trim=1.5cm 1.5cm 1.5cm 1.5cm, width=0.09\textwidth, height=0.05\textheight]{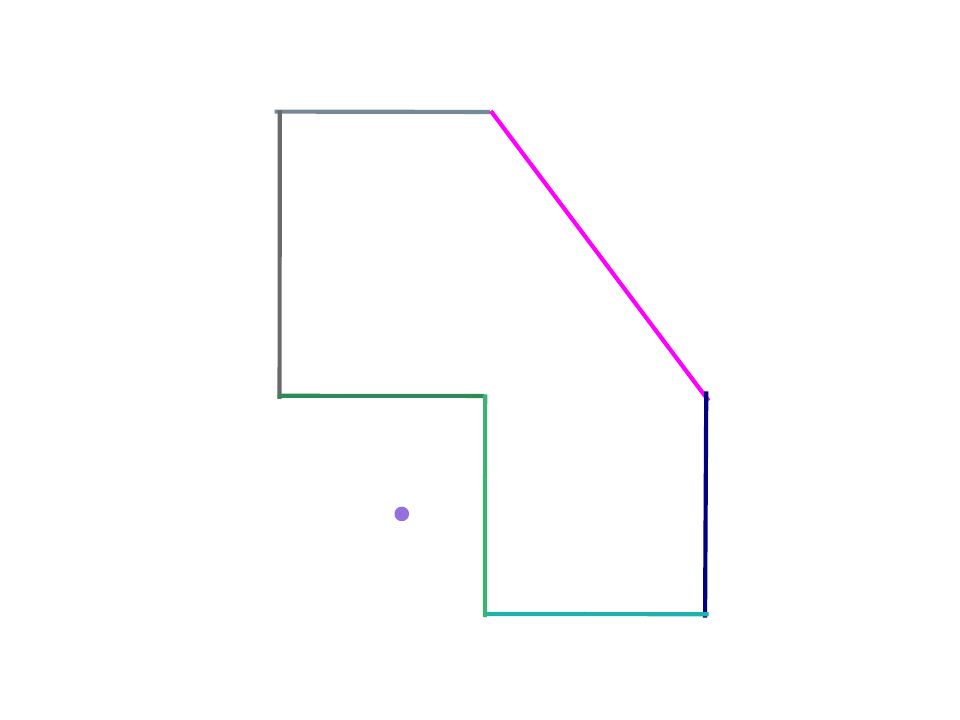} &
        \includegraphics[clip, trim=1.5cm 1.5cm 1.5cm 1.5cm, width=0.09\textwidth, height=0.05\textheight]{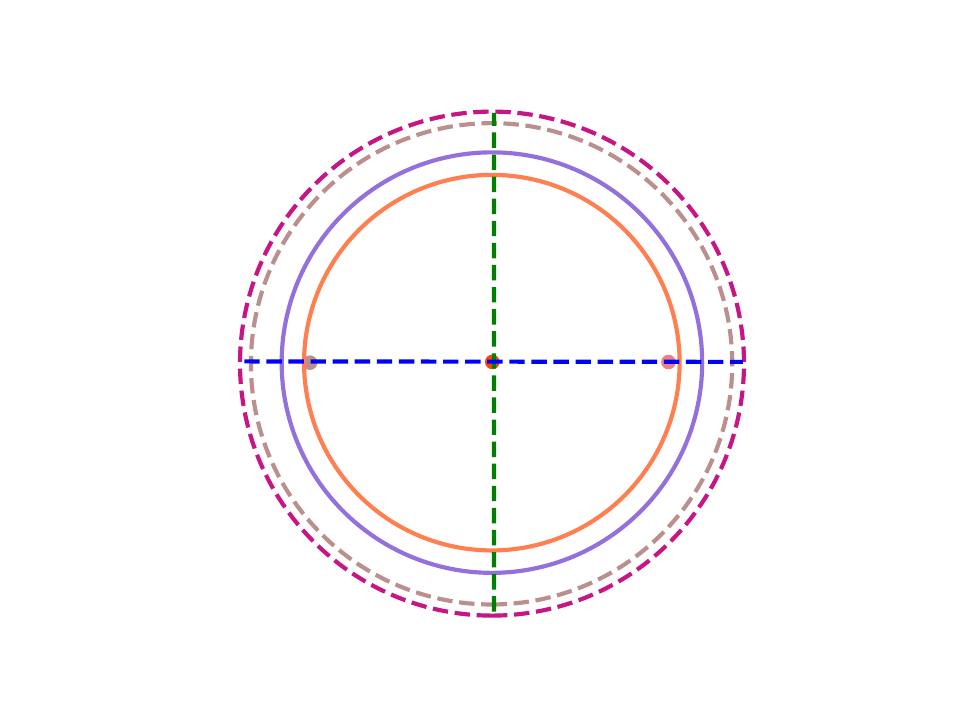} &
        \includegraphics[clip, trim=1.5cm 1.5cm 1.5cm 1.5cm, width=0.09\textwidth, height=0.05\textheight]{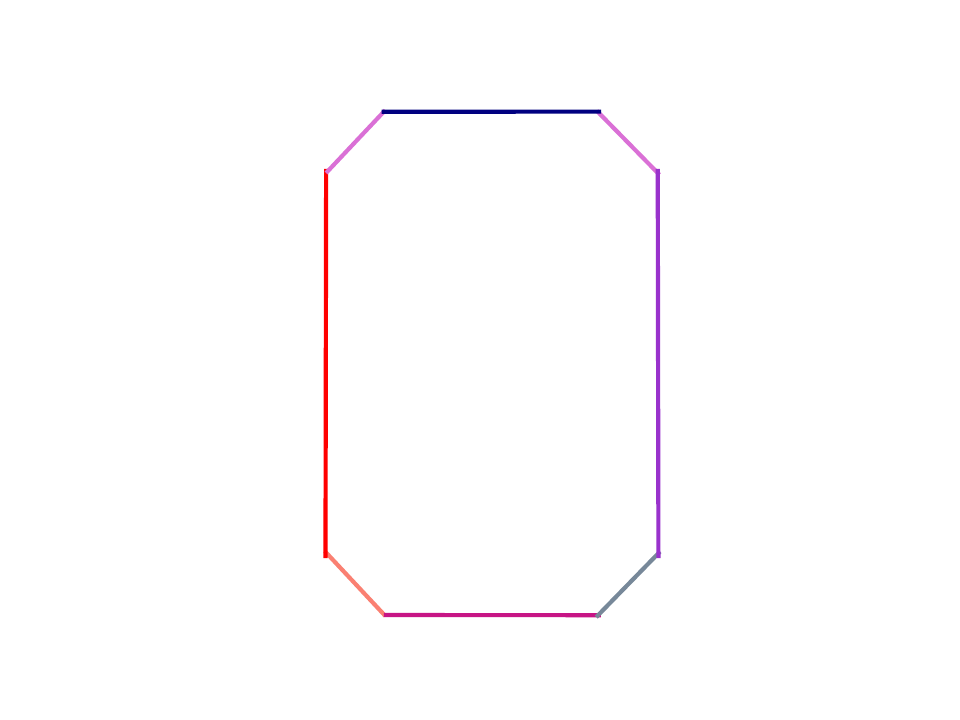} &
        \includegraphics[clip, trim=1.5cm 1.5cm 1.5cm 1.5cm, width=0.09\textwidth, height=0.05\textheight]{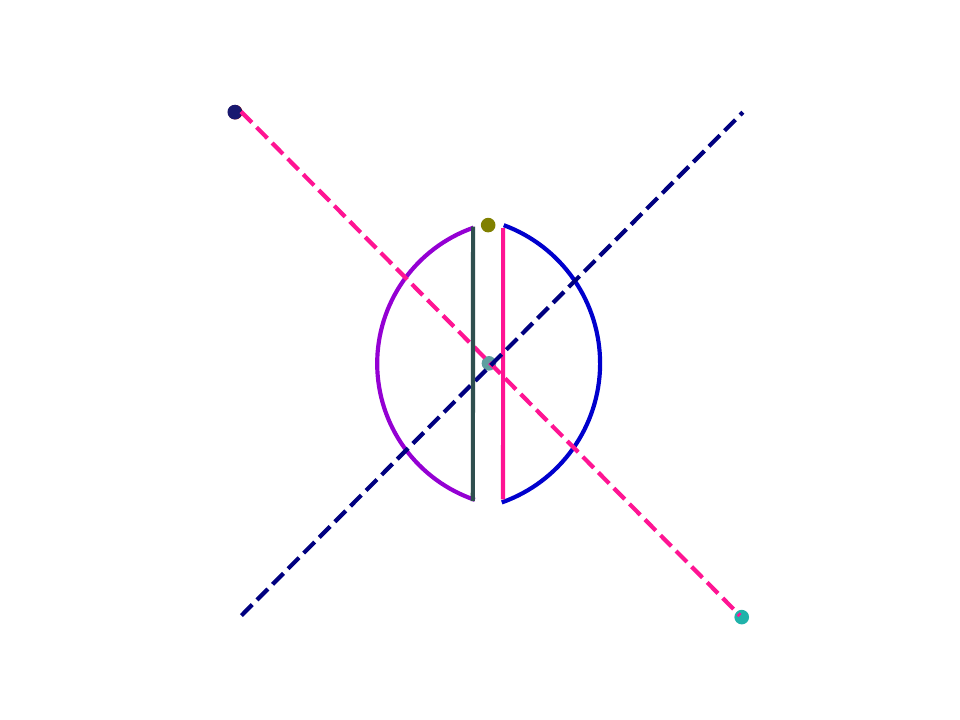} &
        \includegraphics[clip, trim=1.5cm 1.5cm 1.5cm 1.5cm, width=0.09\textwidth, height=0.05\textheight]{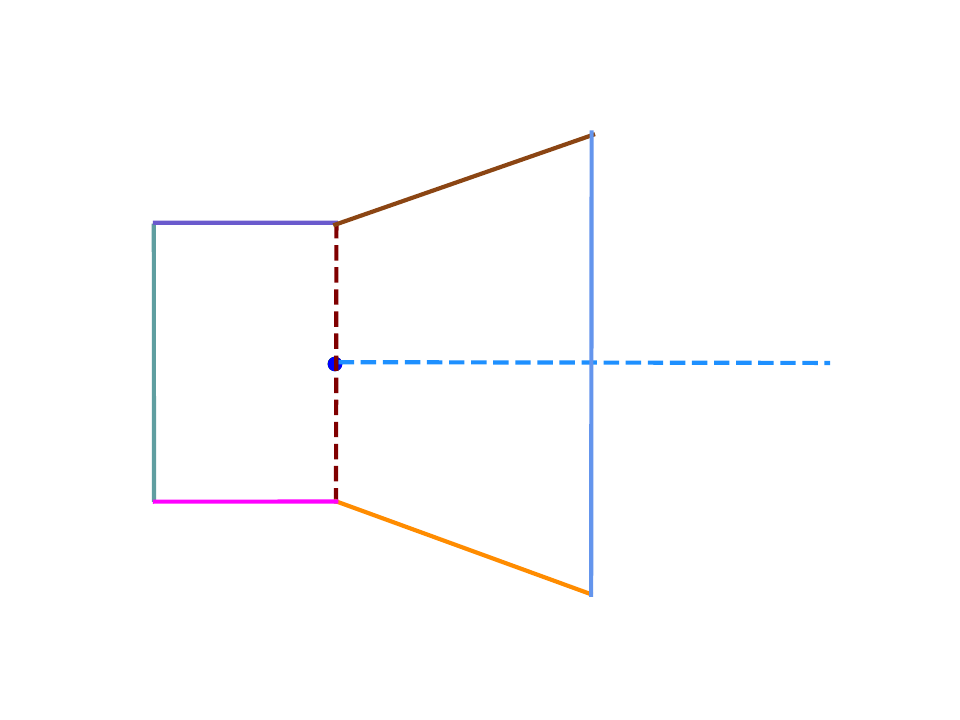} &
        \includegraphics[clip, trim=1.5cm 1.5cm 1.5cm 1.5cm, width=0.09\textwidth, height=0.05\textheight]{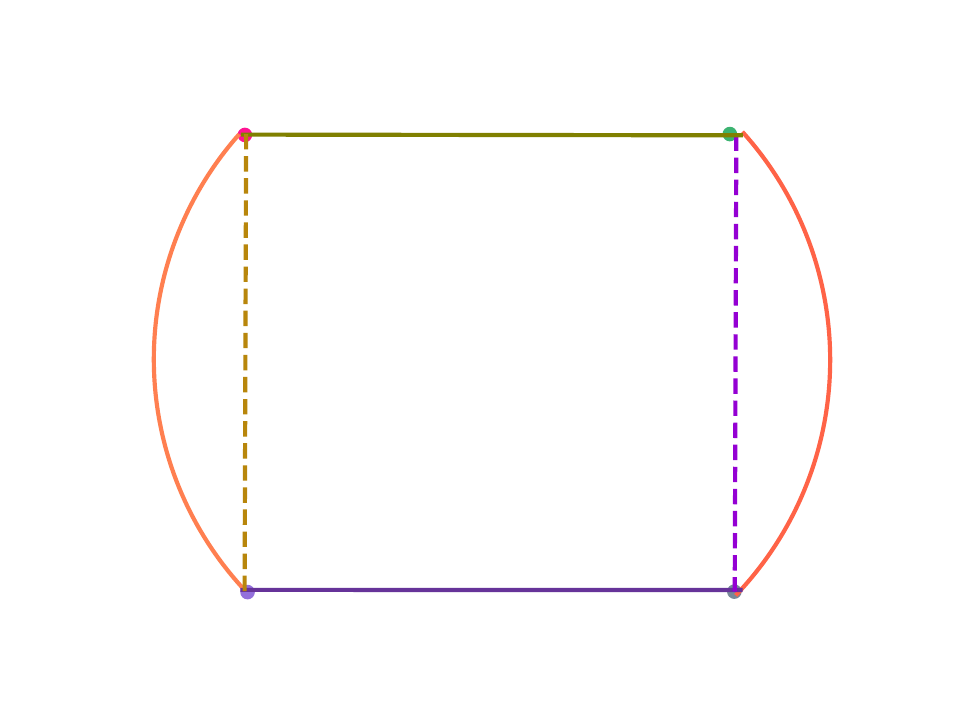} \\

        \includegraphics[clip, trim=1.5cm 1.5cm 1.5cm 1.5cm, width=0.09\textwidth, height=0.05\textheight]{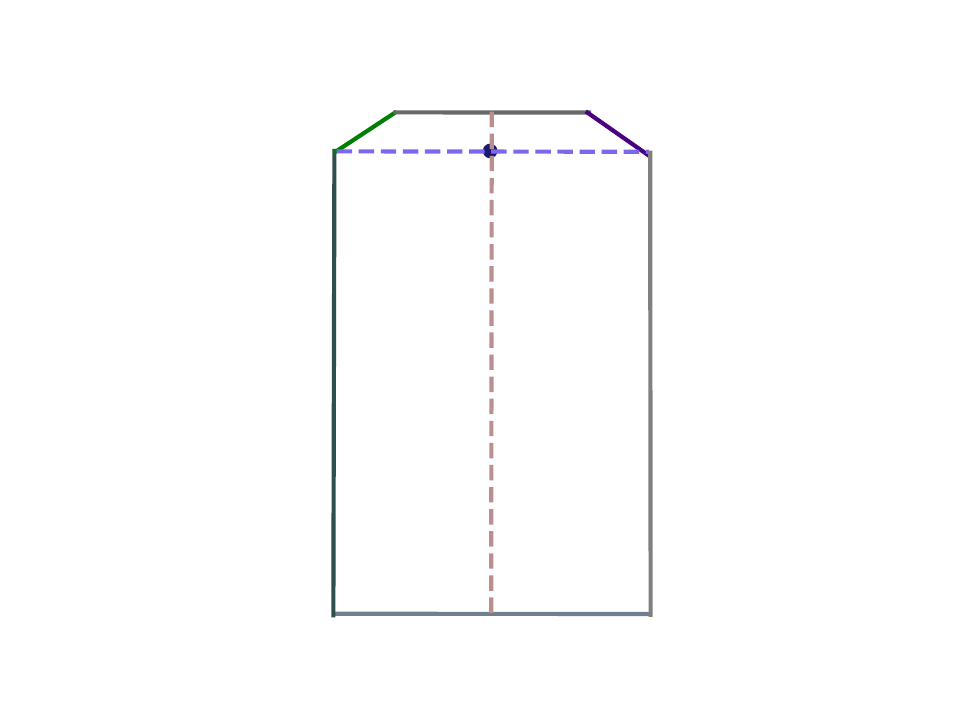} &
        \includegraphics[clip, trim=1.5cm 1.5cm 1.5cm 1.5cm, width=0.09\textwidth, height=0.05\textheight]{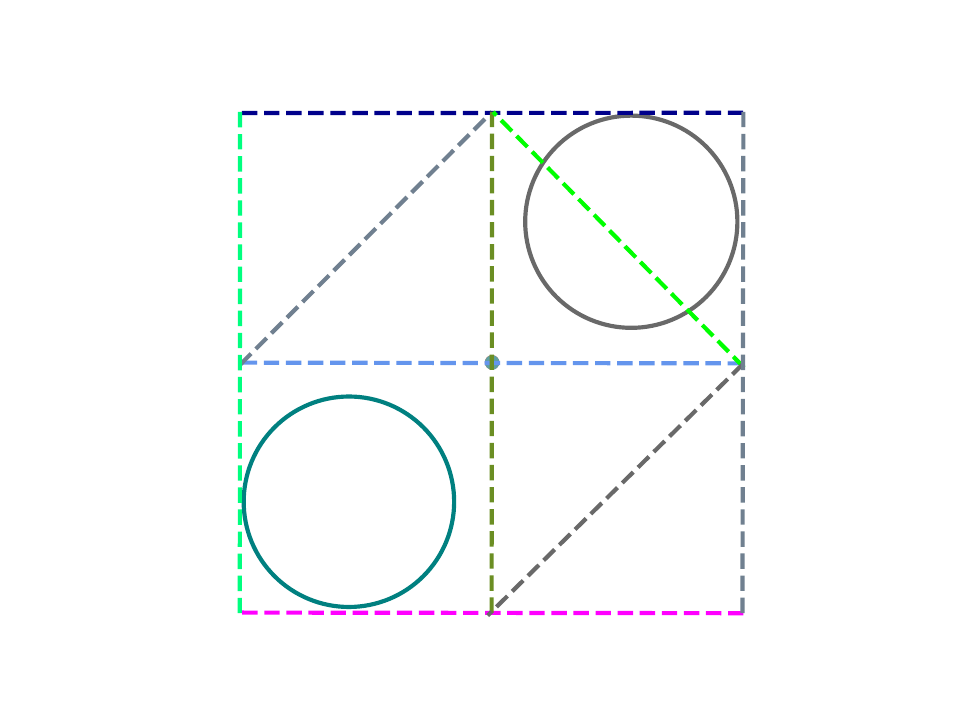} &
        \includegraphics[clip, trim=1.5cm 1.5cm 1.5cm 1.5cm, width=0.09\textwidth, height=0.05\textheight]{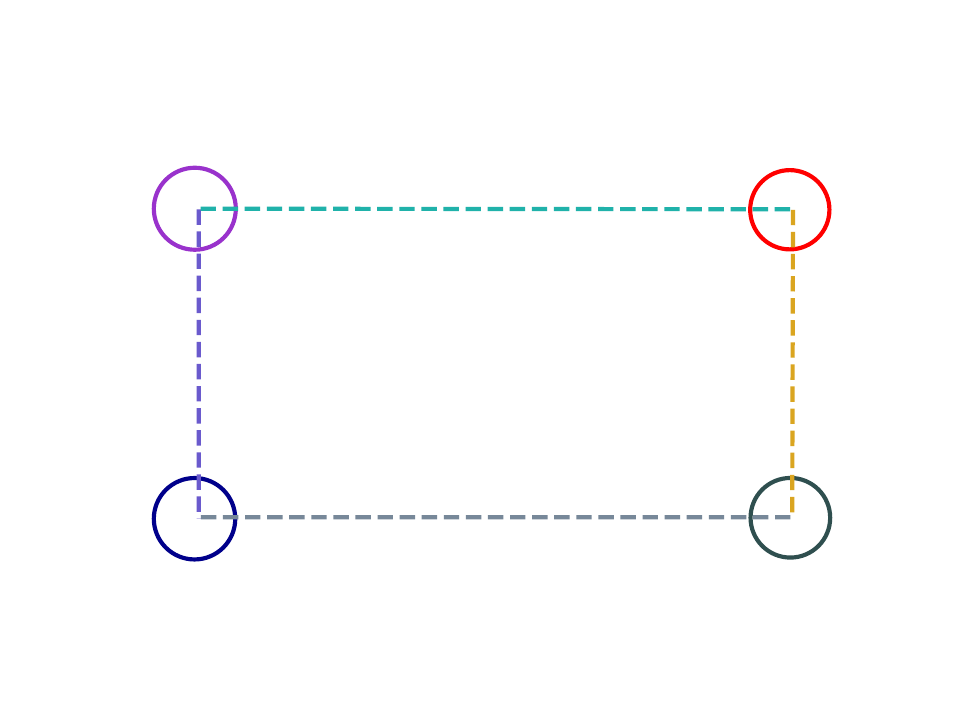} &
        \includegraphics[clip, trim=1.5cm 1.5cm 1.5cm 1.5cm, width=0.09\textwidth, height=0.05\textheight]{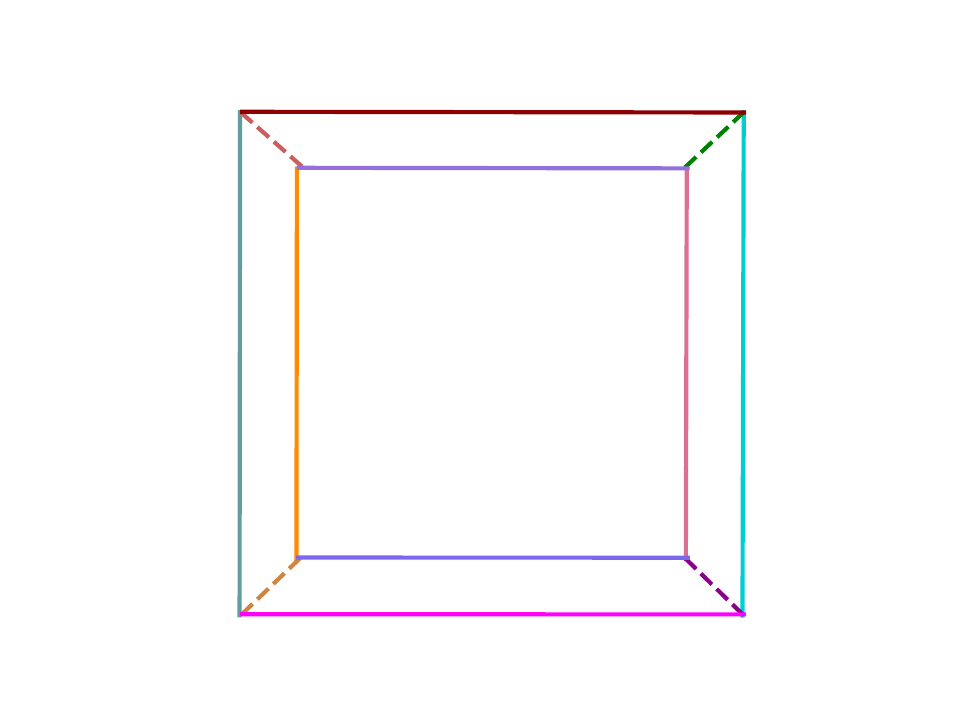} &
        \includegraphics[clip, trim=1.5cm 1.5cm 1.5cm 1.5cm, width=0.09\textwidth, height=0.05\textheight]{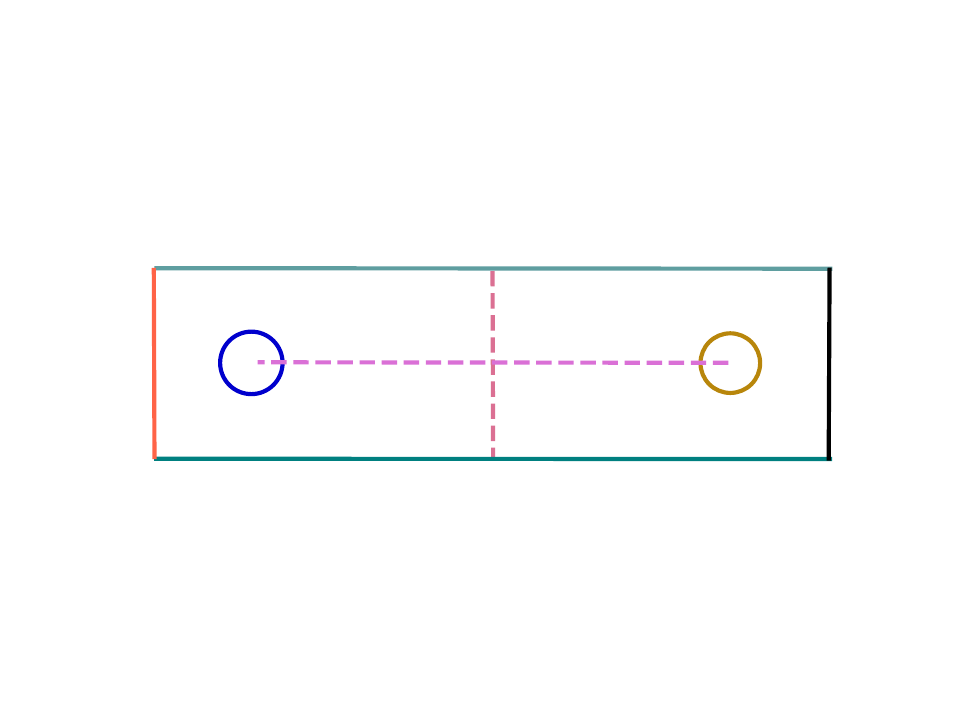} &
        \includegraphics[clip, trim=1.5cm 1.5cm 1.5cm 1.5cm, width=0.09\textwidth, height=0.05\textheight]{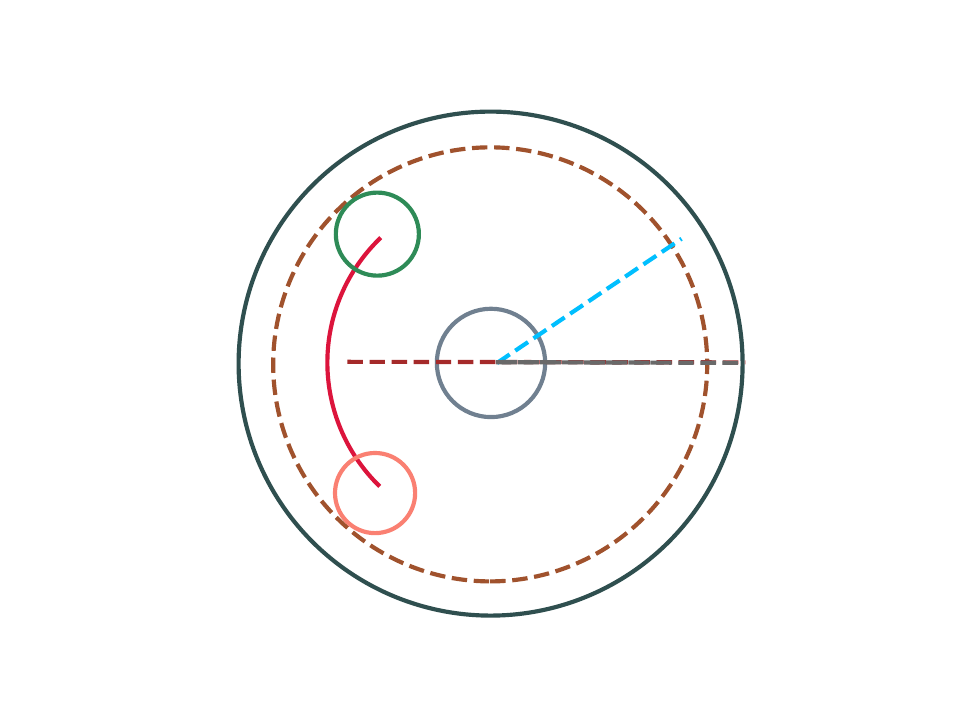} &
        \includegraphics[clip, trim=1.5cm 1.5cm 1.5cm 1.5cm, width=0.09\textwidth, height=0.05\textheight]{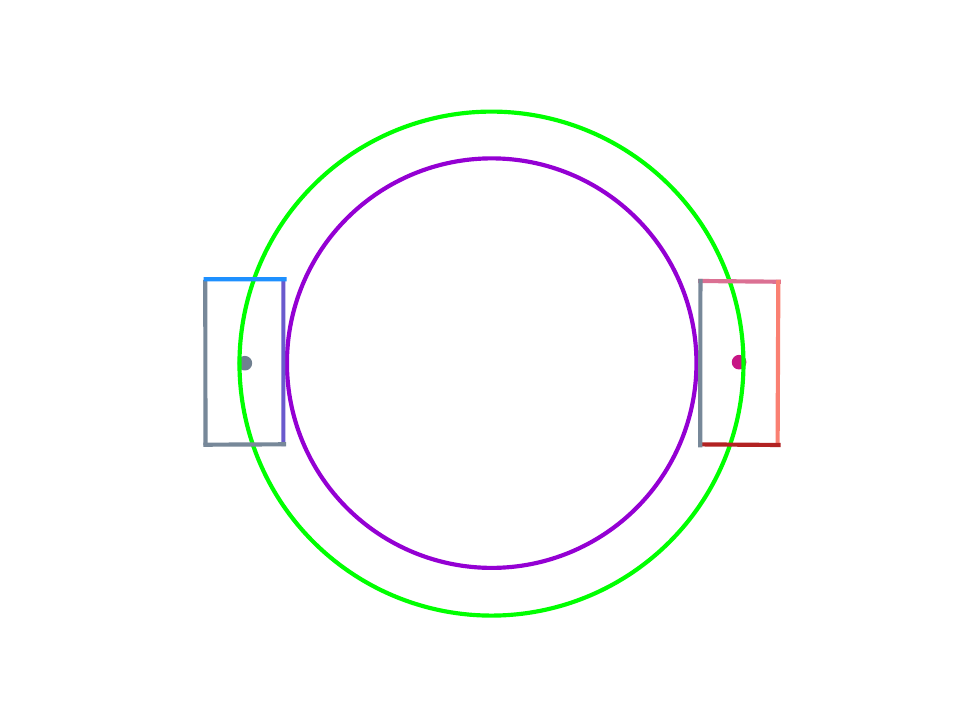} &
        \includegraphics[clip, trim=1.5cm 1.5cm 1.5cm 1.5cm, width=0.09\textwidth, height=0.05\textheight]{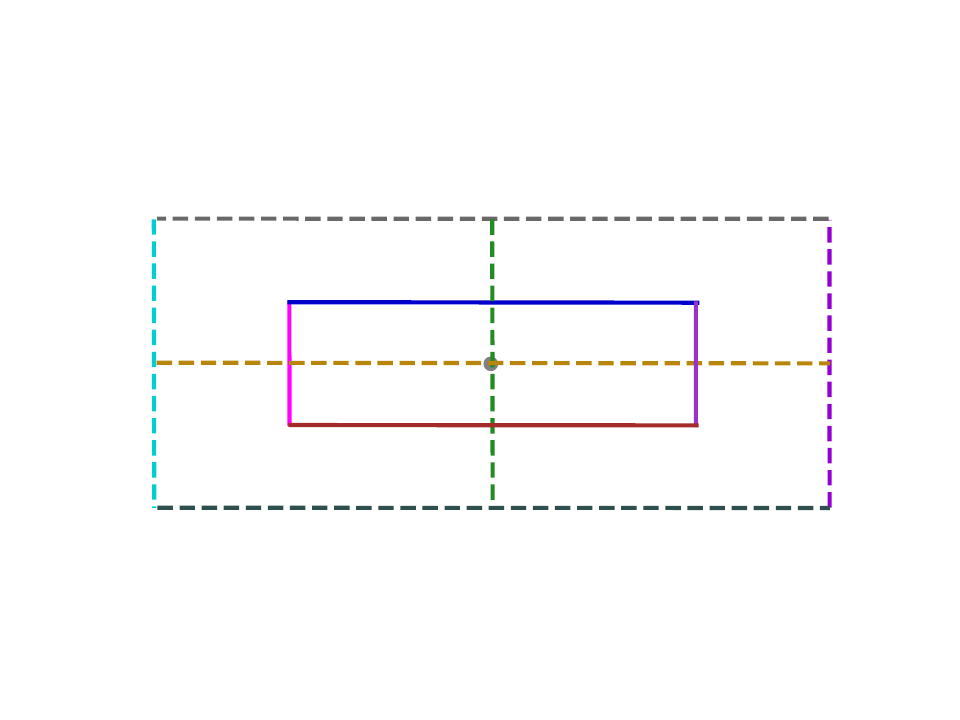} \\
    \end{tabular}

    \vspace{0.5em}
    \rule{0.9\textwidth}{0.4pt}
    \vspace{0.5em}

    % Bottom: Vitruvion Samples
    \begin{tabular}{cccccccc}
        \includegraphics[width=0.09\textwidth, height=0.05\textheight]{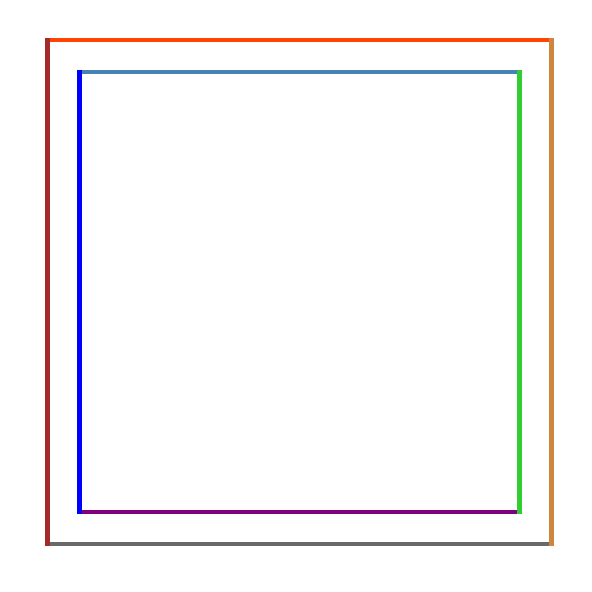} &
        \includegraphics[width=0.09\textwidth, height=0.05\textheight]{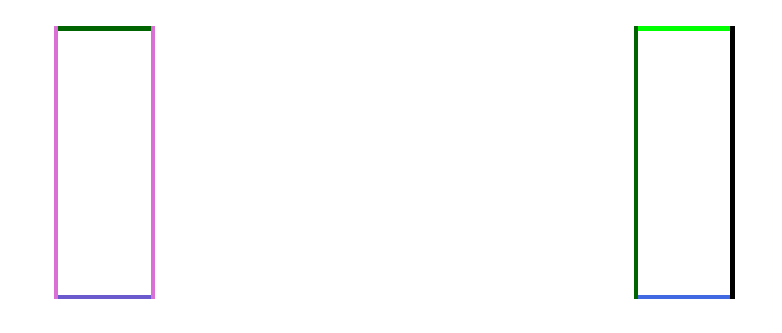} &
        \includegraphics[width=0.09\textwidth, height=0.05\textheight]{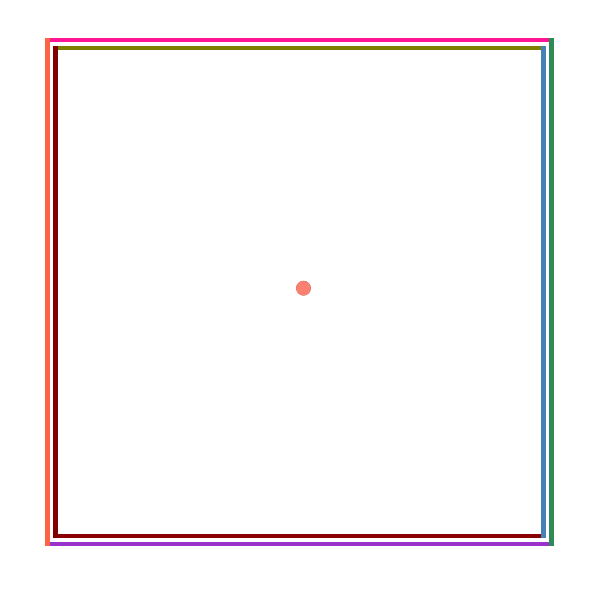} &
        \includegraphics[width=0.09\textwidth, height=0.05\textheight]{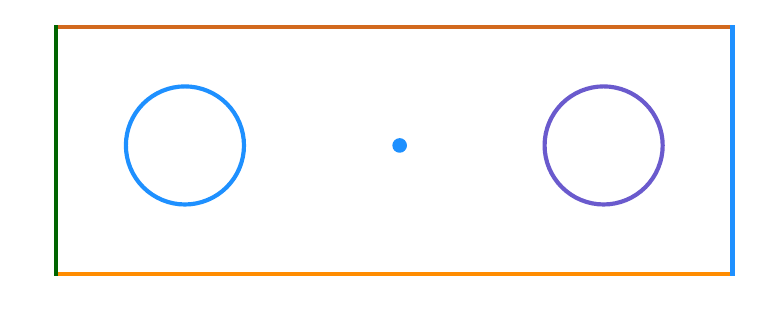} &
        \includegraphics[width=0.09\textwidth, height=0.05\textheight]{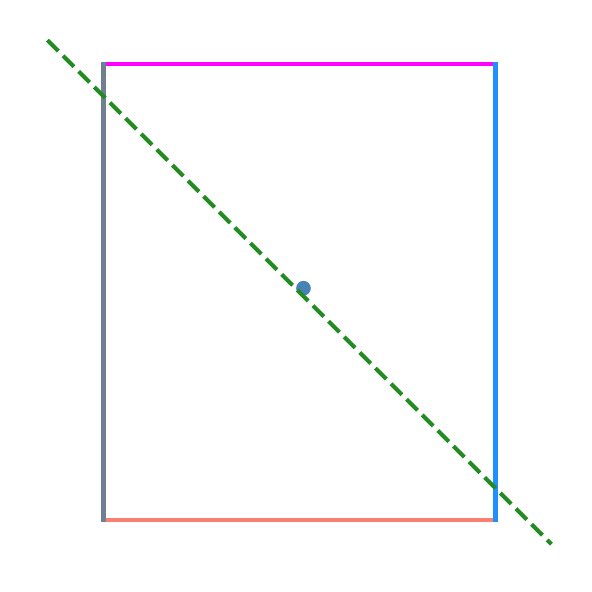} &
        \includegraphics[width=0.09\textwidth, height=0.05\textheight]{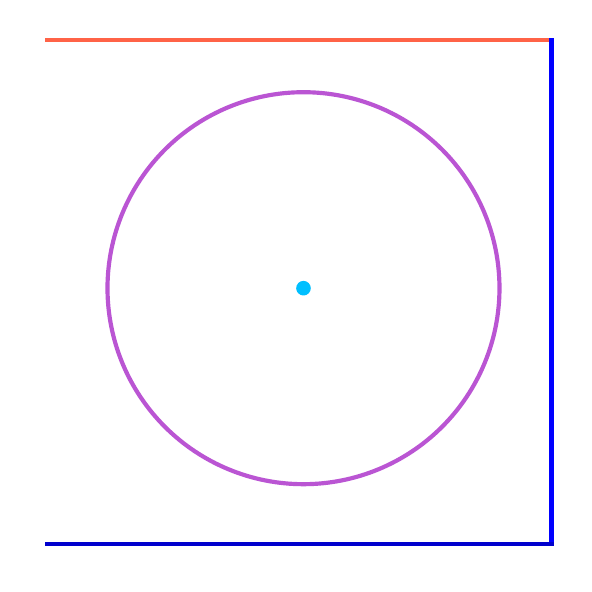} &
        \includegraphics[width=0.09\textwidth, height=0.05\textheight]{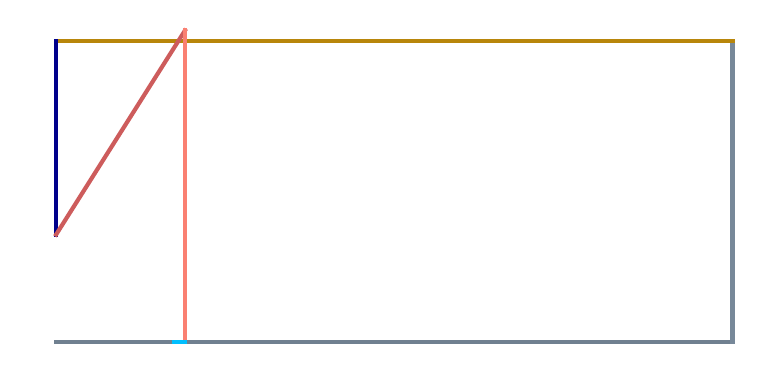} &
        \includegraphics[width=0.09\textwidth, height=0.05\textheight]{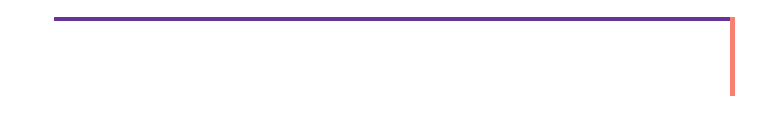} \\

        \includegraphics[width=0.09\textwidth, height=0.05\textheight]{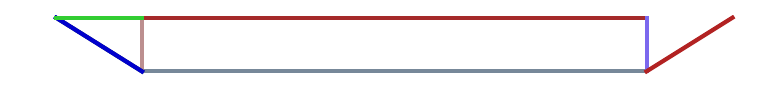} &
        \includegraphics[width=0.09\textwidth, height=0.05\textheight]{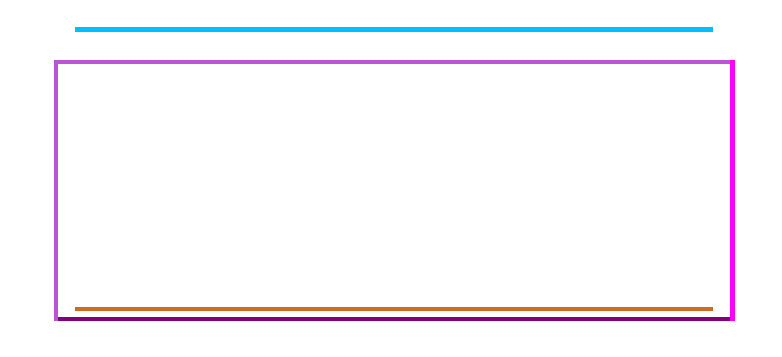} &
        \includegraphics[width=0.09\textwidth, height=0.05\textheight]{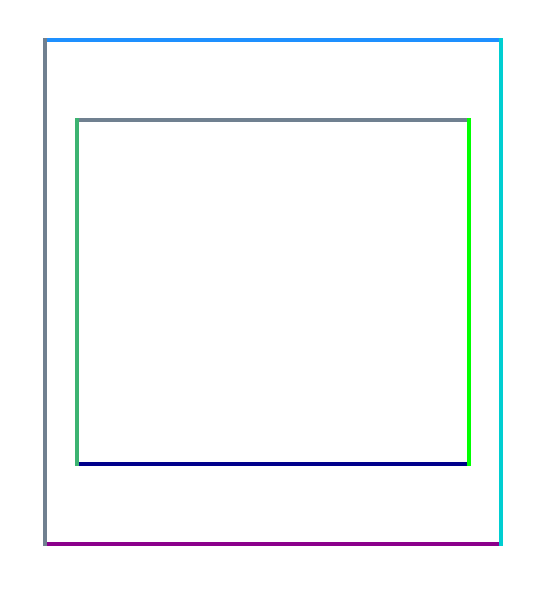} &
        \includegraphics[width=0.09\textwidth, height=0.05\textheight]{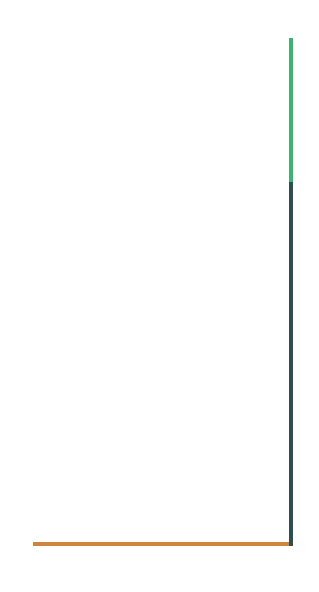} &
        \includegraphics[width=0.09\textwidth, height=0.05\textheight]{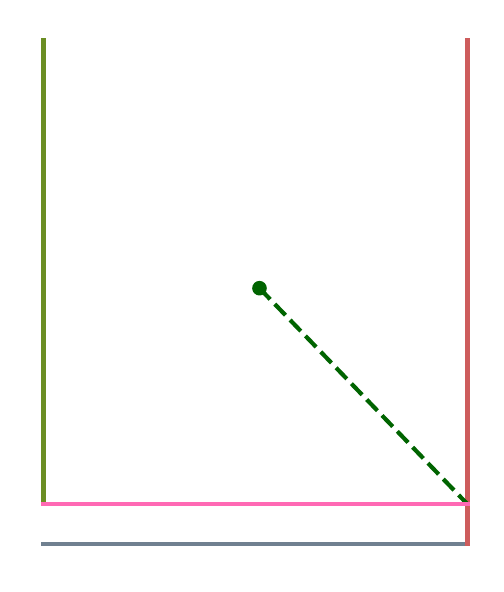} &
        \includegraphics[height=0.05\textheight]{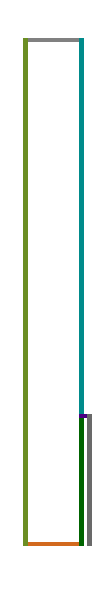} &
        \includegraphics[height=0.05\textheight]{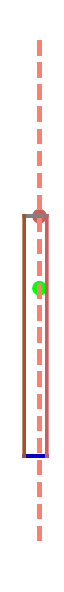} &
        \includegraphics[width=0.09\textwidth, height=0.05\textheight]{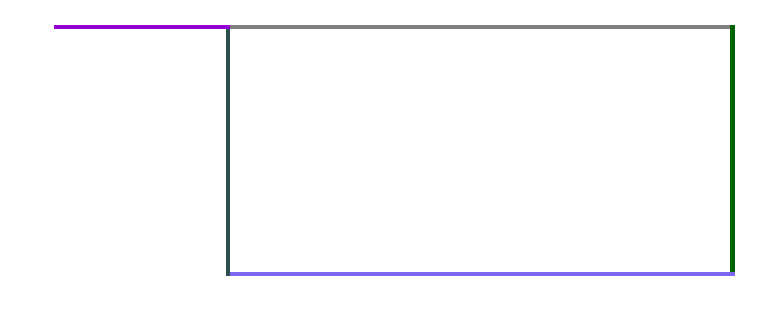} \\
    \end{tabular}

    \caption{Comparison of CAD sketches from the SketchGraphs dataset (top), generations from SketchDNN (middle), and generations from Vitruvion (bottom).}
    \label{fig:dataset_diffusion}
\end{figure*}

We empirically validate our diffusion model, SketchDNN, against prior art. We also perform ablation studies to show that our approaches to address the heterogeneity and permutation invariance of sketches improves the performance of our model. We present metrics for likelihood and sample quality. Additionally we provide a qualitative comparison between our model and the dataset in Figure \ref{fig:dataset_diffusion}.

% \subsection{Dataset Preprocessing}
% We use the CAD sketch dataset introduced in SketchGraphs by Seff et al. \cite{seff_sketchgraphs_2020}. The dataset is comprised of 15 million human-created CAD sketches scraped from Onshape, a cloud centric CAD platform. The majority of sketches in the dataset have less than 8 primitives comprising approximately 84\% of the dataset, they are very simple box-like shapes and are almost identical to one another. Accordingly we filter out sketches with less than 8 primitives to avoid biasing our model towards these simple sketches. We also filter out sketches with more than 16 primitives due to time and resource constraints. Additionally, sketches varied greatly in dimensions from a few millimeters to several meters large, so we perform a normalization procedure where we shift sketches such that their bounding boxes are centered on the origin, and we rescale them so that the dimensions of their bounding box don't exceed 1 meter. After normalization, we remove duplicate sketches from our dataset which arise This arises primarily from
% users creating sketches by following tutorials, reusing previously created sketches, or using collaboration tools provided by OnShape to copy sketches from each other. We group identical sketches and keep 1 sketch from each group. Lastly, we simplify the parameterizations of each primitive inline with the Section \ref{sec:representation}.

\subsection{Dataset Preprocessing}  
We used the CAD sketch dataset introduced in SketchGraphs by Seff et al.~\cite{seff_sketchgraphs_2020}, which consists of 15 million human-created CAD sketches extracted from Onshape, a cloud-based CAD platform. The dataset is highly imbalanced, with approximately 84\% of sketches containing fewer than eight primitives. These sketches predominantly feature simple boxes that are almost identical to one another. To avoid biasing our model towards such sketches, we excluded sketches with fewer than eight primitives. Additionally, we discarded sketches with more than 16 primitives due to time and resource constraints. The sketches also exhibit significant variation in position and scale, ranging from a few millimeters to several meters in size. Accordingly, we applied a normalization procedure where each sketch was translated and rescaled, such that the bounding box was centered at the origin and inscribed within the unit square. 

Next, the dataset contains a substantial number of duplicate sketches, primarily due to users following tutorials, reusing previously created designs, or leveraging Onshape’s collaboration tools to replicate sketches. To again avoid bias, we grouped identical sketches and kept one sketch from each group. For this, we first performed 8-bit quantization on continuous parameters, then sorted the rows in the sketch matrix for a canonical ordering, and finally picked a corresponding unquantized sketch for each group. This was done to avoid sketches with identical geometry but differing orderings from being labeled as nonidentical. Lastly, we split the remaining 1.4 million CAD sketches into 3 subsets: we reserved 90\% for
training, 5\% for validation, and 5\% for testing. Finally, we simplified the parameterization of each primitive in accordance with Section~\ref{sec:representation}. A similar procedure was performed by \cite{para_sketchgen_2021} and \cite{seff2021vitruvion}.

\subsection{Model and Training Setup}  
SketchDNN is based on a simplified variant of the diffusion transformer architecture proposed by Peebles et al.~\cite{peebles2023scalablediffusionmodelstransformers}, with positional encodings and gate conditioning omitted. The omission of positional encodings ensures that the denoiser network remains permutation equivariant, as discussed in Section~\ref{sec:sketch diffusion}. We use an embedding size of $512$ with a depth of $32$ transformer blocks. We trained our model to generate samples over $T=2000$ timesteps. Since the diffusion process involves reconstructing the ground-truth clean sketch from a noisy version, we use reconstruction loss, \(\mathcal{L}_{\text{RECON}}\), composed of Mean-Squared Error (MSE) loss for continuous variables $\mathbf{x}$ and Cross Entropy (CE) loss for discrete variables $\mathbf{y}$:  
\[
\mathcal{L}_{\text{RECON}} = 
\begin{cases} 
\lambda\mathcal{L}_{\text{MSE}}(\hat{\mathbf{x}}^\theta_0, \mathbf{x}_0) + \mathcal{L}_{\text{CE}}(\hat{\mathbf{y}}^\theta_0, \mathbf{y}_0) & \text{if } t \leq 150, \\
\mathcal{L}_{\text{MSE}}(\hat{\mathbf{x}}^\theta_0, \mathbf{x}_0) + \mathcal{L}_{\text{CE}}(\hat{\mathbf{y}}^\theta_0, \mathbf{y}_0) & \text{if } t > 150
\end{cases}
\]
where we set $\lambda=16$. We found weighing the MSE loss of smaller timesteps to be more beneficial to sample quality, than simply weighing the MSE loss higher over all timesteps. The model was trained for 1000 epochs using a batch size of \( 8 \times 512 \), distributed across 8 NVIDIA A30 GPUs. A constant learning rate of \( 1 \times 10^{-4} \) was used throughout training.  

To avoid hindering superposition, we use the ground truth primitive type to mask irrelevant parameters before computing the MSE loss. Thus, our denoiser network is trained to predict parameters for all possible primitive types and assign confidence scores to each. For example, if the true primitive type is a line, the model still predicts parameters for other types, such as circles or arcs, but assigns probabilities reflecting its confidence in each prediction. As a result, during inference, we weight the continuous variables outputted by our model by their corresponding rescaled type probabilities. We rescale the predicted type probabilities by dividing each probability vector with its maximum element. This rescaling prevents relevant parameter values from decaying to zero throughout the reverse diffusion process while irrelevant parameters are masked out, ensuring more accurate predictions of noisy primitives. This adjustment is necessary since the model is trained to expect irrelevant parameters to approach zero at smaller timesteps, discrepancies may arise during the reverse process because we don't train the model to zero out irrelevant parameters. Applying this weighting ensures that the model's predictions better align with the ground truth, particularly at smaller timesteps.

\section{Results}\label{sec:results}  
We evaluate SketchDNN against the most relevant prior methods for CAD sketch generation: SketchGen~\cite{para_sketchgen_2021} and Vitruvion~\cite{seff2021vitruvion}. Both approaches represent CAD sketches as sequences of tokens and generate them autoregressively. Vitruvion, as the previous state of the art, serves as our primary baseline for comparison. Additionally, we assess the effectiveness of our diffusion-based approach by comparing it with conventional discrete diffusion methods as an ablation study, including categorical diffusion and latent diffusion. For latent diffusion, we train a variational autoencoder (VAE) to map sketches into a latent space and train an auxillary diffusion network to generate latents in a manner analogous to Stable Diffusion. For categorical diffusion, we train an alternative version of SketchDNN, which we term SketchDNN (Cat.), that is trained using the categorical diffusion paradigm by \cite{hoogeboom2021argmax} while keeping all other architectural and training details identical. Similarly, we conduct an ablation study to examine the impact of permutation invariance in our generation process. To this end, we train an alternative version of SketchDNN,  which we term SketchDNN (Pos.), that incorporates positional encodings while keeping all other aspects identical.

% \subsection{Negative Log-Likelihood}
% We present the negative log-likelihood (NLL) to measure the aptitude of our model in learning the distribution of CAD graphs. A lower NLL indicates better generalization and improved model performance, though it is not necessarily indicative of sample quality. We present these results in Table \ref{tab:nll-calculations}. We note that SketchDNN outperforms the prior SOA model Vitruvion, demonstrating that our diffusion approach is much better suited to the task of CAD sketch generation than autoregressive modeling. It seems our proposed Gaussian-Softmax diffusion paradigm performs significantly better than traditional categorical diffusion. Infact, SketchDNN (Cat.) performs significantly worse than Vitruvion and SketchGen, indicating that superposition significantly improves the performance of our models. We find this result surprising since the full context of a sketch is always available to SketchDNN (Cat.) because it is still a diffusion model, yet it still performs worse. Furthermore, it appears our permutation-invariant diffusion paradigm also benefits model performance to a noticeable but much lesser degree, as indicated by the relatively close performance of SketchDNN vs SketchDNN (Pos.) in terms of NLL. Thus, it appears superposition is the main driver behind improving CAD sketch generation, and that permutation-invariant denoising plays a significantly smaller role. 

\begin{table}[ht]
\caption{The NLL for each model is reported in bits, where a lower NLL is better. Notably, our diffusion model achieved SOA NLL. SketchDNN (Pos.) is the ablation that includes positional encodings, where SketchDNN (Cat.) is the ablation that uses categorical diffusion. Bold represents state of the art performance, while underlining represents second best performance.}
\label{tab:nll-calculations}
\vskip 0.15in
    \centering
    \begin{tabular}{llll}
        \hline
        Method              & Bits/Sketch$\downarrow$ & Bits/Primitive$\downarrow$ \\ \hline
        SketchDNN (Ours)    & \textbf{81.33}    &  \textbf{5.08} \\
        SketchDNN (Pos.)    & \underline{83.03} &  \underline{5.18}  \\
        SketchDNN (Cat.)    & 106.10         &  6.63  \\
        Vitruvion           & 84.80          &  8.19  \\
        SketchGen           & 88.22          &  8.60  \\
    \end{tabular}
\vskip -0.1in
\end{table}

\subsection{Negative Log-Likelihood}  
We evaluate the negative log-likelihood (NLL) as a measure of our model’s ability to learn the distribution of CAD graphs. A lower NLL indicates better generalization and improved model performance; however, it does not necessarily correlate with sample quality. For diffusion models, computing the exact NLL is intractable, therefore, we follow the standard approach of approximating the NLL using the ELBO, which satisfies the inequality \(\text{ELBO} \geq \text{NLL}\). We calculate the NLL over our test set of 70K CAD sketches. The results are presented in Table~\ref{tab:nll-calculations}. SketchDNN achieves a lower NLL than the previous state-of-the-art model, Vitruvion, demonstrating that our diffusion-based approach is better suited to CAD sketch generation than autoregressive modeling. 

Notably, our Gaussian-Softmax diffusion paradigm significantly outperforms traditional categorical diffusion. In fact, SketchDNN (Cat.), which employs categorical diffusion, performs worse than both Vitruvion and SketchGen. This suggests that the superposition mechanism in Gaussian-Softmax diffusion is a key factor in improving model performance. This finding is particularly surprising, as SketchDNN (Cat.), being a diffusion model, still has access to the full sketch context at any given time, yet performs worse. Additionally, our permutation-invariant diffusion paradigm appears to provide a measurable but smaller benefit, as indicated by the relatively close NLL scores of SketchDNN and SketchDNN (Pos.). These results suggest that while permutation-invariant denoising contributes to performance, the primary factor driving improvements in CAD sketch generation is the use of superposition.  

\begin{table}[ht]
\caption{The FID, precision, and recall are presented for unconditional primitive generation. A lower FID is better, a higher precision is better, and a higher recall is better. We note that our diffusion model has achieved SOA FID and recall. SketchDNN (Pos.) is the ablation that includes positional encodings, where SketchDNN (Cat.) is the ablation that uses categorical diffusion. Bold represents state of the art performance, while underlining represents second best performance.}
\label{tab:fid-calculations}
\vskip 0.15in
    \centering
    \begin{tabular}{llll}
        \hline
        Method              & FID$\downarrow$    & Precision$\uparrow$ & Recall$\uparrow$         \\ \hline
        SketchDNN (Ours)    & \textbf{7.80}   & \underline{0.246}     & \textbf{0.266}          \\
        SketchDNN (Pos.)    & \underline{10.26}           & 0.230     & \underline{0.245} \\
        Latent Diffusion    & 93.34           &  0.134    & 0.033 \\
        SketchDNN (Cat.)    & 148.93          &  0.117    & 0.028 \\
        Vitruvion           & 16.04  &  \textbf{0.294}    & 0.176          \\ \hline
    \end{tabular}
\vskip -0.1in
\end{table}

\subsection{Sample Quality}
Here we present the Fréchet Inception Distance (FID), precision, and recall as metrics to evaluate the fidelity of generated CAD graphs, as shown in Table \ref{tab:fid-calculations}. Fidelity measures how closely synthetic CAD graphs resemble real CAD graphs. Standard image generation metrics are employed to assess sample quality, for this we render CAD sketches as monochromatic images. A lower FID score indicates higher fidelity samples, and our diffusion model achieves state-of-the-art (SOA) FID. Higher precision reflects a closer resemblance between generated and real samples, minimizing irrelevant or low-quality outputs, while higher recall indicates greater diversity in generated samples, capturing the full range of real data variations. Precision and recall often exhibit an inverse relationship, where improvements in one may lead to reductions in the other. As is standard in image generation literature, we compute these metrics using InceptionV3 over 10K CAD graphs from our test set.

% We find similar results as those for NLL, in that our Gaussian-Softmax diffusion paradigm again significantly outperforms other sketch generation paradigms. Similarly, SketchDNN (Cat.) performs the worst out of all methods, indicating that categorical diffusion is not a good fit for CAD sketch generation. The latent diffusion model we train, surprisingly also does not perform well. This is especially surprising considering the performance of latent diffusion models in other domains such as image generation with stable diffusion. Again, SketchDNN and SketchDNN (Pos.) achieve somewhat close results reinforcing that permutation-invariant denoising has a much smaller effect on generation quality than our Gaussian-Softmax diffusion paradigm. Thus, since both SketchDNN and SketchDNN (Pos.) significantly outperform all other models, we conclude that our novel discrete diffusion paradigm improves CAD sketch generation performance specifically because it accomodates superposition.

We observe similar trends as in the NLL evaluation, with our Gaussian-Softmax diffusion paradigm again significantly outperforming alternative sketch generation approaches. Notably, SketchDNN (Cat.) yields the worst performance among all methods, further indicating that superposition plays a large role in CAD sketch generation. Surprisingly, the latent diffusion model also underperforms, indicating that latent space diffusion is ill suited for CAD sketch generation. Once again, SketchDNN and SketchDNN (Pos.) achieve comparable results, reinforcing the conclusion that permutation-invariant denoising has a much smaller effect on generation quality than superposition. Since these models substantially outperform all other models, we conclude that the primary contributor to performance is the ability of our Gaussian-Softmax paradigm to accommodate superposition.

\subsection{Failure Cases}
\begin{figure}[h]
    \centering
    \includegraphics[clip, trim=3.5cm 1.5cm 3.5cm 1.5cm, width=0.32\columnwidth]{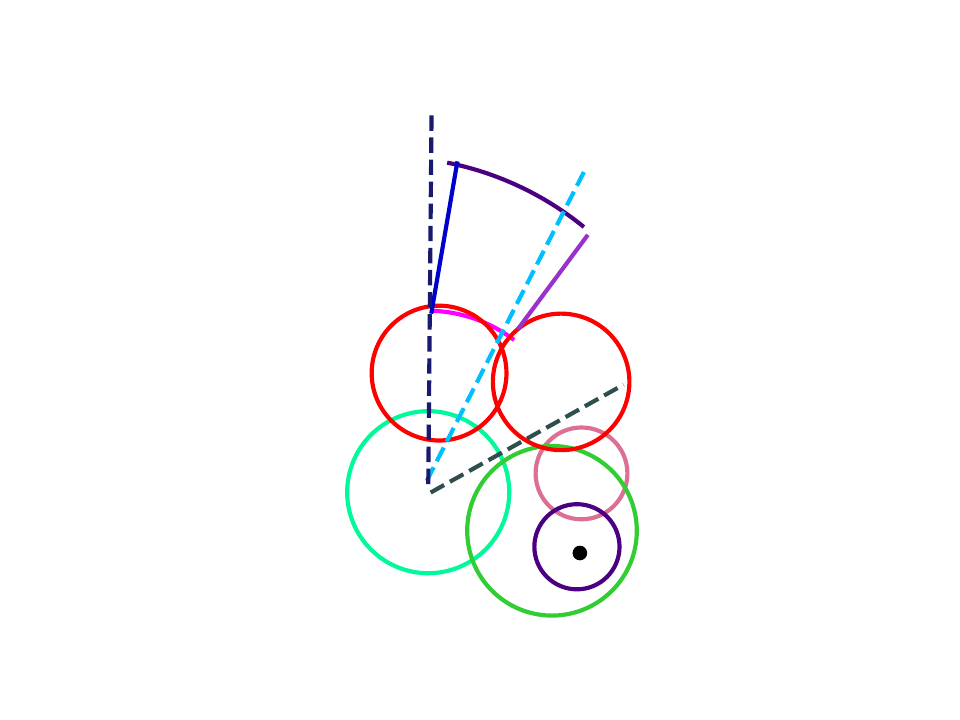}
    \hfill
    \includegraphics[clip, trim=1.5cm 2cm 1.5cm 0cm, width=0.32\columnwidth]{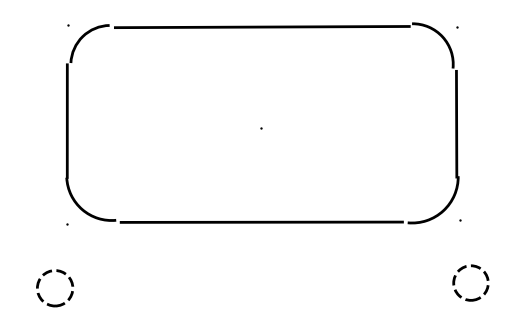}
    \hfill
    \includegraphics[clip, trim=1.5cm 1.5cm 6cm 5.5cm, width=0.32\columnwidth]{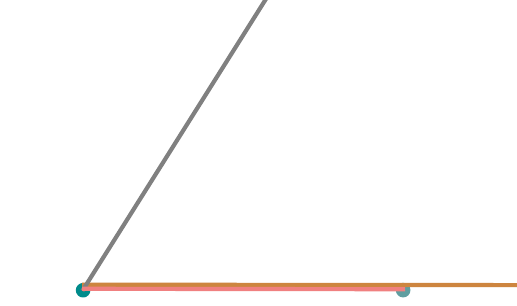}
    \caption{Visualizations of 3 primary failure cases. \textbf{Left:} Generated CAD sketch with no discernible pattern. \textbf{Middle:} Large gaps between primitives. \textbf{Right:} Extraneous primitives, zoom of the bottom left portion of a triangle where the bottom leg has an overlapping redundant line.}
    \label{fig:failure_cases}
\end{figure}

Figure \ref{fig:failure_cases} illustrates three representative failure modes observed in SketchDNN. 1) In some cases, the model generates sketches lacking any discernible geometric structure or pattern. We find that this issue can be mitigated by increasing the weight of the model prediction prior to the reverse interpolation step in Equation \ref{eqn:discrete_reverse_transition}, suggesting that high uncertainty in primitive type impairs the reverse process's ability to reconstruct coherent structure. We hypothesize that this failure mode could also be reduced through conditional generation, which would constrain the model's uncertainty during sampling. 2) A second failure involves primitives whose endpoints should be coincident but are instead separated by large gaps. This issue is likewise alleviated by increasing the contribution of the model prediction during the reverse step, indicating that the error stems from limitations in our parameter masking strategy. 3) The final failure mode involves the presence of redundant or extraneous primitives, often overlapping with valid geometry. We traced this issue to our training data, where several sketches contain overlapping primitives.

\section{Related Work}
Prior work in CAD sketch generation has predominantly focused on autoregressive approaches, treating sketches as sequences of tokens and leveraging language modeling techniques \cite{seff_sketchgraphs_2020, ganin_computer-aided_2021, para_sketchgen_2021, Willis2021EngineeringSG, Wu2021DeepCADAD, seff2021vitruvion, cadvlm}. In tangential domains with sketch-like data, for instance UI layout generation, DLT by \cite{Levi2023DLTCL} uses a joint continuous-discrete approach similar to ours except with Multinomial diffusion instead, whereas others solely use continuous diffusion \cite{Shabani2022HouseDiffusionVF, Cheng2023PLayPC}. More broadly for CAD diffusion as a whole, researchers have primarily utilized latent diffusion to generate CAD models \cite{Wang2024VQCADCD, Xu2022SkexGenAG, brepgen}.

Similarly, prior work on discrete diffusion has largely relied on the categorical distribution for discrete noise \cite{sohlthermo, hoogeboom2021argmax, hoogeboom2022autoregressive, lou2024discretescore, BondTaylor2021UnleashingTP}, or instead on applying continuous diffusion to discrete variables \cite{cohen2022bridge, yu2023latentdiffusionenergybasedmodel, han2023ssdlmsemiautoregressivesimplexbaseddiffusion}. Several researchers have proposed categorical relaxations outside the context of diffusion \cite{aitchisonlogisticnormal, maddison2017the, jang2017categorical, gaussianreparameterization}. \cite{maddison2017the} and \cite{jang2017categorical} propose a similar relaxation referred to as the Gumbel-Softmax distribution, where Gumbel, instead of Gaussian, vectors are mapped onto the probability simplex. \cite{gaussianreparameterization} use a pseudo-softmax transform to approximately map Gaussian values to the simplex, and \cite{aitchisonlogisticnormal} proposed a variant that expands the dimensionality of vectors by appending a zero before the softmax transform. 

In the context of discrete diffusion, the closest works to our Gaussian-Softmax methodology are \cite{han2023ssdlmsemiautoregressivesimplexbaseddiffusion} and \cite{karimi-mahabadi-etal-2024-tess} which superficially resemble our proposed approach, however they differ in 3 significant ways. 1) They use a heuristic reverse process whereas we follow a more principled approach and derive the reverse transition from the posterior of the forward process. 2) They use the cosine variance schedule directly which we demonstrate abruptly destroys information (see Figure \ref{fig:variance-aug}). 3) As a result, they require a one-hot projection scheme that drastically hinders superposition which we show is important for CAD generation.

\section{Conclusion}
In this work, we present SketchDNN, the first data-space diffusion model for CAD sketch generation. We overcome the key challenges of sketch generation, namely the heterogeneity of primitive parameterizations and the permutation invariance of primitives, via superposition and permutation-invariant denoising, respectively. To facilitate this we propose Gaussian-Softmax diffusion, a novel discrete diffusion paradigm that enables blended class labels. In the same vein, we introduce the Gaussian-Softmax distribution a novel variant of the Logistic-Normal distribution. We demonstrate that our model advances the state-of-the-art in terms of both fidelity and diversity of sketches. We hope that future work will build off of framework to improve CAD sketch generation, or apply our framework to other domains such as text generation.

\section*{Impact Statement}
This paper presents work whose goal is to advance the field of 
Machine Learning. There are many potential societal consequences 
of our work, none which we feel must be specifically highlighted here.

% Authors are \textbf{required} to include a statement of the potential 
% broader impact of their work, including its ethical aspects and future 
% societal consequences. This statement should be in an unnumbered 
% section at the end of the paper (co-located with Acknowledgements -- 
% the two may appear in either order, but both must be before References), 
% and does not count toward the paper page limit. In many cases, where 
% the ethical impacts and expected societal implications are those that 
% are well established when advancing the field of Machine Learning, 
% substantial discussion is not required, and a simple statement such 
% as the following will suffice:

% ``This paper presents work whose goal is to advance the field of 
% Machine Learning. There are many potential societal consequences 
% of our work, none which we feel must be specifically highlighted here.''

% The above statement can be used verbatim in such cases, but we 
% encourage authors to think about whether there is content which does 
% warrant further discussion, as this statement will be apparent if the 
% paper is later flagged for ethics review.

% In the unusual situation where you want a paper to appear in the
% references without citing it in the main text, use \nocite
\nocite{langley00}

\bibliographystyle{icml2025}

\begin{thebibliography}{34}
\providecommand{\natexlab}[1]{#1}
\providecommand{\url}[1]{\texttt{#1}}
\expandafter\ifx\csname urlstyle\endcsname\relax
  \providecommand{\doi}[1]{doi: #1}\else
  \providecommand{\doi}{doi: \begingroup \urlstyle{rm}\Url}\fi

\bibitem[Aitchison \& Shen(1980)Aitchison and Shen]{aitchisonlogisticnormal}
Aitchison, J. and Shen, S.~M.
\newblock Logistic-normal distributions: Some properties and uses.
\newblock \emph{Biometrika}, 67\penalty0 (2):\penalty0 261--272, 1980.
\newblock ISSN 00063444, 14643510.

\bibitem[Austin et~al.(2021)Austin, Johnson, Ho, Tarlow, and van~den Berg]{austin2021structured}
Austin, J., Johnson, D.~D., Ho, J., Tarlow, D., and van~den Berg, R.
\newblock Structured denoising diffusion models in discrete state-spaces.
\newblock In Beygelzimer, A., Dauphin, Y., Liang, P., and Vaughan, J.~W. (eds.), \emph{Advances in Neural Information Processing Systems}, 2021.

\bibitem[Bond-Taylor et~al.(2021)Bond-Taylor, Hessey, Sasaki, Breckon, and Willcocks]{BondTaylor2021UnleashingTP}
Bond-Taylor, S., Hessey, P., Sasaki, H., Breckon, T., and Willcocks, C.~G.
\newblock Unleashing transformers: Parallel token prediction with discrete absorbing diffusion for fast high-resolution image generation from vector-quantized codes.
\newblock In \emph{European Conference on Computer Vision}, 2021.

\bibitem[Cheng et~al.(2023)Cheng, Huang, Li, and Li]{Cheng2023PLayPC}
Cheng, C.-Y., Huang, F., Li, G., and Li, Y.
\newblock Play: Parametrically conditioned layout generation using latent diffusion.
\newblock In \emph{International Conference on Machine Learning}, 2023.

\bibitem[Cohen et~al.(2022)Cohen, Quispe, Corff, Ollion, and Moulines]{cohen2022bridge}
Cohen, M., Quispe, G., Corff, S.~L., Ollion, C., and Moulines, E.
\newblock Diffusion bridges vector quantized variational autoencoders.
\newblock In Chaudhuri, K., Jegelka, S., Song, L., Szepesvari, C., Niu, G., and Sabato, S. (eds.), \emph{Proceedings of the 39th International Conference on Machine Learning}, volume 162 of \emph{Proceedings of Machine Learning Research}, pp.\  4141--4156. PMLR, 17--23 Jul 2022.

\bibitem[Ganin et~al.(2021)Ganin, Bartunov, Li, Keller, and Saliceti]{ganin_computer-aided_2021}
Ganin, Y., Bartunov, S., Li, Y., Keller, E., and Saliceti, S.
\newblock Computer-aided design as language.
\newblock In Ranzato, M., Beygelzimer, A., Dauphin, Y., Liang, P., and Vaughan, J.~W. (eds.), \emph{Advances in Neural Information Processing Systems}, volume~34, pp.\  5885--5897. Curran Associates, Inc., 2021.

\bibitem[Han et~al.(2022)Han, Kumar, and Tsvetkov]{han2023ssdlmsemiautoregressivesimplexbaseddiffusion}
Han, X., Kumar, S., and Tsvetkov, Y.
\newblock Ssd-lm: Semi-autoregressive simplex-based diffusion language model for text generation and modular control.
\newblock In \emph{Annual Meeting of the Association for Computational Linguistics}, 2022.

\bibitem[Ho et~al.(2020)Ho, Jain, and Abbeel]{ho2020denoising}
Ho, J., Jain, A., and Abbeel, P.
\newblock Denoising diffusion probabilistic models.
\newblock In Larochelle, H., Ranzato, M., Hadsell, R., Balcan, M., and Lin, H. (eds.), \emph{Advances in Neural Information Processing Systems}, volume~33, pp.\  6840--6851. Curran Associates, Inc., 2020.

\bibitem[Hoogeboom et~al.(2021)Hoogeboom, Nielsen, Jaini, Forr{\'e}, and Welling]{hoogeboom2021argmax}
Hoogeboom, E., Nielsen, D., Jaini, P., Forr{\'e}, P., and Welling, M.
\newblock Argmax flows and multinomial diffusion: Learning categorical distributions.
\newblock In Beygelzimer, A., Dauphin, Y., Liang, P., and Vaughan, J.~W. (eds.), \emph{Advances in Neural Information Processing Systems}, 2021.

\bibitem[Hoogeboom et~al.(2022)Hoogeboom, Gritsenko, Bastings, Poole, van~den Berg, and Salimans]{hoogeboom2022autoregressive}
Hoogeboom, E., Gritsenko, A.~A., Bastings, J., Poole, B., van~den Berg, R., and Salimans, T.
\newblock Autoregressive diffusion models.
\newblock In \emph{International Conference on Learning Representations}, 2022.

\bibitem[Huijben et~al.(2021)Huijben, Kool, Paulus, and van Sloun]{Huijben2021ARO}
Huijben, I. A.~M., Kool, W., Paulus, M.~B., and van Sloun, R. J.~G.
\newblock A review of the gumbel-max trick and its extensions for discrete stochasticity in machine learning.
\newblock \emph{IEEE Transactions on Pattern Analysis and Machine Intelligence}, 45:\penalty0 1353--1371, 2021.

\bibitem[Jang et~al.(2017)Jang, Gu, and Poole]{jang2017categorical}
Jang, E., Gu, S., and Poole, B.
\newblock Categorical reparameterization with gumbel-softmax.
\newblock In \emph{International Conference on Learning Representations}, 2017.

\bibitem[Karimi~Mahabadi et~al.(2024)Karimi~Mahabadi, Ivison, Tae, Henderson, Beltagy, Peters, and Cohan]{karimi-mahabadi-etal-2024-tess}
Karimi~Mahabadi, R., Ivison, H., Tae, J., Henderson, J., Beltagy, I., Peters, M., and Cohan, A.
\newblock {TESS}: Text-to-text self-conditioned simplex diffusion.
\newblock In Graham, Y. and Purver, M. (eds.), \emph{Proceedings of the 18th Conference of the European Chapter of the Association for Computational Linguistics (Volume 1: Long Papers)}, pp.\  2347--2361, St. Julian{'}s, Malta, March 2024. Association for Computational Linguistics.

\bibitem[Langley(2000)]{langley00}
Langley, P.
\newblock Crafting papers on machine learning.
\newblock In Langley, P. (ed.), \emph{Proceedings of the 17th International Conference on Machine Learning (ICML 2000)}, pp.\  1207--1216, Stanford, CA, 2000. Morgan Kaufmann.

\bibitem[Levi et~al.(2023)Levi, Brosh, Mykhailych, and Perez]{Levi2023DLTCL}
Levi, E., Brosh, E., Mykhailych, M., and Perez, M.
\newblock Dlt: Conditioned layout generation with joint discrete-continuous diffusion layout transformer.
\newblock \emph{2023 IEEE/CVF International Conference on Computer Vision (ICCV)}, pp.\  2106--2115, 2023.

\bibitem[Lou et~al.(2024)Lou, Meng, and Ermon]{lou2024discretescore}
Lou, A., Meng, C., and Ermon, S.
\newblock Discrete diffusion modeling by estimating the ratios of the data distribution.
\newblock In Salakhutdinov, R., Kolter, Z., Heller, K., Weller, A., Oliver, N., Scarlett, J., and Berkenkamp, F. (eds.), \emph{Proceedings of the 41st International Conference on Machine Learning}, volume 235 of \emph{Proceedings of Machine Learning Research}, pp.\  32819--32848. PMLR, 21--27 Jul 2024.

\bibitem[Maddison et~al.(2017)Maddison, Mnih, and Teh]{maddison2017the}
Maddison, C.~J., Mnih, A., and Teh, Y.~W.
\newblock The concrete distribution: A continuous relaxation of discrete random variables.
\newblock In \emph{International Conference on Learning Representations}, 2017.

\bibitem[Nichol \& Dhariwal(2021)Nichol and Dhariwal]{nichol2021}
Nichol, A.~Q. and Dhariwal, P.
\newblock Improved denoising diffusion probabilistic models.
\newblock In Meila, M. and Zhang, T. (eds.), \emph{Proceedings of the 38th International Conference on Machine Learning}, volume 139 of \emph{Proceedings of Machine Learning Research}, pp.\  8162--8171. PMLR, 18--24 Jul 2021.

\bibitem[Para et~al.(2024)Para, Bhat, Guerrero, Kelly, Mitra, Guibas, and Wonka]{para_sketchgen_2021}
Para, W.~R., Bhat, S.~F., Guerrero, P., Kelly, T., Mitra, N., Guibas, L., and Wonka, P.
\newblock Sketchgen: generating constrained cad sketches.
\newblock In \emph{Proceedings of the 35th International Conference on Neural Information Processing Systems}, NIPS '21, Red Hook, NY, USA, 2024. Curran Associates Inc.
\newblock ISBN 9781713845393.

\bibitem[Peebles \& Xie(2022)Peebles and Xie]{peebles2023scalablediffusionmodelstransformers}
Peebles, W.~S. and Xie, S.
\newblock Scalable diffusion models with transformers.
\newblock \emph{2023 IEEE/CVF International Conference on Computer Vision (ICCV)}, pp.\  4172--4182, 2022.

\bibitem[Potapczynski et~al.(2020)Potapczynski, Loaiza-Ganem, and Cunningham]{gaussianreparameterization}
Potapczynski, A., Loaiza-Ganem, G., and Cunningham, J.~P.
\newblock Invertible gaussian reparameterization: revisiting the gumbel-softmax.
\newblock In \emph{Proceedings of the 34th International Conference on Neural Information Processing Systems}, NIPS '20, Red Hook, NY, USA, 2020. Curran Associates Inc.
\newblock ISBN 9781713829546.

\bibitem[Seff et~al.(2020)Seff, Ovadia, Zhou, and Adams]{seff_sketchgraphs_2020}
Seff, A., Ovadia, Y., Zhou, W., and Adams, R.~P.
\newblock Sketchgraphs: A large-scale dataset for modeling relational geometry in computer-aided design.
\newblock \emph{ArXiv}, abs/2007.08506, 2020.

\bibitem[Seff et~al.(2022)Seff, Zhou, Richardson, and Adams]{seff2021vitruvion}
Seff, A., Zhou, W., Richardson, N., and Adams, R.~P.
\newblock Vitruvion: A generative model of parametric {CAD} sketches.
\newblock In \emph{International Conference on Learning Representations}, 2022.

\bibitem[Shabani et~al.(2022)Shabani, Hosseini, and Furukawa]{Shabani2022HouseDiffusionVF}
Shabani, M.~A., Hosseini, S., and Furukawa, Y.
\newblock Housediffusion: Vector floorplan generation via a diffusion model with discrete and continuous denoising.
\newblock \emph{2023 IEEE/CVF Conference on Computer Vision and Pattern Recognition (CVPR)}, pp.\  5466--5475, 2022.

\bibitem[Sohl-Dickstein et~al.(2015)Sohl-Dickstein, Weiss, Maheswaranathan, and Ganguli]{sohlthermo}
Sohl-Dickstein, J., Weiss, E., Maheswaranathan, N., and Ganguli, S.
\newblock Deep unsupervised learning using nonequilibrium thermodynamics.
\newblock In Bach, F. and Blei, D. (eds.), \emph{Proceedings of the 32nd International Conference on Machine Learning}, volume~37 of \emph{Proceedings of Machine Learning Research}, pp.\  2256--2265, Lille, France, 07--09 Jul 2015. PMLR.

\bibitem[Vaswani et~al.(2017)Vaswani, Shazeer, Parmar, Uszkoreit, Jones, Gomez, Kaiser, and Polosukhin]{vaswani_attention_2017}
Vaswani, A., Shazeer, N., Parmar, N., Uszkoreit, J., Jones, L., Gomez, A.~N., Kaiser, L., and Polosukhin, I.
\newblock Attention is all you need.
\newblock In \emph{Proceedings of the 31st International Conference on Neural Information Processing Systems}, NIPS'17, pp.\  6000–6010, Red Hook, NY, USA, 2017. Curran Associates Inc.
\newblock ISBN 9781510860964.

\bibitem[Wang et~al.(2024)Wang, Zhao, Wang, Quan, and Yan]{Wang2024VQCADCD}
Wang, H., Zhao, M., Wang, Y., Quan, W., and Yan, D.-M.
\newblock Vq-cad: Computer-aided design model generation with vector quantized diffusion.
\newblock \emph{Comput. Aided Geom. Des.}, 111:\penalty0 102327, 2024.

\bibitem[Willis et~al.(2021)Willis, Jayaraman, Lambourne, Chu, and Pu]{Willis2021EngineeringSG}
Willis, K. D.~D., Jayaraman, P.~K., Lambourne, J., Chu, H., and Pu, Y.
\newblock Engineering sketch generation for computer-aided design.
\newblock \emph{2021 IEEE/CVF Conference on Computer Vision and Pattern Recognition Workshops (CVPRW)}, pp.\  2105--2114, 2021.

\bibitem[Wu et~al.(2021)Wu, Xiao, and Zheng]{Wu2021DeepCADAD}
Wu, R., Xiao, C., and Zheng, C.
\newblock Deepcad: A deep generative network for computer-aided design models.
\newblock \emph{2021 IEEE/CVF International Conference on Computer Vision (ICCV)}, pp.\  6752--6762, 2021.

\bibitem[Wu et~al.(2025)Wu, Khasahmadi, Katz, Jayaraman, Pu, Willis, and Liu]{cadvlm}
Wu, S., Khasahmadi, A.~H., Katz, M., Jayaraman, P.~K., Pu, Y., Willis, K., and Liu, B.
\newblock Cadvlm: Bridging language and vision in the generation of parametric cad sketches.
\newblock In Leonardis, A., Ricci, E., Roth, S., Russakovsky, O., Sattler, T., and Varol, G. (eds.), \emph{Computer Vision -- ECCV 2024}, pp.\  368--384, Cham, 2025. Springer Nature Switzerland.
\newblock ISBN 978-3-031-72897-6.

\bibitem[Xu et~al.(2024{\natexlab{a}})Xu, Xiang, Ye, Yao, Chu, and Li]{Xu_2024_CVPR}
Xu, H., Xiang, L., Ye, H., Yao, D., Chu, P., and Li, B.
\newblock Permutation equivariance of transformers and its applications.
\newblock In \emph{Proceedings of the IEEE/CVF Conference on Computer Vision and Pattern Recognition (CVPR)}, pp.\  5987--5996, June 2024{\natexlab{a}}.

\bibitem[Xu et~al.(2022)Xu, Willis, Lambourne, Cheng, Jayaraman, and Furukawa]{Xu2022SkexGenAG}
Xu, X., Willis, K. D.~D., Lambourne, J., Cheng, C.-Y., Jayaraman, P.~K., and Furukawa, Y.
\newblock Skexgen: Autoregressive generation of cad construction sequences with disentangled codebooks.
\newblock \emph{ArXiv}, abs/2207.04632, 2022.

\bibitem[Xu et~al.(2024{\natexlab{b}})Xu, Lambourne, Jayaraman, Wang, Willis, and Furukawa]{brepgen}
Xu, X., Lambourne, J., Jayaraman, P.~K., Wang, Z., Willis, K. D.~D., and Furukawa, Y.
\newblock Brepgen: A b-rep generative diffusion model with structured latent geometry.
\newblock \emph{ACM Transactions on Graphics (TOG)}, 43:\penalty0 1 -- 14, 2024{\natexlab{b}}.

\bibitem[Yu et~al.(2022)Yu, Xie, Ma, Jia, Pang, Gao, Zhu, Zhu, and Wu]{yu2023latentdiffusionenergybasedmodel}
Yu, P., Xie, S., Ma, X., Jia, B., Pang, B., Gao, R., Zhu, Y., Zhu, S.-C., and Wu, Y.~N.
\newblock Latent diffusion energy-based model for interpretable text modelling.
\newblock In Chaudhuri, K., Jegelka, S., Song, L., Szepesvari, C., Niu, G., and Sabato, S. (eds.), \emph{Proceedings of the 39th International Conference on Machine Learning}, volume 162 of \emph{Proceedings of Machine Learning Research}, pp.\  25702--25720. PMLR, 17--23 Jul 2022.

\end{thebibliography}

%%%%%%%%%%%%%%%%%%%%%%%%%%%%%%%%%%%%%%%%%%%%%%%%%%%%%%%%%%%%%%%%%%%%%%%%%%%%%%%
%%%%%%%%%%%%%%%%%%%%%%%%%%%%%%%%%%%%%%%%%%%%%%%%%%%%%%%%%%%%%%%%%%%%%%%%%%%%%%%
% APPENDIX
%%%%%%%%%%%%%%%%%%%%%%%%%%%%%%%%%%%%%%%%%%%%%%%%%%%%%%%%%%%%%%%%%%%%%%%%%%%%%%%
%%%%%%%%%%%%%%%%%%%%%%%%%%%%%%%%%%%%%%%%%%%%%%%%%%%%%%%%%%%%%%%%%%%%%%%%%%%%%%%
\newpage
\appendix
\onecolumn
\section{Gaussian-Softmax Distribution}

The Gaussian-Softmax distribution ($\mathcal{GS}$), a variation of the Logistic-Normal distribution introduced by Aitchison \cite{aitchisonlogisticnormal}, is the distribution of a Gaussian vector that has undergone the softmax transformation. The probability density is:
\begin{align*}
    p(\vec{y} | \vec{\mu}, \sigma^2) &= D^{-\frac{1}{2}} (2\pi\sigma^2)^{\frac{1-D}{2}} \left( \prod_{i=1}^{D} y_i \right)^{-1} \\
    &\quad \times \exp\left\{-\frac{1}{2\sigma^2} \left[ \sum_{i \ne D} \left( \log\frac{y_i}{y_D} - \mu_i + \mu_D \right)^2 - \frac{1}{D} \left( \sum_{i \ne D} \log\frac{y_i}{y_D} - \mu_i + \mu_D \right)^2 \right] \right\}
\end{align*}
and the derivation of the density is provided in \ref{gauss-soft-density}.

 \subsection{Derivation of Cumulative Forward Transition}\label{cumulative_forward}
 A simple derivation for the cumulative transition is:
% \begin{align*}
%     % Equation 1 with an explanatory comment above
%     &\text{\textit{Conditional probability for the next timestep $t+2$:}}\\
%     p(\vec{x}_{t+2}|\vec{x}_t) &= \text{softmax}\left\{\sqrt{a_{t+2}}\log{\left[\text{softmax}\left\{\sqrt{a_{t+1}}\log{\vec{x}_t}+\sqrt{1-a_{t+1}}\vec{\epsilon}\right\}\right]} + \sqrt{1-a_{t+2}}\vec{\epsilon}\right\} \\
%     &\text{\textit{Expand:}} \\
%     &= \text{softmax}\left\{\sqrt{a_{t+2}}\left(\sqrt{a_{t+1}}\log{\vec{x}_t}+\sqrt{1-a_{t+1}}\vec{\epsilon}+C\right) + \sqrt{1-a_{t+2}}\vec{\epsilon}\right\} \\
%     &\text{\textit{Combine coefficients:}} \\
%     &= \text{softmax}\left\{\sqrt{a_{t+2}a_{t+1}}\log{\vec{x}_t} + \sqrt{a_{t+2}(1-a_{t+1})}\vec{\epsilon} + \sqrt{a_{t+2}}C + \sqrt{1-a_{t+2}}\vec{\epsilon}\right\} \\
%     &\text{\textit{Constant disappears due to shift invariance of softmax:}} \\
%     &= \text{softmax}\left\{\sqrt{a_{t+2}a_{t+1}}\log{\vec{x}_t} + \sqrt{a_{t+2}(1-a_{t+1})}\vec{\epsilon} + \sqrt{1-a_{t+2}}\vec{\epsilon}\right\} \\
%     &\text{\textit{Sum of Gaussians is another Gaussian with variances summed:}} \\
%     &= \text{softmax}\left\{\sqrt{a_{t+2}a_{t+1}}\log{\vec{x}_t} + \sqrt{a_{t+2}(1-a_{t+1}) + 1-a_{t+2}}\vec{\epsilon}\right\} \\
%     &\text{\textit{Merging terms simply accumulates `a' terms:}} \\
%     &= \text{softmax}\left\{\sqrt{a_{t+2}a_{t+1}}\log{\vec{x}_t} + \sqrt{1-a_{t+2}a_{t+1}}\vec{\epsilon}\right\}
% \end{align*}
\begin{align*}
    p\left(\vec{x}_{t+2} \mid \vec{x}_t\right) 
    &= \text{softmax}\left\{\sqrt{\alpha_{t+2}} \log\left[\text{softmax}\left\{\sqrt{\alpha_{t+1}} \log\left(\vec{x}_t\right) + \sqrt{1 - \alpha_{t+1}} \vec{\epsilon}\right\}\right] + \sqrt{1 - \alpha_{t+2}} \vec{\epsilon}\right\} \\
    \intertext{Expand terms:}
    &= \text{softmax}\left\{\sqrt{\alpha_{t+2}}\left(\sqrt{\alpha_{t+1}} \log\left(\vec{x}_t\right) + \sqrt{1 - \alpha_{t+1}} \vec{\epsilon} + C\right) + \sqrt{1 - \alpha_{t+2}} \vec{\epsilon}\right\} \\
    \intertext{Reduce terms:}
    &= \text{softmax}\left\{\sqrt{\alpha_{t+2} \alpha_{t+1}} \log\left(\vec{x}_t\right) + \sqrt{\alpha_{t+2}(1 - \alpha_{t+1})} \vec{\epsilon} + \sqrt{\alpha_{t+2}} C + \sqrt{1 - \alpha_{t+2}} \vec{\epsilon}\right\} \\
    \intertext{Constants disappear due to shift invariance of softmax:}
    &= \text{softmax}\left\{\sqrt{\alpha_{t+2} \alpha_{t+1}} \log\left(\vec{x}_t\right) + \sqrt{\alpha_{t+2}(1 - \alpha_{t+1})} \vec{\epsilon} + \sqrt{1 - \alpha_{t+2}} \vec{\epsilon}\right\} \\
    \intertext{Sum of Gaussians is another Gaussian with summed variances:}
    &= \text{softmax}\left\{\sqrt{\alpha_{t+2} \alpha_{t+1}} \log\left(\vec{x}_t\right) + \sqrt{\alpha_{t+2}(1 - \alpha_{t+1}) + 1 - \alpha_{t+2}} \vec{\epsilon}\right\} \\
    \intertext{Merging terms accumulates the \(a\) terms:}
    &= \text{softmax}\left\{\sqrt{\alpha_{t+2} \alpha_{t+1}} \log\left(\vec{x}_t\right) + \sqrt{1 - \alpha_{t+2} \alpha_{t+1}} \vec{\epsilon}\right\}
\end{align*}

therefore iteratively applying the forward transition will simply accumulate the variance schedule terms.

\subsection{Derivation of Gaussian Softmax Density}\label{gauss-soft-density}
% PROOF OF GAUSSIAN SOFTMAX DENSITY
 Our strategy is to use the change of variables formula:
 $$p'(\vec{y})=p\left(h^{-1}\left(\vec{y}\right)\right)\text{Det}\left[J\left(\vec{h}\left(\vec{y}\right)\right)\right]$$ where $\vec{h}(\vec{y})$ is some invertible function and $\text{Det}(J(\vec{h}(\vec{y}))$ is the determinant of the jacobian. More specifically, 
 $$\text{softmax}\{\vec{y}\}=\text{softmax}\{\vec{y}-\vec1\cdot y_D\}$$ holds due to the shift invariance of softmax, thus our strategy is to first find the density of $\vec{y'}=[y_1-y_D,y_2-y_D,...,0]$, as this ``centered" form turns the softmax into an invertible function $\vec{h}(\vec{y'})=\text{softmax}\{\vec{y'}\}$ where the inverse is $\vec{h}^{-1}(\vec{x})=\log(\vec{x}/x_D)$. The derivation is as follows for $\vec{y} \sim \mathcal{N}(\vec\mu, \vec\sigma^2\mathbf{I})$, additionally for brevity we aggregate all factors into a single variable $C$:
\begin{align*}
    &\intertext{Marginalizing over $y_D$ yields the density of the ``centered" density:}
    p(\vec{y'}) &= \int_{-\infty}^{\infty} p(\vec{y'} \mid y_D) p(y_D) \, dy_D \\
    &= \int_{-\infty}^{\infty} \left[\prod_{i \ne D} \left(2 \pi \sigma^2 \right)^{-\frac{1}{2}} \exp\left\{-\frac{1}{2 \sigma^2} (y_i - y_D - \mu_i)^2 \right\}\right] \left[\left(2 \pi \sigma^2 \right)^{-\frac{1}{2}} \exp\left\{-\frac{1}{2 \sigma^2} (y_D - \mu_D)^2 \right\}\right] \, dy_D \\
    &\intertext{Expand terms:} 
    &= (2 \pi \sigma^2)^{-\frac{D}{2}} \int_{-\infty}^{\infty} \left[\prod_{i \ne D} \exp\left\{-\frac{1}{2 \sigma^2} (y_i^2 - 2 y_i \mu_i + \mu_i^2) \right\}\right] \\
    &\quad \times \exp\left\{-\frac{1}{2 \sigma^2} \left(y_D^2 - 2 y_D \mu_D + \mu_D^2 + (D - 1) y_D^2 + 2 y_D \sum_{i \ne D} \mu_i - y_i \right)\right\} \, dy_D \\
\end{align*}
\begin{align*}
    &\intertext{Reduce $y_D$ terms:} 
    &= C \left[\prod_{i \ne D} \exp\left\{-\frac{1}{2 \sigma^2} (y_i - \mu_i)^2 \right\}\right] \\
    &\quad \times \int_{-\infty}^{\infty} \exp\left\{-\frac{1}{2 \sigma^2} \left(y_D^2 - 2 y_D \mu_D + \mu_D^2 + (D - 1) y_D^2 + 2 y_D \sum_{i \ne D} \mu_i - y_i\right)\right\} \, dy_D \\
    &\intertext{Reduce terms:} 
    &= C \int_{-\infty}^{\infty} \exp\left\{-\frac{1}{2 \sigma^2} \left(D y_D^2 - 2 y_D \left(\mu_D + \sum_{i \ne D} y_i - \mu_i \right) + \mu_D^2 \right)\right\} \, dy_D \\
    &\intertext{Complete the square:} 
    &= C \exp\left\{-\frac{1}{2 \sigma^2} \mu_D^2 \right\} \int_{-\infty}^{\infty} \exp\left\{-\frac{1}{2 \sigma^2} \left(D y_D^2 - 2 y_D \left(\mu_D + \sum_{i \ne D} y_i - \mu_i \right)\right)\right\} \, dy_D \\
    &= C \int_{-\infty}^{\infty} \exp\left\{-\frac{1}{2 \sigma^2} \left(\sqrt{D} y_D - \frac{\left(\mu_D + \sum_{i \ne D} y_i - \mu_i \right)}{\sqrt{D}}\right)^2 \right\} \, dy_D \\
    &\intertext{The integrand is a Gaussian density in terms of $y_D$:}
    &= (2 \pi \sigma^2)^{-\frac{D}{2}} \left[\prod_{i \ne D} \exp\left\{-\frac{1}{2 \sigma^2} (y_i - \mu_i)^2 \right\}\right] \exp\left\{-\frac{1}{2 \sigma^2} \mu_D^2 \right\} \\
    &\quad \times \exp\left\{-\frac{1}{2 \sigma^2} \frac{\left(\mu_D + \sum_{i \ne D} y_i - \mu_i \right)^2}{D}\right\} \sqrt{\frac{2 \pi \sigma^2}{D}}
\end{align*}

 We can simplify this further using the fact that shifting the mean by any constant scalar does not affect the density due to the shift invariance of the softmax operation i.e., $p(\vec{y}|\vec{u},\sigma^2)=p(\vec{y}|\vec{u} + c\vec{1},\sigma^2)$, thus if we use $c=-\mu_D$ the density becomes:
 % \[
 % =D^{-\frac{1}{2}}(2\pi\sigma^2)^\frac{1-D}{2}\exp{\{-\frac{1}{2\sigma^2}[\sum_{i \ne D}(y_i-\mu_i+\mu_D)^2-\frac{1}{D}(\sum_{i \ne D}y_i-\mu_i+\mu_D)^2]\}}
 % \]
 \[
= D^{-\frac{1}{2}} \left(2 \pi \sigma^2\right)^{\frac{1 - D}{2}} 
\exp\left\{-\frac{1}{2 \sigma^2} \left[\sum_{i \ne D} \left(y_i - \mu_i + \mu_D\right)^2 
- \frac{1}{D} \left(\sum_{i \ne D} y_i - \mu_i + \mu_D\right)^2\right]\right\}
\]

 Now that we have the density of $\vec{y'}$ we can use a straightforward application of the change of variables formula, with the known result that the determinant of the Jacobian of the softmax is $(\prod_i^Dy_i)^{-1}$ \cite{jang2017categorical,maddison2017the} to obtain:
\begin{align*}
    p(\vec{y} | \vec{\mu}, \sigma^2) &= D^{-\frac{1}{2}} (2\pi\sigma^2)^{\frac{1-D}{2}} \left( \prod_{i=1}^{D} y_i \right)^{-1} \\
    &\quad \times \exp\left\{-\frac{1}{2\sigma^2} \left[ \sum_{i \ne D} \left( \log\frac{y_i}{y_D} - \mu_i + \mu_D \right)^2 - \frac{1}{D} \left( \sum_{i \ne D} \log\frac{y_i}{y_D} - \mu_i + \mu_D \right)^2 \right] \right\}
\end{align*}

\subsection{Derivation of Posterior for Gaussian-Softmax}\label{reverse-derivation}
 % Proof of the posterior of gaussian softmax diffusion
 For the reverse process we sample from the posterior distribution $p(\vec{x}_{t-1}|\vec{x}_t,\vec{x}_0)$, using the same setup as in DDPM \cite{ho2020denoising}: 
 $$p(\vec{x}_{t-1}|\vec{x}_t,\vec{x}_0)
 =\frac{p(\vec{x}_{t-1},\vec{x}_t,\vec{x}_0)}{p(\vec{x}_t,\vec{x}_0)}
 =\frac{p(\vec{x}_t|\vec{x}_{t-1},\vec{x}_0)p(\vec{x}_{t-1},\vec{x}_0)}{p(\vec{x}_t,\vec{x}_0)}
 =\frac{p(\vec{x}_t|\vec{x}_{t-1},\vec{x}_0)p(\vec{x}_{t-1}|\vec{x}_0)}{p(\vec{x}_t|\vec{x}_0)}$$ 
 and due to the Markov property 
 $p(\vec{x}_t|\vec{x}_{t-1},\vec{x}_0)
 =p(\vec{x}_t|\vec{x}_{t-1})$ 
the posterior simplifies to: 
 $$p(\vec{x}_{t-1}|\vec{x}_t,\vec{x}_0)=\frac{p(\vec{x}_t|\vec{x}_{t-1})p(\vec{x}_{t-1}|\vec{x}_0)}{p(\vec{x}_t|\vec{x}_0)} \propto p(\vec{x}_t|\vec{x}_{t-1})p(\vec{x}_{t-1}|\vec{x}_0)$$
 fortunately we have access to the necessary densities which are simply:
\begin{enumerate}
\item
%p(x_t-1|x_0) --------------
\[
p(\vec{x}_{t-1}|\vec{x}_0) \propto
\left(\prod_i^Dx_{t-1,i}\right)^{-1}
\exp{\left[-\frac{1}{2(1-\overline{a}_{t-1})}
\sum_{i \ne D}
v_i^2\right]}
\exp{\left[\frac{1}{2D(1-\overline{a}_{t-1})}
\left(\sum_{i \ne D}v_i\right)^2\right]}
\] \qquad where $v_i=\log\frac{x_{t-1,i}}{x_{t-1,D}}-\sqrt{\overline{a}_{t-1}}\log\frac{x_{0,i}}{x_{0,D}}$
\item
%p(x_t|p_x-1) ------------
\[
p(\vec{x}_t|\vec{x}_{t-1}) \propto
\exp{\left[-\frac{1}{2(1-a_t)}
\sum_{i \ne D}
r_i^2\right]}\exp{\left[\frac{1}{2D(1-a_t)}
\left(\sum_{i \ne D}
r_i\right)^2\right]}
\] \qquad where $r_i=\log\frac{x_{t,i}}{x_{t,D}}-\sqrt{a_t}\log\frac{x_{t-1,i}}{x_{t-1,D}}$
\end{enumerate}
Focusing on the first exponential terms with simplified notation where $z_i=\log\frac{x_{t,i}}{x_{t,D}}$, $y_i=\log\frac{x_{t-1,i}}{x_{t-1,D}}$, $x_i=\log\frac{x_{0,i}}{x_{0,D}}$, $\sigma^2_t=1-a_t$, and $\overline\sigma^2_{t-1}=1-\overline a_{t-1}$
\begin{align*}
    &\exp{\bigg[-\frac{1}{2(1-\overline{a}_{t-1})}
    \sum_{i \ne D}
    v_i^2\bigg]}
    \exp{\bigg[-\frac{1}{2(1-a_t)}
    \sum_{i \ne D}
    r_i^2\bigg]} \\[1em]
    &= \exp\bigg\{
    -\frac{1}{2\overline\sigma^2_{t-1}}\sum_{i \ne D}
    \left(y_i-\sqrt{\overline{a}_{t-1}}x_i\right)^2
    -\frac{1}{2\sigma^2_t}
    \sum_{i \ne D}
    \left(z_i-\sqrt{a_t}y_i\right)^2\bigg\} \\[1em]
    &= \exp\bigg\{
    -\frac{1}{2}\bigg[\sum_{i \ne D}
    \frac{1}{\overline\sigma^2_{t-1}}\left(y_i-\sqrt{\overline{a}_{t-1}}x_i\right)^2
    +\frac{1}{\sigma^2_t}
    \left(z_i-\sqrt{a_t}y_i\right)^2\bigg]\bigg\} \\[1em]
    &= \exp\bigg\{
    -\frac{1}{2}\bigg[\sum_{i \ne D}
    \frac{1}{\overline\sigma^2_{t-1}}\left(y_i^2-2y_i\sqrt{\overline{a}_{t-1}}x_i+\overline{a}_{t-1}x_i^2\right) \\
    &\qquad\qquad\qquad +\frac{1}{\sigma^2_t}
    \left(z_i^2-2\sqrt{a_t}y_iz_i+a_ty_i^2\right)\bigg]\bigg\} \\[1em]
    &\propto \exp\bigg\{
    -\frac{1}{2}\bigg[\sum_{i \ne D}
    \frac{y_i^2}{{\overline\sigma^2_{t-1}}}
    -\frac{2y_i\sqrt{\overline{a}_{t-1}}x_i}{{\overline\sigma^2_{t-1}}}
    -\frac{2\sqrt{a_t}y_iz_i}{\sigma^2_t}
    +\frac{a_ty_i^2}{\sigma^2_t}\bigg]\bigg\} \\[1em]
    &= \exp\bigg\{
    -\frac{1}{2}\bigg[\sum_{i \ne D}
    \Big(\frac{1}{{\overline\sigma^2_{t-1}}} + \frac{a_t}{\sigma^2_t}\Big)y_i^2
    -2\Big(\frac{\sqrt{\overline{a}_{t-1}}x_i}{{\overline\sigma^2_{t-1}}} + \frac{\sqrt{a_t}z_i}{\sigma^2_t}\Big)y_i\bigg]\bigg\}
\end{align*}
observe that we can read the posterior mean and variance as:
$$\mu_{t-1,i}=\left(\frac{\sqrt{\overline{a}_{t-1}}x_i}{{\overline\sigma^2_{t-1}}} + \frac{\sqrt{a_t}z_i}{\sigma^2_t}\right)
\sigma^2_{t-1}=\frac{\sqrt{a_t}(1-\overline a_{t-1})z_i+\sqrt{\overline a_{t-1}}(1-a_t)x_i}{1-\overline a_t}$$ 
$$\sigma^2_{t-1}=\left(\frac{1}{{\overline\sigma^2_{t-1}}} + \frac{a_t}{\sigma^2_t}\right)^{-1}=\frac{(1-a_t)(1-\overline a_{t-1})}{1-\overline a_t}$$
since the form has to be proportional to
$\exp{\{-\frac{1}{2\sigma^2}(y_i-\mu)^2\}}$.

Similarly for the second exponential term:
\begin{align*}
&\exp{\left[\frac{1}{2D(1-\overline{a}_{t-1})}
\left(\sum_{i \ne D}v_i\right)^2\right]}
\exp{\left[\frac{1}{2D(1-a_t)}
\left(\sum_{i \ne D}
r_i\right)^2\right]} \\
&=
\exp{\left\{\frac{1}{2D}\left[\frac{1}{\overline\sigma^2_{t-1}}
\left(\sum_{i \ne D}
y_i-\sqrt{\overline{a}_{t-1}}x_i\right)^2
+\frac{1}{\sigma^2_t}
\left(\sum_{i \ne D}
z_i-\sqrt{a_t}y_i\right)^2\right]\right\}} \\
&=
\exp\bigg\{\frac{1}{2D}\bigg[\frac{1}{\overline\sigma^2_{t-1}}
\bigg(\mathbf{\sum_{i \ne D}
(y_i-\sqrt{\overline{a}_{t-1}}x_i)^2} +2\sum_{j<i,i \ne D}
(y_i-\sqrt{\overline{a}_{t-1}}x_i)(y_j-\sqrt{\overline{a}_{t-1}}x_j)\bigg) \\
& +\frac{1}{\sigma^2_t}
\bigg(\mathbf{\sum_{i \ne D}
(z_i-\sqrt{a_t}y_i)^2} +2\sum_{j<i,i \ne D}(z_i-\sqrt{a_t}y_i)(z_j-\sqrt{a_t}y_j)\bigg)\bigg]\bigg\}
\end{align*}
The terms in bold correspond exactly with the first exponential term, and imply the same posterior mean and variance, further more the remaining terms are proportional to:
\begin{align*}
&\exp\frac{1}{2D}\left[2\sum_{j<i,i \ne D}
\frac{1}{\overline\sigma^2_{t-1}}
\left(y_iy_j-\sqrt{\overline{a}_{t-1}}x_jy_i-\sqrt{\overline{a}_{t-1}}x_iy_j\right)
+\frac{1}{\sigma^2_t}
\left(-\sqrt{a_t}z_iy_j-\sqrt{a_t}z_jy_i+a_ty_iy_j\right)\right] \\
&=
\exp\frac{1}{2D}\left[2\sum_{j<i,i \ne D}
\frac{1}{\overline\sigma^2_{t-1}}
\left(y_iy_j-\sqrt{\overline{a}_{t-1}}x_jy_i-\sqrt{\overline{a}_{t-1}}x_iy_j\right)
+\frac{1}{\sigma^2_t}
\left(-\sqrt{a_t}z_iy_j-\sqrt{a_t}z_jy_i+a_ty_iy_j\right)\right] \\
&=
\exp\frac{1}{2D}\left[2\sum_{j<i,i \ne D}
\left(\frac{1}{\overline\sigma^2_{t-1}}+\frac{a_t}{\sigma^2_t}\right)
y_iy_j
-\left(\frac{\sqrt{\overline{a}_{t-1}}x_j}{\overline\sigma^2_{t-1}} + \frac{\sqrt{a_t}z_j}{\sigma^2_t}\right)
y_i
-\left(\frac{\sqrt{\overline{a}_{t-1}}x_i}{\overline\sigma^2_{t-1}} + \frac{\sqrt{a_t}z_i}{\sigma^2_t}\right)
y_j\right]
\end{align*}
which again imply the same posterior mean and variance since the form has to be proportional to:
$$
\exp{\left\{2\left[\sum_{j<i,i \ne D}
\frac{y_iy_j}{\sigma^2}
-\frac{\mu_j}{\sigma^2}
y_i
-\frac{\mu_i}{\sigma^2}
y_j
+\frac{\mu_i\mu_j}{\sigma^2}\right]\right\}}
$$
Thus all terms agree on the same posterior mean and variance of:
$$\mu_{t-1,i}=\left(\frac{\sqrt{\overline{a}_{t-1}}x_i}{{\overline\sigma^2_{t-1}}} + \frac{\sqrt{a_t}z_i}{\sigma^2_t}\right)
\sigma^2_{t-1}=\frac{\sqrt{a_t}(1-\overline a_{t-1})z_i+\sqrt{\overline a_{t-1}}(1-a_t)x_i}{1-\overline a_t}$$ 
$$\sigma^2_{t-1}=\left(\frac{1}{{\overline\sigma^2_{t-1}}} + \frac{a_t}{\sigma^2_t}\right)^{-1}=\frac{(1-a_t)(1-\overline a_{t-1})}{1-\overline a_t}$$
We can simplify the posterior mean by utilizing the shift invariance property, where we observe that:
\begin{align*}
&\mu_{t-1,i}
=\frac{\sqrt{a_t}(1-\overline a_{t-1})(\log{x_{t,i}}-\log{x_{t,D}})+\sqrt{\overline a_{t-1}}(1-a_t)(\log{x_{0,i}}-\log{x_{0,D}})}{1-\overline a_t} \\
&=\frac{\sqrt{a_t}(1-\overline a_{t-1})\log{x_{t,i}}+\sqrt{\overline a_{t-1}}(1-a_t)\log{x_{0,i}}}{1-\overline a_t}
-
\frac{\sqrt{a_t}(1-\overline a_{t-1})\log{x_{t,D}}+\sqrt{\overline a_{t-1}}(1-a_t)\log{x_{0,D}}}{1-\overline a_t} \\
&=
\frac{\sqrt{a_t}(1-\overline a_{t-1})\log{x_{t,i}}+\sqrt{\overline a_{t-1}}(1-a_t)\log{x_{0,i}}}{1-\overline a_t}+C
\end{align*}
so the posterior mean can be simplified as:
$$\mu_{t-1,i}=\frac{\sqrt{a_t}(1-\overline a_{t-1})\log{x_{t,i}}+\sqrt{\overline a_{t-1}}(1-a_t)\log{x_{0,i}}}{1-\overline a_t}$$

Since all the terms agree on the same posterior mean and variance, and furthermore the posterior density has the same form as the Gaussian-Softmax distribution, we can conclude that $p(\vec{x}_{t-1}|\vec{x}_t,\vec{x}_0)=p(\vec{x}_{t-1}|\vec{\mu}_{t-1},\sigma_{t-1}^2\mathbf{I})$

\subsection{Derivation of Variance Schedule Augmentation}\label{variance-augmentation}
As shown in Figure \ref{fig:variance-aug}, we need to augment our chosen variance schedule to ensure that the class labels are gradually noised. Taking inspiration from Categorical diffusion, our desideratum is to smoothly noise the class label such that the argmax of $\mathbf{x}_t$ follows the distribution $\mathcal{C}(\overline{b_t} \mathbf{x}_0 + \frac{1}{D}(1-\overline{b_t})\mathbf{1})$ where $\mathcal{C}$ is the categorical distribution and $b_0,b_1,\ldots,b_T$ is a variance schedule of our choosing. Unfortunately there is no closed form formula to determine the argmax of a Gaussian vector, so we instead approximate a Gaussian vector with a Gumbel vector. A useful property of the Gumbel distribution is that it can be used to reparameterize the Categorical distribution where $\text{argmax}\{a\log \mathbf{p} + \mathbf{g}\} \sim \mathcal{C}(\text{softmax}\{a\log \mathbf{p}\})$, $\mathbf{g}\sim \mathcal{G}(0,1)$, and $\mathcal{G}$ is the Gumbel distribution \cite{Huijben2021ARO}. Then considering the forward process in Gaussian-Softmax diffusion we can derive:
\begin{align*}
    \text{argmax}\{\mathbf{x}_t\} 
    &\approx \text{argmax}\left\{\text{softmax}\left(\sqrt{\overline{\alpha_t}}\log{\left(k\mathbf{x}_0 + \frac{1-k}{D}\right)} + \sqrt{1 - \overline{\alpha_t}} \mathbf{g}\right)\right\}.
\end{align*}

Simplifying the expression inside the argmax:
\begin{align*}
    &= \text{argmax}\left\{\sqrt{\overline{\alpha_t}} \log{\left(k\mathbf{x}_0 + \frac{1-k}{D} \mathbf{1}\right)} + \sqrt{1 - \overline{\alpha_t}} \mathbf{g}\right\} \\
    &= \text{argmax}\left\{\sqrt{\frac{\overline{\alpha_t}}{1 - \overline{\alpha_t}}} \log{\left(k\mathbf{x}_0 + \frac{1-k}{D} \mathbf{1}\right)} + \mathbf{g}\right\} 
    \sim 
    \text{softmax}\left\{\sqrt{\frac{\overline{\alpha_t}}{1 - \overline{\alpha_t}}} \log{\left(k\mathbf{x}_0 + \frac{1-k}{D} \mathbf{1}\right)}\right\}
\end{align*}

We aim to satisfy:
\[
    \text{softmax}\left\{\sqrt{\frac{\overline{\alpha_t}}{1 - \overline{\alpha_t}}} \log{\left(k\mathbf{x}_0 + \frac{1-k}{D} \mathbf{1}\right)}\right\} = \overline{b}_t \mathbf{x}_0 + \frac{1 - \overline{b}_t}{D} \mathbf{1}
\]

Taking the logarithm of both sides gives:
\[
    \sqrt{\frac{\overline{\alpha_t}}{1 - \overline{\alpha_t}}} \log{\left(k\mathbf{x}_0 + \frac{1-k}{D} \mathbf{1}\right)} + \mathbf{c} = \log{\left(\overline{b}_t \mathbf{x}_0 + \frac{1 - \overline{b}_t}{D} \mathbf{1}\right)}
\]

Assuming without loss of generality that \(\mathbf{x}_0 = [1, 0, \ldots, 0]\), we have:
\[
    \sqrt{\frac{\overline{\alpha_t}}{1 - \overline{\alpha_t}}} \left[\log{\left(k + \frac{1-k}{D}\right)}, \log{\left(\frac{1-k}{D}\right)}, \ldots\right] + \mathbf{c} = \left[\log{\left(\overline{b}_t + \frac{1 - \overline{b}_t}{D}\right)}, \log{\left(\frac{1 - \overline{b}_t}{D}\right)}, \ldots\right]
\]

Since \(\mathbf{c}\) is a free parameter, we can set:
\[
    \mathbf{c} = \left[\log{\left(\overline{b}_t + \frac{1 - \overline{b}_t}{D}\right)} - \sqrt{\frac{\overline{\alpha_t}}{1 - \overline{\alpha_t}}} \log{\left(k + \frac{1-k}{D}\right)}\right] \mathbf{1}
\]

This reduces the equation to:
\[
    \sqrt{\frac{\overline{\alpha_t}}{1 - \overline{\alpha_t}}} \left[0, \log{\left(\frac{1-k}{(D-1)k + 1}\right)}, \ldots\right] = \left[0, \log{\left(\frac{1 - \overline{b}_t}{(D-1)\overline{b}_t + 1}\right)}, \ldots\right]
\]

Thus, we deduce:
\[
    \sqrt{\frac{\overline{\alpha_t}}{1 - \overline{\alpha_t}}} = \log{\left(\frac{1 - \overline{b}_t}{(D-1)\overline{b}_t + 1}\right)}/\log{\left(\frac{1-k}{(D-1)k + 1}\right)}
\]

Finally, isolating \(\overline{\alpha_t}\) yields:
\[
    \overline{\alpha_t} = \frac{n^2}{n^2 + m^2},
\text{where} \quad
    n = \log{\left(\frac{1 - \overline{b}_t}{(D-1)\overline{b}_t + 1}\right)}, \quad m = \log{\left(\frac{1-k}{(D-1)k + 1}\right)}.
\]

% You can have as much text here as you want. The main body must be at most $8$ pages long.
% For the final version, one more page can be added.
% If you want, you can use an appendix like this one.  

% The $\mathtt{\backslash onecolumn}$ command above can be kept in place if you prefer a one-column appendix, or can be removed if you prefer a two-column appendix.  Apart from this possible change, the style (font size, spacing, margins, page numbering, etc.) should be kept the same as the main body.
%%%%%%%%%%%%%%%%%%%%%%%%%%%%%%%%%%%%%%%%%%%%%%%%%%%%%%%%%%%%%%%%%%%%%%%%%%%%%%%
%%%%%%%%%%%%%%%%%%%%%%%%%%%%%%%%%%%%%%%%%%%%%%%%%%%%%%%%%%%%%%%%%%%%%%%%%%%%%%%

\end{document}